\pdfoutput=1
\documentclass[11pt]{article}
\pdftrailerid{}
\usepackage[letterpaper,margin=1in]{geometry}
\usepackage[utf8]{inputenc}
\usepackage[T1]{fontenc}
\usepackage{times}
\usepackage{microtype}
\usepackage{graphicx}
\usepackage{booktabs}
\usepackage{xcolor}
\usepackage{amsmath,amssymb}
\usepackage{array}
\usepackage{longtable}
\usepackage{fvextra}
\usepackage[round]{natbib}
\setcitestyle{authoryear,round,citesep={;},aysep={,},yysep={;}}
\usepackage{caption}
\usepackage{hyperref}
\definecolor{linkblue}{rgb}{0,0.08,0.45}
\hypersetup{
  colorlinks=true,
  linkcolor=linkblue,
  citecolor=linkblue,
  filecolor=linkblue,
  urlcolor=linkblue,
  pdftitle={Asymmetric Activations Keep Sinking Preactivations: A Mechanism of Plasticity Loss in MLPs and CNNs, with the kunekune Activation},
  pdfauthor={Issa Nakatsuka and Masayoshi Kubo}
}
\usepackage{newunicodechar}
\newunicodechar{−}{\ensuremath{-}}
\newunicodechar{×}{\ensuremath{\times}}
\newunicodechar{≥}{\ensuremath{\geq}}
\newunicodechar{≤}{\ensuremath{\leq}}
\newunicodechar{≈}{\ensuremath{\approx}}
\newunicodechar{≃}{\ensuremath{\simeq}}
\newunicodechar{→}{\ensuremath{\rightarrow}}
\newunicodechar{←}{\ensuremath{\leftarrow}}
\newunicodechar{±}{\ensuremath{\pm}}
\newunicodechar{′}{\ensuremath{'}}
\newunicodechar{∼}{\ensuremath{\sim}}
\newunicodechar{≠}{\ensuremath{\neq}}
\newunicodechar{·}{\ensuremath{\cdot}}
\newunicodechar{≻}{\ensuremath{\succ}}
\newunicodechar{⊥}{\ensuremath{\perp}}
\newunicodechar{√}{\ensuremath{\surd}}
\newunicodechar{∈}{\ensuremath{\in}}
\newunicodechar{²}{\ensuremath{^2}}
\newunicodechar{¹}{\ensuremath{^1}}
\newcolumntype{P}[1]{>{\raggedright\arraybackslash}p{\dimexpr#1*26/27\relax}}
\newunicodechar{⇒}{\ensuremath{\Rightarrow}}
\newunicodechar{…}{\ldots}
\newunicodechar{ᵢ}{\ensuremath{_i}}
\newunicodechar{ₓ}{\ensuremath{_x}}
\newunicodechar{‖}{\ensuremath{\|}}
\newunicodechar{⁻}{\ensuremath{^{-}}}
\newunicodechar{⁵}{\ensuremath{^{5}}}
\newunicodechar{⁴}{\ensuremath{^{4}}}
\newunicodechar{³}{\ensuremath{^{3}}}
\newunicodechar{α}{\ensuremath{\alpha}}
\newunicodechar{κ}{\ensuremath{\kappa}}
\newunicodechar{λ}{\ensuremath{\lambda}}
\newunicodechar{η}{\ensuremath{\eta}}
\newunicodechar{ρ}{\ensuremath{\rho}}
\newunicodechar{τ}{\ensuremath{\tau}}
\newunicodechar{σ}{\ensuremath{\sigma}}
\newunicodechar{ε}{\ensuremath{\varepsilon}}
\newunicodechar{φ}{\ensuremath{\varphi}}
\newunicodechar{Φ}{\ensuremath{\Phi}}
\newunicodechar{θ}{\ensuremath{\theta}}
\newunicodechar{π}{\ensuremath{\pi}}
\newunicodechar{÷}{\ensuremath{\div}}
\title{Asymmetric Activations Keep Sinking Preactivations: \\
    A Mechanism of Plasticity Loss in MLPs and CNNs, with the kunekune Activation}
\author{%
Issa Nakatsuka\textsuperscript{1}\qquad Masayoshi Kubo\textsuperscript{2}\\[2ex]
\normalsize\textsuperscript{1}School of Informatics and Mathematical Science, Faculty of Engineering, Kyoto University\\
\normalsize\textsuperscript{2}Graduate School of Informatics, Kyoto University\\[1.5ex]
\normalsize Correspondence: \href{mailto:nakatsuka.issa.23f@st.kyoto-u.ac.jp}{\texttt{nakatsuka.issa.23f@st.kyoto-u.ac.jp}} (Issa Nakatsuka)%
}
\date{}

\begin{document}

\maketitle

\begin{abstract}
Three phenomena known in continual learning (the negative shift of preactivations, the decline of activation derivatives, and the linearization of units) are connected here by a single mechanism. Each time a task is learned, the readout aligns with the current task, and at the next switch this alignment produces a gradient that lowers the mean of the preactivation. This gradient is carried by the paths of the input mean, the bias, and the upstream that moves with learning; with asymmetric activations the mean shifts deep into negative values, and growth of the preactivation scale increases the depth. Units that have shifted negative lose their activation derivative, and adaptation to new tasks slows. The first stage is shown by derivation (a lower bound on the alignment and an identity that decomposes the switch gradient) and observation; stages 2--4 are tested by interventions in MLPs and CNNs without normalization layers. On random-label sequences, blocking the paths preserved adaptation (MLP, 200 tasks; CNN, 30 tasks); capping the scale kept the ELU MLP on random labels from collapsing; and swapping only the activation derivative, with activations and predictions kept identical, changed adaptation in both directions. In the MLP with real labels, the network with the paths blocked scored lower than the unmodified network (post hoc). On the standard Continual ImageNet (500 tasks), at the learning rate at which the unmodified network loses plasticity (0.001), the path-blocking intervention (centering without blocking the bias worked equally well) and kunekune, an activation whose curvature scale is matched to the standard deviation of the preactivation, removed the loss, and the derivative swap moved adaptation in both directions. On Permuted MNIST, at the same learning rate, they reduced the loss by about 70--80\% but did not remove it. The seven existing methods compared by Kumar et al.\ were placed alongside, with learning rate and strength calibrated by the same rule. In a Transformer and a small language model that share the output layer or embeddings across all tasks, the interventions that worked acted on those shared parts (the boundary of the mechanism's scope).
\end{abstract}

\section{Introduction}
\begin{figure*}[t]
\centering
\includegraphics[width=0.94\textwidth]{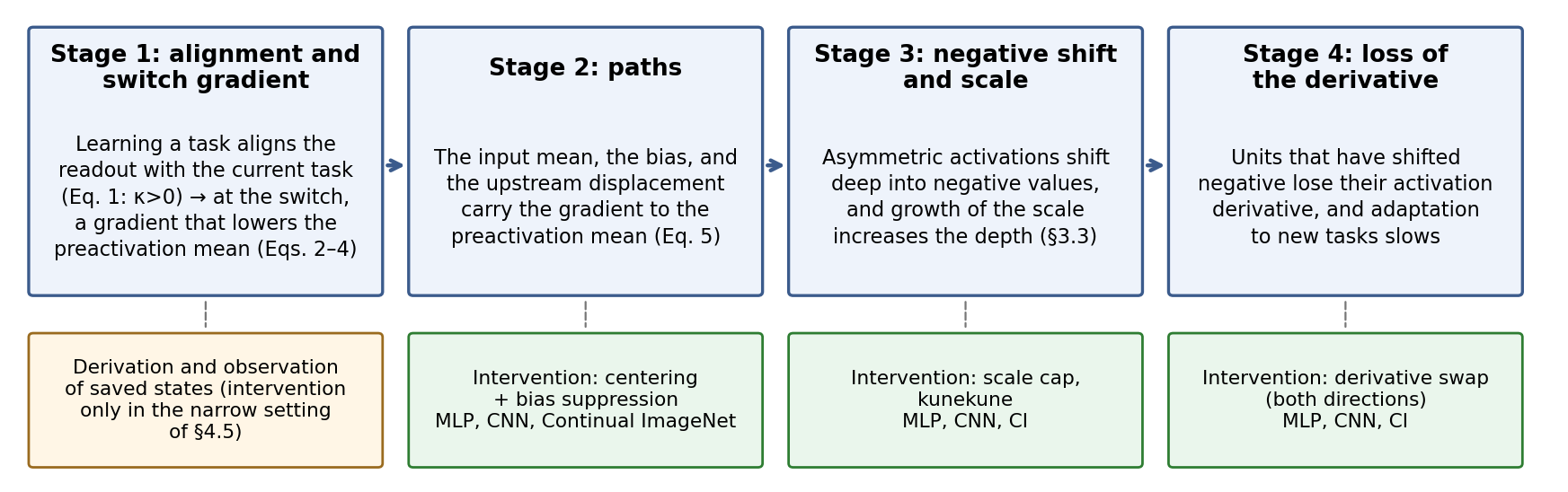}
\caption{The chain of four stages (top) and the intervention applied to each stage (bottom). CI = Continual ImageNet. Networks that share the output layer or embeddings across all tasks are outside the scope (\S\ref{sec:scope}).}
\label{fig:chain}
\end{figure*}
In the loss of plasticity, in which a network that keeps learning becomes unable to learn new tasks, units die, weights grow, and the rank of the representation falls \citep{dohare2024loss}. Another view holds that preactivations are pushed into the tails of the activation function, where gradients no longer pass \citep{lillo2025activation}. In ReLU networks at initialization, when the readout weights are random, gradient descent acts to shrink positive preactivations (the proof is limited to ReLU; for GELU and SiLU this is reported experimentally) \citep{li2023lazy,shvetsov2026bug}. We ask why that downward gradient is rebuilt at every task switch, which paths carry it to the mean of the preactivation, and what is lost so that adaptation to new tasks slows.

The answer is a mechanism with four stages (Figure~\ref{fig:chain}). (1)~Learning a task aligns the features and the readout with the labels of the current task, and at the next switch this alignment produces a downward gradient proportional to the product of the activation value and its derivative. (2)~This gradient is carried to the preactivation mean by the input-mean and bias paths, and, in layers whose upstream moves with learning, also by that upstream displacement. (3)~Asymmetric activations shift deep into negative values, and growth of the preactivation scale increases the depth. (4)~Units that have shifted negative lose their activation derivative, and adaptation to new tasks slows.

We establish a reference in MLPs (\S\ref{sec:mlp}, Figure~\ref{fig:mlp}) and test it in CNNs (\S\ref{sec:cnn}, Figure~\ref{fig:cnn}) and on the standard task sequences of Kumar et al.\ (\S\ref{sec:std}). Stages 2--4 are tested by interventions; stage 1 is shown by derivation and observation. Centering, bias suppression, the scale cap, and replacing the activation are interventions that each move one stage; we do not propose them as methods.

\textbf{Relation to prior work.} The downward gradient at initialization is described as not applying after training starts, or as decreasing once the readout aligns with the labels (\citealp[\S3.4]{li2023lazy}; \citealp[App.~A.4]{shvetsov2026bug}). That biases drift toward negative values under random labels so that ReLUs die \citep{cornacchia2021regularization}, and that the preactivations of many inputs drop right after a switch \citep{lyle2024disentangling}, have been described as observations, and loss of plasticity has also been reported to occur without saturated units \citep{lyle2023understanding}. We add that the learned readout rebuilds the downward gradient at every switch, the paths that carry it, and that what is lost is the activation derivative, which is tied to the growth of the scale; and we test these by interventions.

Centering activities and errors is a classical way to speed up learning \citep{schraudolph1998centering,raiko2012deep}, and normalization that subtracts the minibatch mean also centers the backpropagated gradient \citep[\S4]{salimans2016weight}. We use these as interventions that block the paths carrying the gradient, and ask which stage must be blocked for the loss to disappear.

\section{Setup and measurement}
\subsection{Terms and measures}
\begin{table}[t]
\caption{Terms used in this paper (a detailed version is in App.~B).}
\label{tab:terms}
\small
\begin{tabular}{@{}>{\raggedright\arraybackslash}p{0.23\columnwidth}>{\raggedright\arraybackslash}p{0.73\columnwidth}@{}}
\toprule
Term & Meaning \\
\midrule
online accuracy & Accuracy on each minibatch before the update, averaged within a task. Its average over all tasks is called the total average online accuracy, following Kumar et al. \\
gap to scratch & Online accuracy of a network trained from scratch on the same task with the same rule, minus that of the network that kept learning (points; positive means plasticity was lost) \\
preactivation mean $m_i$, scale $s_i$ & Mean and standard deviation of the preactivation of unit $i$ over the evaluation inputs. Standardized mean $r_i=m_i/s_i$; open fraction $p_i=\Pr_x[z_i>0]$ \\
activation derivative & $\varphi'(z_i(x))$, the factor by which backpropagated gradients pass through the unit. A unit with $\lvert\varphi'\rvert<10^{-6}$ on all evaluation inputs is called dead \\
\bottomrule
\end{tabular}
\end{table}

Tasks are learned in sequence, and the weights and the optimizer state carry over at switches. We measure adaptation to new tasks, not retention of past tasks. Let $x$ be the input to a layer and $\mu$ the mean of the evaluation input distribution, and split $z_i(x)=w_i^\top x+b_i=m_i+w_i^\top(x-\mu)$ ($m_i=w_i^\top\mu+b_i$). Because changes in $r$ also include changes in the denominator $s$, we state the dynamics in terms of $m$ and use $r$ only to describe states.

\subsection{Tasks and the roles of the architectures}
A random-label task assigns 10-class random labels, independent across tasks, to a fixed set of 1,200 MNIST or CIFAR-10 images (a test of relearning that requires no generalization; \citealp{lyle2022understanding}; \citealp{lee2024slow}). Real-label 5+1 CIFAR alternates a hard task with 5 classes and an easy task with 1 class from CIFAR-100 (evaluated on the hard tasks). Both are settings of \citet{kumar2024maintaining}; in \S\ref{sec:std} we use the remaining two of Kumar et al., Continual ImageNet (1,200 images from 2 ImageNet classes; 500 tasks; output layer shared across all tasks) and Permuted MNIST (10,000 images with permuted pixels; 500 tasks). The Continual ImageNet data are the $32\times32$ images released by \citet{dohare2024loss}, and our padding and input standardization differ from the description of Kumar et al.\ (App.~J.1). The main networks have no normalization layers and are trained with Adam. Inputs scaled as pixel value/255 are called raw, and inputs standardized per channel are called std (App.~A).

\section{Mechanism: the learned state moves the preactivation mean}
\subsection{Learning creates the switch gradient}
Learning before a switch aligns the features and the readout with the labels of the current task. Let $\phi_n$ be the features of input $n=1,\ldots,N$, $y_n$ the one-hot label of the previous task ($C$ classes), and $\ell_n=V\phi_n+b$ the logits, and let $V^c,b^c$ be the readout and bias centered along the class direction. We measure the strength of the alignment by the coefficient $\kappa$ of the projection of $V^c$ onto the direction of $F=\sum_n(y_n-\mathbf1/C)\phi_n^\top$. With $\mathcal L$ the mean cross-entropy, when $F\ne0$,
\begin{equation}
\kappa=\frac{\langle V^c,F\rangle_F}{\|F\|_F^2}\ge\frac{N(\log C-\mathcal L)-\Gamma}{\|F\|_F^2}
\label{eq:kappa}
\end{equation}
holds (proof in App.~C.1). $\Gamma=\sum_c(n_c-N/C)b^c_c$ is a term arising from the imbalance of the class counts $n_c$ and the bias, and is 0 when the classes are balanced. Hence $\kappa>0$ whenever $\mathcal L<\log C-\Gamma/N$; neither independence of features and labels nor perfect accuracy is required.

This alignment becomes a downward gradient at the switch. With the layer input fixed and the weights and bias updated by single-sample SGD with the same learning rate $\eta$, the preactivation mean moves by $\Delta m_i=-\eta\,\delta_{in}K_n$ ($K_n=\mu^\top x_n+1$; \S\ref{sec:paths}) for $\delta_{in}=\partial\ell_n/\partial z_i(x_n)$. Defining the switch gradient averaged over the new labels as
\begin{equation}
S_i=\sum_nK_n\varphi'(z_{in})\langle v_i,\pi_n-\mathbf1/C\rangle
\label{eq:S}
\end{equation}
($\pi_n$ is the prediction before the switch), $S_i>0$ is the lowering direction. In a one-hidden-layer model with fixed features, under assumption H1 (App.~C.2)---labels independent and uniform across inputs, a readout $v_{ic}=v^0_{ic}+\kappa\sum_m(y_{mc}-1/C)\varphi(z_{im})$, and the limit of confident correct predictions before the switch---averaging over the old labels gives
\begin{equation}
\mathbb E_{\rm old}S_i=\kappa(1-1/C)\sum_nK_n\varphi'(z_{in})\varphi(z_{in}).
\label{eq:Sold}
\end{equation}
In this model the direction of the switch gradient is set by the sign of the product of the activation value and its derivative, and its size is proportional to the alignment $\kappa$ created by learning.

With real labels, the new labels are not independent of the features, but the switch gradient splits identically as
\begin{equation}
S_i^{\rm real}=S_i^{\rm rand}-L_i,\qquad L_i=\operatorname{Cov}_{c\sim\bar y}\!\big(v_{i,c},\,\bar g_{i,c}\big)
\label{eq:split}
\end{equation}
(App.~C.7). $S_i^{\rm rand}$ is the term in which the labels of the new task are replaced by the class frequencies $\bar y$, and does not depend on the label assignment. $L_i$ is the covariance between the readout weight $v_{i,c}$ and the mean $\bar g_{i,c}$, over the images of class $c$, of the backpropagation factor $g_{in}=K_n\varphi'(z_{in})$; random labels are the limit in which it is exactly 0.

\subsection{The paths that carry the gradient to the preactivation mean}
\label{sec:paths}
The finite displacement from time $t$ to $t+1$ is exactly
\begin{equation}
\Delta m_i=(\Delta w_i)^\top\mu_t+w_{i,t+1}^\top\Delta\mu+\Delta b_i
\label{eq:paths}
\end{equation}
(independently of the optimizer). The first term is the weight path, the third the bias path, and the second the path of upstream change. $K_n=\mu^\top x_n+1$ is the sum of what the weight path ($\mu^\top x_n$) and the bias path ($+1$) carry in one SGD step with a fixed input; if every component of the input is nonnegative, as for raw images or ReLU outputs, then $K_n\ge1$, and a positive local gradient lowers the mean. With standardized inputs $K_n$ can also be negative. Exactly centering the input gives $\mu=0$ and $K_n=1$, which removes the weight path but leaves the bias path, so whether centering alone suppresses the negative shift also depends on how the bias acts. In layers whose upstream moves with learning, the second term is added (\S\ref{sec:cnn}).

\subsection{Sign boundaries and scale for each activation}
\label{sec:sign}
For ReLU, $\varphi'(z)\varphi(z)=\max(z,0)$, so if every $K_n$ is positive the sum in Eq.~\ref{eq:Sold} is positive and nothing lifts units back from the negative side (there is no zero in the live regime). Under $z\sim\mathcal N(m,s^2)$ and a constant positive $K$, the sign boundary of the sum for leaky ReLU (negative slope 0.1) lies at the standardized mean $r\simeq-1.72$ (open fraction about 4\%) independently of the scale, so the absolute position $m=rs$ is deeper for a larger scale. For ELU, the zero itself deepens as the scale grows (App.~C.3--C.6, including SiLU and GELU). The zero is a sign boundary, not a stopping point of the training trajectory, and we do not derive the growth of the scale itself. Conversely, for kunekune, an activation whose two tails have slope 1 and whose curvature scale follows the preactivation scale (it keeps the curvature of Snake \citep{ziyin2020neural} on a finite interval and joins it outside to straight lines of slope 1; App.~F.1), the zero stays at $r\simeq-1.0$ independently of the scale, and the mean derivative there is 0.55 (numerical; App.~C.6). We predict that as the scale grows the mean does not deepen and the derivative remains (unlike an activation that maintains plasticity with sines and cosines \citep{lewandowski2024plastic}, we use kunekune as an intervention that moves only stage 3). After the negative shift, ReLU's derivative becomes 0 on the inputs for which it is closed, ELU's becomes small on the deep side, and leaky ReLU linearizes when most inputs fall on the same branch. All of these occur separately from death, in which a unit is closed on all inputs (\S\ref{sec:swap}).

\section{MLP: testing the four stages in order}
\label{sec:mlp}
\begin{figure*}[t]
\centering
\includegraphics[width=\textwidth]{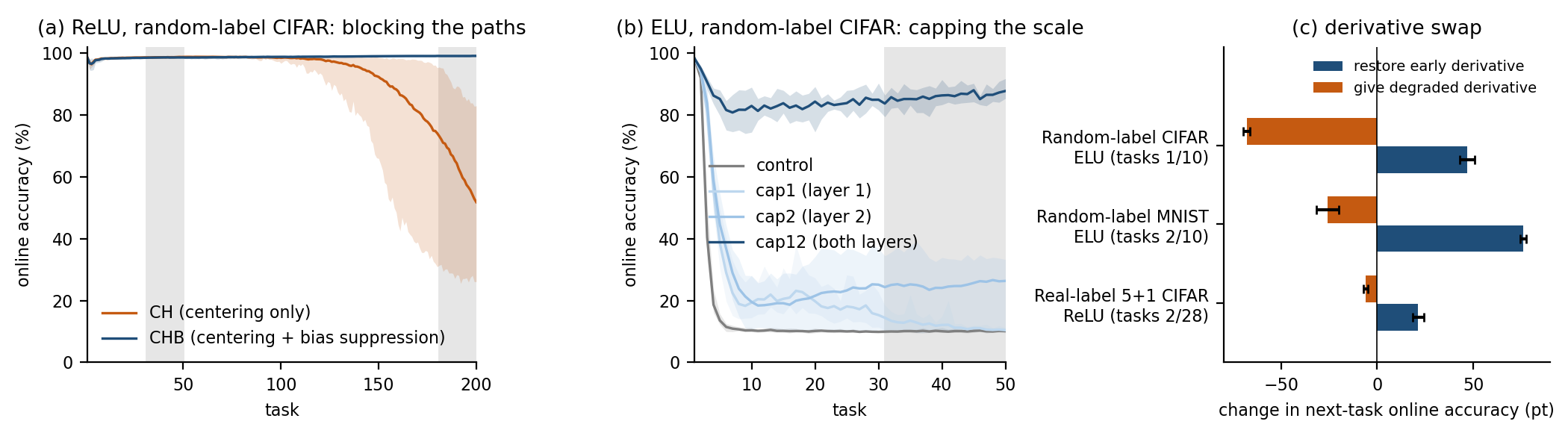}
\caption{MLP. (a) Online accuracy per task of CH (centering only) and CHB (centering + bias suppression) on random-label CIFAR, ReLU, raw (200 tasks; mean of seeds 10--19; band = seed minimum to maximum; gray bands = tasks 31--50 and 181--200; App.~D.5). (b) Control and scale caps cap1, cap2, cap12 on random-label CIFAR, ELU, std (50 tasks; mean and minimum to maximum of seeds 0--9; App.~D.3). (c) Change in next-task online accuracy under the derivative swap in three settings (mean of the paired differences over 10 seeds and 97.5\% intervals; App.~E.1).}
\label{fig:mlp}
\end{figure*}
\subsection{The switch gradient is created by learning and rebuilt at each task}
In saved states of a one-hidden-layer MNIST network with five activations (2 seeds each), the expected switch gradient rose at a nearly constant multiple of the amount $\log C-\mathcal L$ by which the task-1 loss fell from chance level, and $\kappa$ was rebuilt at the end of every task. In the first 200 updates after a switch (Adam), 82--100\% of live units actually shifted negative. The sum in Eq.~\ref{eq:Sold} agreed in direction with the expected gradient for 90--100\% of units for all activations except leaky ReLU (40--52\% from task 4 on for leaky ReLU), but carried only 13--60\% of its magnitude (App.~D.1). The theory predicts the direction, not the magnitude (in actual training, most of the alignment is carried by changes in the features, and that part of the gradient falls into terms outside H1).

\subsection{Blocking the paths: centering and bias suppression}
\label{sec:block}
In a two-hidden-layer network on random-label CIFAR (raw, ReLU), we compared subtracting the input mean (C), subtracting the means of the outputs of both hidden layers (H), and both (CH) (these are names of operations, distinct from the number of classes $C$). With C alone the second layer was lost, and with H alone the first layer: the layer whose input-mean path remained open was lost. Online accuracy over tasks 31--50 was 11.33\% for C, 98.67\% for CH, and 98.54\% for centering + bias suppression (CHB), which adds to CH the zeroing of the first-layer bias and decoupled weight decay on the second-layer bias (medians over 10 seeds). Extended to 200 tasks with 10 other seeds, the mean over tasks 181--200 was 62.98\% for CH (falling in all 10 seeds) and 99.03\% for CHB (Figure~\ref{fig:mlp}a; App.~D.5). In the first layer with centered input, the preactivation mean equals the bias ($m_1=b_1$), and under CH this shifted negative. Blocking the weight path by centering leaves the gradient in the bias path, and only when that path was also blocked was adaptation preserved over the long sequence. This, however, is a result on random-label sequences (on Continual ImageNet, centering alone sufficed; \S\ref{sec:lr001}, \S\ref{sec:scope}). On real-label 5+1, the network with both paths blocked (output-bias learning rate ×100) was 6--12 points below the unmodified network on every hard task from task 3 to 29, and at task 29 it trailed a network trained from scratch by 6--17 points in all 10 seeds (post hoc, descriptive; App.~G).

\subsection{Stopping the scale; an activation that follows the scale}
To test whether the difference between ELU, whose zero deepens with scale, and ReLU, which has no zero in the live regime, appears in adaptation, we stopped the growth of the scale. The row norm of each unit's input weights at the end of task 1 is used as a cap, and after every Adam update from task 2 on, rows that exceed it are projected back (both layers: cap12; one layer: cap1, cap2; App.~A.5). On random-label CIFAR (std, ELU), online accuracy over tasks 31--50 was 86.0\% with cap12 against 10.1\% for the control, preventing the collapse in all 10 seeds; a cap on only one layer was not enough (Figure~\ref{fig:mlp}b; App.~D.3).

For ReLU on the same random-label CIFAR, the control was at the floor (11.3\%) and cap12 did not change it. For ReLU on real-label 5+1 CIFAR, cap12 lowered late online accuracy by 3.6 points [97.5\% interval −6.7 to −0.5; the simultaneous interval for the 8 comparisons of D.3 contains 0], and did not lower the fraction of dead units (App.~D.2, D.3). Up to this point the results agree with the reading of \S\ref{sec:sign}. What does not agree is ELU on real-label 5+1, where cap12 made accuracy 2.0 points worse. The preregistered prediction that ``the effect splits by activation alone'' failed here, and the failure cannot be explained by the cost of the cap on networks trained from scratch (App.~D.3).

Moving the same stage from the side of the activation: in the random-label CIFAR MLP, late online accuracy of ELU was at the floor of about 10\% for both raw and std, whereas kunekune reached 95.7\% and 98.7\% (medians of 10 seeds; App.~F.2). On real-label 5+1, kunekune and adaptive Snake, whose curvature scales are matched to the preactivation, were 4.6--6.1 points above Snake with the curvature scale fixed at its initial value in late online accuracy (10/10; preregistered; App.~F.1), and when each condition selected its learning rate from \{0.0001, 0.0003, 0.001\} by the same rule, the median difference was 11.4--11.5 points (20/20; App.~F.3).

\subsection{Swapping the derivative}
\label{sec:swap}
To test whether the loss of the derivative cuts adaptation, we swap only the derivative of the second layer. We fix the difference $o_i=z_i^{\rm src}-z_i^0$ between the preactivation $z_i^0$ at the branch point and the donor's preactivation $z_i^{\rm src}$, and after the branch the output is $\varphi(z_i^0)+\varphi(z_i+o_i)-\varphi(z_i^0+o_i)$. At the branch point the activities, logits, loss, and weights coincide with those of the control; only the derivative used for learning becomes $\varphi'(z_i^{\rm src})$, which thereafter moves with the preactivation.

Swapping the derivative in both directions between early and post-degradation points of the same trajectory, in all three settings (random-label CIFAR, ELU/std; random-label MNIST, ELU/raw; real-label 5+1 CIFAR, ReLU/std), restoring the early derivative raised online accuracy on the next task (+21.3 to +76.0 points) and giving the degraded derivative lowered it (−67.9 to −6.0 points; all 10 seeds in the same direction in each comparison; Figure~\ref{fig:mlp}c; 97.5\% intervals in the table of App.~E.1). Swapping the derivative between a scale-capped network and a control network also moved adaptation in both directions (random-label MNIST, ELU; this is the effect of an artificial intervention, not the share of the cap's benefit that it mediates; App.~E.2). The effect of restoring the derivative could not be reproduced by fixed learning rates matched to the ratio of gradient norms or of Adam displacements; both controls fell below the restored network in all 6 comparisons (App.~E.3).

\subsection{An intervention on stage 1: a necessity of limited scope}
\label{sec:stage1}
In the real-label 5+1 MLP with the hidden layers centered and the biases fixed at 0, the temporary delay from task 2 to 3 was prevented by removing, from the error backpropagated to the hidden layers, only the part equal to the minibatch mean of the output error (error centering \citep[\S2.2]{schraudolph1998centering} applied to backpropagation into the hidden layers) (10/10; criterion fixed before the run; App.~G). In the unmodified network and on the two standard sequences, however, the same removal was harmful (on Continual ImageNet, −20.9 points from task 2, 0/10, among others; App.~G, J.10). What the stage-1 intervention supports reaches only the temporary delay at the first switch.

\section{CNN: the same four stages in layers whose upstream moves}
\label{sec:cnn}
\begin{figure*}[t]
\centering
\includegraphics[width=\textwidth]{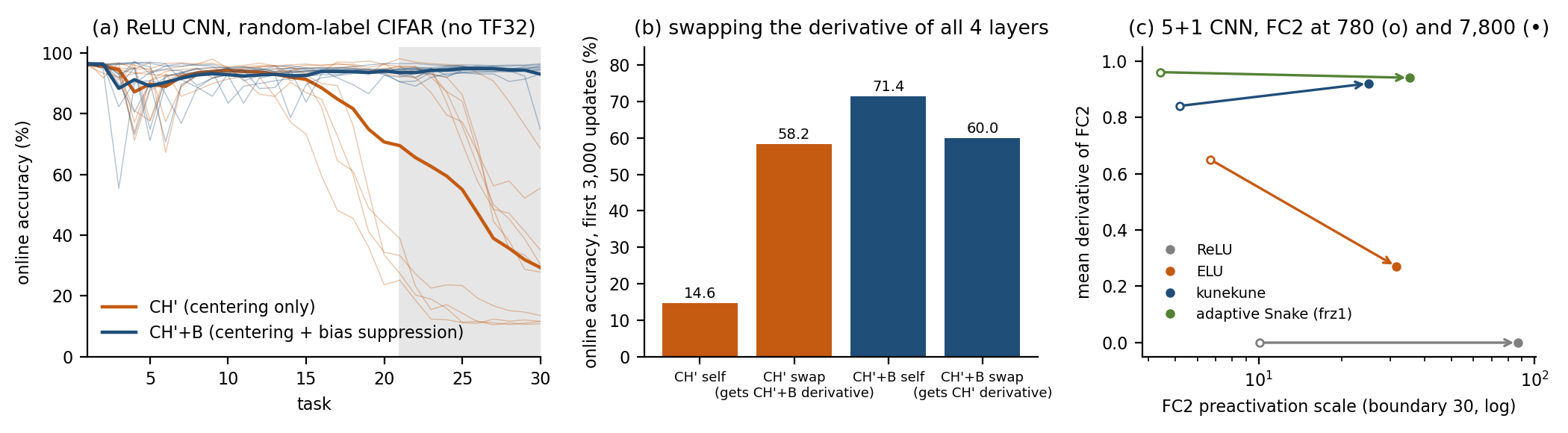}
\caption{CNN. (a) Online accuracy per task of CH′ and CH′+B in the ReLU CNN on random-label CIFAR (run without TF32; thin lines = each of seeds 10--19; thick lines = mean; gray band = tasks 21--30; App.~H.3). (b) Online accuracy over the first 3,000 updates of the next task after swapping the derivatives of all 4 hidden layers from the end of task 30 (mean of 10 seeds; App.~H.4). (c) FC2 scale at boundary 30 (log axis) and mean derivative when the updates per task of the 5+1 CNN are extended from 780 (open) to 7,800 (filled) (App.~H.5).}
\label{fig:cnn}
\end{figure*}
A network with 2 convolutional layers and 2 fully connected hidden layers was trained on random-label CIFAR (30 tasks) and real-label 5+1 CIFAR (10 seeds each; App.~A.1), and we tested the four stages in the same order as in \S\ref{sec:mlp} (Figure~\ref{fig:cnn}; App.~H).

\textbf{Alignment.} In saved states of the leaky ReLU network, the readout of the last hidden layer f2 realigned with the current task at every boundary observed (ratio of $\kappa$ to that at the end of task 1: 1.46--7.05, 10/10), with 0 violations of the lower bound of Eq.~\ref{eq:kappa}. In 200 replayed updates, 84--100\% of f2 shifted negative, but 97--100\% of the first-order displacement of the mean was carried by the upstream; f2 recovered within the task, and the negative shift accumulated in the layers below it, f1, c2, and c1 (App.~H.1). On real-label 5+1 CIFAR too, the readout was aligned in all 467 states at the ends of hard tasks; split by Eq.~\ref{eq:split}, the direction for FC2 agreed with $S^{\rm rand}$ for 95--98\% of units, and the label term was 7--12\% of $S^{\rm rand}$ (App.~H.2). In the three conditions other than ReLU, however, 49--55\% of units were on the lowering side, and the negative shift of the real-label CNN proceeds within tasks.

\textbf{Paths.} Centering the input and all features (channel × position) of the ReLU network by an exponential moving average (CH′) closed the input-mean path, but the gradient moved to the bias path. In the first run, which used TF32, the biases of all 4 layers of CH′ shifted negative and the units closed (fraction of dead units at task 30: 0.77--0.88), and online accuracy over tasks 21--30 had a median of 17.8\%. The comparison that adds bias suppression (B), which projects the biases of the 4 hidden layers after every Adam update to within ±0.2 times the standard deviation of the preactivation of the current minibatch, was made in a separate run without TF32: CH′ averaged 49.5\% (4/10 seeds collapsed) and CH′+B 94.2\% (improved in 10/10; Figure~\ref{fig:cnn}a; App.~H.3). CH′ moved substantially with the numerical-precision setting, but only the network with both paths blocked exceeded 50\% in all 10 seeds (CH′: 1/10 and 6/10).

\textbf{Scale.} Extending the updates per task on real-label 5+1 CIFAR from 780 to 7,800 grew the scale of FC2 under all of ReLU, ELU, and two kinds of Snake (Figure~\ref{fig:cnn}c). Only ELU shifted negative and lost its derivative together with the scale (mean derivative 0.65→0.27), and online accuracy over the first 780 updates of late hard tasks fell by 1.17 points. Online accuracy over the whole task was nearly the same as kunekune's (98.2\% vs.\ 98.3\%), so the delay is confined to the start. ReLU already had 85\% of FC2 dead at 780 updates, a loss separate from the scale. Extending the budget is not an intervention that moves only the scale. kunekune kept a mean derivative of 0.9 as the scale grew, and its online accuracy over the first 780 updates rose by 4.6 points (preregistered criterion; App.~H.5). In level, ELU at 780 updates ($\alpha=3.6$, outside the sign condition of \S\ref{sec:sign}) ranked highest; the intervention asks whether loss occurs and does not compare levels (App.~F.4).

In this setting, ELU has no loss that the scale cap can rescue. Capping the FC2 scale from 34 to 1.4 and keeping the mean derivative at 1.00 lowered online accuracy over the first 780 updates by 5.1 points (10/10), while end-of-task test accuracy was higher by 0.2 points on average (descriptive; App.~H.5). This ELU does not decline late as tasks accumulate; extending the budget delays only the first 780 updates.

\textbf{Derivative swap.} When the derivatives of all 4 hidden layers were swapped between CH′ and CH′+B from the end of task 30, online accuracy over the first 3,000 updates of the next task rose from 14.6\% to 58.2\% for the CH′ recipient (10/10) and fell from 71.4\% to 60.0\% for the CH′+B recipient (9/10; Figure~\ref{fig:cnn}b). The improvement on the rescuing side persisted to the end of the task; on the opposite side the low derivative reopened within 100 updates, and the deterioration did not persist to the end. Because the cross-entropy at random-label switches reflects confidence on the previous task, adaptation was compared in online accuracy. The cross-entropy of the CH′+B network, however, was as large as 169 even over the last 3,000 updates, where accuracy was 98.6\% (App.~H.4), so adaptation can be said to be preserved only in terms of online accuracy. Adding the same B after degradation opened the derivative, but the improvement in accuracy was not certain (+2.9 points, 7/10), and the cross-entropy worsened in 10/10. Blocking is prevention, not cure.

\section{Standard task sequences: Continual ImageNet and Permuted MNIST}
\label{sec:std}
\subsection{Comparison protocol (fixed before the runs)}
\label{sec:protocol}
Our interventions (centering + bias suppression, kunekune, and the scale cap [Continual ImageNet only]), the 7 methods compared by Kumar et al.\ (L2 Init; L2; Shrink \& Perturb [S\&P] \citep{ash2020warm}; Continual Backprop [CBP] \citep{dohare2024loss}; ReDo \citep{sokar2023dormant}; Layer Norm \citep{ba2016layer}; Concat ReLU \citep{shang2016understanding}), and the unmodified network were run for 500 tasks with 10 seeds. The learning rate was selected from 4 values, the protocol of Kumar et al., \{0.001, 0.0001\} \citep[\S5]{kumar2024maintaining}, plus \{0.003, 0.01\}, and the strength from the grid of each method, by the same rule (largest total average online accuracy on calibration seeds 100--102). The primary measure is the gap to scratch (mean of tasks 400 and 500), and the secondary is the total average online accuracy. The loss criterion (a loss is present if the mean gap is at least 0.5 points and at least 9/10 seeds are positive), the pairwise criterion (intervention better = better in 10/10 and by at least 0.5 points on average; the other better = worse in at least 8/10 and by at least 0.5 points on average; otherwise the difference is undecided), and the wording rules were fixed before the runs (App.~J.2).

\subsection{Loss and the effect of interventions at learning rate 0.001}
\label{sec:lr001}
\textbf{Continual ImageNet.} The unmodified network at learning rate 0.001 fell from 81.8\% in the early window (tasks 11--60) to 54.9\% in the late window (tasks 451--500), trailed a network trained from scratch by 28.2 points (10/10), and had 90.0\% of the units of the last hidden layer dead at task 500. The network at learning rate 0.0001 does not lose plasticity (−4.1 points), but its total average online accuracy stays at 75.0\% \citep[\S5.1]{kumar2024maintaining}. Keeping learning rate 0.001, centering + bias suppression and kunekune removed the loss of plasticity (gap to scratch −7.8 and −2.1 points), preserved 36.0 and 30.3 points more plasticity than the unmodified network (10/10 each), and reduced the dead units of the last hidden layer from 90.0\% to 0.0\% (kunekune cannot die by construction). Centering without bias suppression worked equally well (App.~J.3). The scale cap narrowed the gap by 21.4 points, but the loss remained (+6.8 points).

When the derivatives of all 4 hidden layers were swapped, restoring the derivative of the early network (end of task 10) to the post-degradation network (end of task 490) changed online accuracy on the next task by +15.7 points (10/10), and giving it in the other direction by −21.2 points (0/10); the second task went in the same direction. The restored effect corresponds to about 55\% of the difference between the early window and the control's online accuracy on task 491 (App.~J.9). Stage 1 remains an observation. When the switch gradient is split by Eq.~\ref{eq:split}, in early tasks (2--11) $S^{\rm rand}$ matched the direction in a fraction 0.88, and the median ratio of the label term was 0.48, larger on this sequence, whose output is shared across tasks, than in the real-label 5+1 CNN (0.07--0.12). Late (tasks 491--500), the live units fell to 9.6\% and the directional agreement dropped to 0.21 (App.~J.8).

The fraction of dead units did not correspond one-to-one to the loss. The unmodified network at learning rate 0.0001 did not lose plasticity with 48.6\% of the last hidden layer dead, and the scale cap reduced death in the last hidden layer from 90.0\% to 71.3\% but increased death in the lower layers, and the loss remained (App.~J.12).

\textbf{Permuted MNIST.} The unmodified network at learning rate 0.001 trailed a network trained from scratch more as tasks accumulated (+8.1 points; 10/10; App.~J.3). At the same learning rate, centering + bias suppression and kunekune reduced the loss by fractions of 0.79 and 0.75, but gaps of +1.7 and +2.0 points remained (10/10). The row-norm cap on the weights (cap12) removed the loss (gap −3.4 points), but at task 500, 91.3\% of the units of the first hidden layer were dead. Giving the degraded derivative changed the next task by −12.5 points (0/10), whereas the restoring direction gave −3.4 points (1/10) on the next task (App.~J.9, J.12).

\subsection{Comparison with calibrated existing methods}
\label{sec:calib}
\textbf{Continual ImageNet.} In the comparison with learning rate and strength selected by the rule of \S\ref{sec:protocol}, centering + bias suppression (learning rate 0.001, current strength) had an undecided difference from L2 Init and L2 in preserving plasticity (seeds above them: 5/10 and 8/10; 95\% intervals of the difference [points]: [−0.4, +0.5] and [+0.1, +1.5]) and was better than S\&P, CBP, ReDo, Layer Norm, and Concat ReLU; in total average online accuracy it had an undecided difference from L2 Init, L2, and Concat ReLU and was higher than S\&P, CBP, ReDo, and Layer Norm (App.~J.3). kunekune was worse than L2 Init, L2, ReDo, and Concat ReLU in preserving plasticity (ibid.). Of the difference from L2 Init in total average online accuracy (+2.4 points, 9/10), 66\% (82\% when applied to tasks 2--500) corresponds to the difference in learning speed between the networks trained from scratch (+2.0 points, 10/10) (post hoc; App.~J.11). The total average online accuracy of centering + bias suppression was 66.7\% at learning rate 0.003 and 92.3\% at 0.001 (calibration seeds; App.~J.6).

\textbf{Permuted MNIST.} Under the same reselection, centering + bias suppression (learning rate 0.003; second-layer bias decay at half its current value) was outperformed by L2 Init, L2, S\&P, CBP, ReDo, and Concat ReLU in preserving plasticity, and by L2 Init, L2, S\&P, CBP, and ReDo in total average online accuracy. In the reselected settings, centering + bias suppression reduced the loss by about 60\% and kunekune by about 70\% (relative to the calibrated unmodified network). Changes of setting and selections at the edge of a grid (5 on Continual ImageNet, 3 on Permuted MNIST) are given in App.~J.3. Layer Norm retained a loss on both sequences even at its selected learning rate (+17.1 and +10.5 points; 10/10 each); we have not examined how this differs from the report that layer normalization maintained plasticity \citep{lyle2023understanding}.

\section{The boundary of the mechanism's scope}
\label{sec:scope}
In networks with normalization layers, the results split (learning rate 0.001; 10 seeds; App.~J.14). The Layer Norm network lost plasticity on Continual ImageNet (gap +17.8 points; 10/10); with centering alone it had no loss (+1.3; positive seeds 7/10; the expectation stated before the run failed), and neither did the network in which all paths were blocked by also fixing the bias and the mean offset after normalization (shift) at 0 (+0.7; difference in preservation +17.1 points; 10/10; at learning rate 0.0001 the loss decreased but +4.0 remained). The batch-normalization (BN) network lost plasticity on both sequences (+11.0 and +9.3 points; 10/10 each), but fixing the shift at 0 increased the loss (+12.9 and +12.2), and the loss of the Layer Norm network on Permuted MNIST decreased with centering but remained (+8.8) and did not change when everything was blocked (+10.5 → +10.7). The losses of these networks cannot be explained by our paths. In networks that share the output layer or embeddings across all tasks, the place where interventions work changed (App.~K). An image Transformer with 2 blocks and no layer normalization (5 seeds) lost adaptation, but the stage criteria for the input, the upstream, and the scale did not hold (1/5, 0/5, 0/5), and 4 of the 8 predictions made before the run failed. The delay at task 30 disappeared when only the output layer was reset to 0 (post hoc), but the loss from task 60 on was not reversed in all 5 seeds by any intervention on the side of the output layer and the last FFN. In a pre-LN character Transformer (10 seeds), the four stages held. In a small 4-layer language model with tied input and output embeddings, the interventions that worked acted on the embeddings, and when moved to the body of Pythia-14m \citep{biderman2023pythia} the loss itself did not appear. In a reduced ResNet-18 with BN (Split CIFAR-100), output-layer-side methods and L2 Init raised late online accuracy (App.~J.4).

\section{Discussion and limitations}
\textbf{Common connection.} A single mechanism held in small MLPs and CNNs without normalization layers (stage 1 by derivation and observation). We have not, however, tested the four stages linked in order within a single setting. There are also settings in which keeping the scale and the derivative even slows the start (the scale cap on ELU in the real-label CNN). What changed with the architecture was the identity of what carries the mean: it moved from the fixed input mean to the upstream that moves with learning.

\textbf{Scope.} We do not claim a monotone negative shift of all layers or the irreversibility of the loss. Centering + bias suppression removes the loss on the random-label sequences and on Continual ImageNet; on Permuted MNIST, about 20\% remains at learning rate 0.001 and about 40\% in the reselected setting, and on real-label 5+1 the blocked network is worse than the unmodified network. The need to block the bias path was shown only on random-label sequences. The preservation in the random-label CNN holds only in terms of online accuracy (\S\ref{sec:cnn}). The losses of the BN network and of the Layer Norm network on Permuted MNIST did not disappear when the paths were blocked, and cannot be explained by our paths (App.~J.14). The comparisons use only Adam and were not tested with SGD \citep[App.~A.2.1]{kumar2024maintaining}. Only one placement of Layer Norm (after the linear map, before the activation) was tried. In the calibrated comparison on Permuted MNIST, 6 existing methods in preserving plasticity and 5 in total average online accuracy outperformed centering + bias suppression (\S\ref{sec:calib}). The replication control, in which the same setting was re-run in a different process, did not agree on Continual ImageNet (deviation up to 0.9 points; App.~J.5), and per the rules we raised the bar for comparisons across runs. Most interventions were preregistered runs (App.~A.4), and the CNN alignment is an analysis of saved states. We do not report mediation ratios that divide the benefit of natural interventions by the derivative or the scale alone.

\section*{AI Use Statement}
The authors formulated the research questions, hypotheses and claims. Generative AI assisted with experimental implementation, debugging, analysis, figures, reports, drafting, translation and editing. The authors verified the reported results, read and revised the manuscript, and take responsibility for the text, claims, code and experiments.

\clearpage
\appendix
\section*{Overview of the appendices}
\addcontentsline{toc}{section}{Overview}
{\footnotesize\setlength{\tabcolsep}{3pt}
\begin{longtable}{@{}P{3.0cm}P{11.3cm}P{2.3cm}@{}}
\toprule
\textbf{Appendix} & \textbf{Contents} & \textbf{Main text} \\
\midrule
A Setup & Tasks, architectures and budgets; online accuracy and networks trained from scratch; measurement of the preactivation mean, scale and death; aggregation (time of registration = A.4); definitions of centering, bias suppression and the scale cap (the setup of the standard sequences is in J.1) & §2, §4, §8 \\
B Terms and symbols & Detailed version of Table~1; identity B.1 for the term removed by error-mean removal; families of intervals and tests & §2.1 \\
C Theory and proofs & C.1 lower bound on the alignment (Eq.~1); C.2 H1 and Eq.~3; C.3 sign boundary of leaky ReLU; C.4 inequality for SiLU/GELU; C.5 model after the sum weakens (H2--H4); C.6 scale and sign boundary; C.7 label term of the switch gradient (Eq.~4; measured on a sequence whose output is shared across tasks in J.8) & §3 \\
D MLP checks & D.1 alignment and Eq.~3 (quartiles of the ratio 0.18--0.22, etc.); D.2--D.3 families of intervals of the scale cap and its cost on networks trained from scratch (norm of the second-layer input mean 52.6 → 1157 vs 40 = D.3); D.4 checks after the sum weakens; D.5 centering over 50 and 200 tasks & §4.1--§4.3 \\
E Derivative swap & E.1 two-way swaps in three settings (numbers and 97.5\% intervals of §4.4); E.2 swaps between the scale-capped (cap12′) and natural networks (80.8\% → 19.7\%, 16.9\% → 84.6\%; the offset is not fixed but follows the donor's continuation); E.3 learning-rate controls calibrated to gradient norms and Adam displacements & §4.4 \\
F Activations and existing methods & F.1 definitions of kunekune, adaptive Snake and fixed scales; F.2 all random-label conditions; F.3 the 19 conditions of the real-label MLP and existing methods, with the comparison in which the learning rate is reselected by the same rule (11.4--11.5 points of §4.3, 20/20); F.4 the 12 conditions of the real-label CNN & §3.3, §4.3, §5 \\
G Error-mean removal & Interventions on the temporary delay of the centered network (all conditions); 30 tasks of the unmodified network (including aligning the output bias at boundaries); comparison of the network with both paths blocked and the unmodified network & §4.2, §4.5 \\
H CNN & H.1 alignment and layer-wise gradients of the random-label CNN; H.2 alignment and label term of the real-label CNN; H.3 centering and bias suppression of the CNN; H.4 swap of the derivatives of all layers, and bias suppression added later; H.5 budget extension and scale cap & §5 \\
J Standard task sequences & Setup; rules fixed before the runs; tables of all arms and pairs (4 learning rates and reselected strengths); Split CIFAR-100; replication controls and post hoc checks; learning-rate and strength tables; all points of the calibration grids; split of the switch gradient; derivative swap; error-mean removal; speed of networks trained from scratch (J.11); fraction of dead units (J.12); per-task online accuracy (J.13); networks with normalization layers (J.14) & §6, §4.5 \\
K Boundary of the scope & K.0 sources and limits of the sentences of §7; K.1 verdicts for the image Transformer; K.2 character Transformer; K.3 all conditions of the language models; K.4 where the common-error gradient lands (analysis of saved weights) & §7, §8 \\
\bottomrule
\end{longtable}}

Log excerpts (verbatim environments) are copied mechanically from the analysis and verdict outputs; internal run identifiers are replaced by run A/B/C, free-text labels are rewritten in plain words, and the numbers are unchanged.

Condition symbols used in the tables of Appendices A, C--F and H: kunekune = the activation of the main text (F.1; symbol KKT1 in the tables); SNA = adaptive Snake; frz1 = the version fixed at the scale reached at the end of task 1; SN06, SN3 = Snake with $\alpha$ fixed at 0.6 and 3; KKA, KKA23 = the $C^2$ version of F.1 and its 2/3 multiple; NEO = loss-gated Snake (F.1); RSL = the activation of the setting of \citet{lillo2025activation} (F.4); SL = Smooth-Leaky; LR, LK03, LK001 = leaky ReLU with negative slope 0.1, 0.3, 0.01 (LR\_G and LR\_U in E.3 are learning-rate controls); R = ReLU (F.3, F.4; R in E.3 is the network with the derivative restored); L2I = L2 Init. Control (symbols ref and natural in the tables) = the unmodified network. RC = random-label CIFAR; RM = random-label MNIST; 5+1 = real-label 5+1 CIFAR (A.1). The families of intervals and tests are listed at the end of Appendix~B.

\section{Setup}
\label{app:A}
Main text: §2, §4.2--4.3.

\subsection{Tasks, architectures and training budgets}
With random labels, each seed fixes 1,200 images, and each task assigns independent uniform labels over 10 classes. 5+1 uses the fine labels of CIFAR-100 and alternates hard tasks with 500 images of each of 5 classes and easy tasks with 500 images of 1 class; 30 tasks use 90 classes, and no class is revisited. raw is pixel/255; std is standardization by per-channel constants. Weights and Adam states carry over at task switches. Only the branch interventions follow the rules of Appendices E and H.

\begin{center}\footnotesize\setlength{\tabcolsep}{3pt}
\begin{tabular}{@{}P{2.4cm}P{3.0cm}P{2.5cm}P{1.8cm}P{3.0cm}P{3.0cm}@{}}
\toprule
\textbf{Setting} & \textbf{Architecture / input} & \textbf{Tasks × updates / minibatch} & \textbf{Learning rate} & \textbf{seed} & \textbf{Main window} \\
\midrule
1-hidden-layer MNIST & 784–100–10 / raw & 30×4,000 / 16 & 1e−3 & 0, 1; 5 activations & saved task ends and 200 updates after a switch \\
RC: random-label CIFAR & 3072–100–100–10 / raw, std & 50×30,000 / 16 & 1e−3 & 0–9 & tasks 31–50 \\
RC long-run centering & same /raw, ReLU & 200×30,000 / 16 & 1e−3 & confirmation run 10–19 & tasks 181–200 \\
RM: random-label MNIST & 784–100–100–10 / raw, ELU & 100×6,000 / 16 & 1e−3 & 0–9 & tasks 51–100 \\
5+1 MLP & 3072–100–100–100 / std & 30×780 / 32 & 1e−4 fixed & main 0–9; follow-up of methods 0–19 & hard21, 23, 25, 27, 29 \\
RC CNN & 4 hidden layers below, 10 outputs & 30×30,000 / 16 & 1e−3 & 10–19 & tasks 21–30 \\
5+1 CNN & same hidden part, 100 outputs / std & 30×780, 2,340, 7,800 / 32 & calibrated per condition, F.4 & calibration 100–102, evaluation 2000–2009 & late hard tasks; common first 780 / whole task kept separate \\
\bottomrule
\end{tabular}\end{center}

The CNN is conv 3→16–$\varphi$–pool–conv 16→16–$\varphi$–pool–FC 1024→100–$\varphi$–FC 100→100–$\varphi$ (layers c1/c2/f1/f2; 16×16 and 8×8 after pooling; a bias in every layer). Adam $\beta=(.9,.999)$, $\epsilon=10^{-8}$, no weight decay. Runs with TF32 (true) and without TF32 (false) are separate comparisons.

Note: the seeds in a row are independent replication units; 2 seeds × 5 activations are not counted as 10 seeds.

\subsection{Online accuracy, comparison with networks trained from scratch, and evaluation times}
online is the accuracy on the minibatch before the update; train/test at the end of a task are separate measures taken after learning. fresh (a network trained from scratch) is a control that learns the same task, minibatch sequence and budget from initialized weights and a zero Adam state.

\begin{center}\footnotesize\setlength{\tabcolsep}{3pt}
\begin{tabular}{@{}P{3.5cm}P{6.0cm}P{6.5cm}@{}}
\toprule
\textbf{Comparison} & \textbf{Task and budget trained from scratch} & \textbf{Sign of the difference, aggregation} \\
\midrule
5+1 MLP, 19 conditions & last hard task 29, 780 updates & seed median of fresh − continued; positive means the continued network is worse \\
5+1 MLP, cost of the cap on fresh networks & hard 29, 780 updates, networks trained from scratch with / without the cap & interaction of the continued difference and the from-scratch difference, D.3 \\
5+1 CNN, 12 conditions & task 29, 780 updates & fresh − continued; positive means the continued network is worse; seed mean \\
CNN budget extension & tasks 9, 19, 29, with 780 / 2340 / 7800 updates each & task 29 is primary; fresh − continued; common first 780 and full budget kept separate \\
\bottomrule
\end{tabular}\end{center}

Note: the sign of the gap to scratch (fresh gap) differed between the original experiments and is therefore stated in the table. All comparisons use 10-seed controls; only the follow-up of the methods in the MLP uses 20 seeds.

\subsection{Measurement of the preactivation mean, the activation derivative and death}
On the same evaluation set we measure, per unit, $m=\mathrm{mean}(z)$, $s=\mathrm{sd}(z)$, $r=m/s$ ($s>0$) and $p=\Pr(z>0)$. For convolutional layers, images and spatial positions are pooled per channel; for diagnostics of the derivative through max-pooling, the position of the winner is used. The scale is the standard deviation of the preactivation distribution, not a number of units. Using the old $w,\mu$, the finite displacement of the preactivation mean is
\[\Delta m=(\Delta w)^\top\mu+w^\top\Delta\mu+(\Delta w)^\top\Delta\mu+\Delta b.\]
In bookkeeping that uses the new $w$, the cross term enters the upstream term. We distinguish the ratio of the absolute values of the terms from their shares of the signed total displacement.

\begin{center}\footnotesize\setlength{\tabcolsep}{3pt}
\begin{tabular}{@{}P{2.5cm}P{5.5cm}P{8.0cm}@{}}
\toprule
\textbf{Series} & \textbf{Set} & \textbf{Definition of death and of low derivative} \\
\midrule
image MLP & all training images of the task (RC/RM 1,200 images, 5+1 hard 2,500 images) & units with maxₓ|φ′(z)|<1e−6; distinct from the rate of low derivative over pairs (x, unit) \\
RC CNN centering & fixed 1,200 training images; for conv, spatial positions included & units with no input on which the ReLU is open; the first drop of a layer is defined separately by a median unit open fraction <.01 \\
5+1 CNN & 20 images from each of all 100 classes; fixed training probe of 2,000 images & units with max over the probe of φ′<1e−6 (for ReLU, equivalent to open fraction 0 on the probe; the open fraction of §2.1 has this meaning). Low derivative is φ′<.1 per pair (x, unit), not a threshold on the open fraction \\
\bottomrule
\end{tabular}\end{center}

We distinguish the autograd derivative used in training from diagnostics by the analytic formula. Unit means and medians follow the notes of each table.

Note: death on the probe does not mean permanent death on all inputs and at all future times. The definitions follow each registration and are not statistical interventions.

\subsection{Aggregation, families of comparisons and times of registration}
\label{app:A4}
Within a seed, tasks and updates are aggregated to form paired differences. We distinguish means and medians of levels from means and medians of differences, and we do not treat non-significance as equivalence. We also distinguish post hoc interventions designed from saved trajectories from verdicts fixed before the runs.

The 97.5\% paired $t$ intervals of the original cap and of the swaps are stated in D.2/E.1; the simultaneous intervals of the additional 8/6/15 comparisons of cap/fresh/learning rate in D.3/E.3; Holm and rank intervals of the MLP in F.3; the separate-family bootstraps of the CNN in F.4/H.5; exact sign tests with Holm in H.4.

The times and rules of fixing are kept in the registration records of each section; for the additional controls, the registration record of each result is authoritative.

Note: re-aggregation of the same saved conditions and the same seeds is not counted as independent replication. The CNN budget of 780 updates is the same trajectory as the 12 conditions.

\subsection{Centering, bias suppression and the scale cap in the MLP}
\label{app:A5}
C subtracts a fixed mean before learning starts: for RC/RM, the per-feature mean of the 1,200 images chosen by the seed; for the 5+1 MLP, the mean image of the \textbf{whole training set} of CIFAR-100; per-task means are not tracked. H subtracts per-unit EMA means from the activation outputs of each hidden layer: EMA coefficient .01, initial value 0, statistics detached, not reset at task boundaries, the same values used in training and evaluation. No variance normalization or affine transform is added.

\begin{center}\footnotesize\setlength{\tabcolsep}{3pt}
\begin{tabular}{@{}P{3.2cm}P{8.0cm}P{4.8cm}@{}}
\toprule
\textbf{Operation} & \textbf{Done after each update} & \textbf{Not changed directly} \\
\midrule
C / H / CH & fixed input centering / EMA centering of both hidden outputs / both & Adam history \\
CHB (RC) & CH plus: first-layer bias set to 0 after every update; second-layer bias b←(1−ηλ)b, λ=.1713 & weights, output layer; B is prevention from task 1 \\
CHB (5+1 MLP) & same form, λ=.835 calibrated on a separate probe & the λ of RC is not reused \\
cap1 / cap2 / cap12 (RC, 5+1) & fix the row norm Rᵢ at the end of task 1; from task 2 on, after Adam, wᵢ←min(1,Rᵢ/‖wᵢ‖)wᵢ & bias, Adam moments; zero rows left as they are \\
cap12\_bfix (auxiliary) & cap12 plus: bias returned to its value at the end of task 1 & kept separate from the main comparison of cap12 \\
cap12′ (RM) & caps on the first-layer norms along and orthogonal to the input-mean direction, and on the second-layer row norms & details and operations after the branch in E.2 \\
\bottomrule
\end{tabular}\end{center}

Centering by an EMA lags and does not guarantee an exact zero mean at each time. CHB combines two bias operations. The cap is a norm projection and can change the scale, the direction of updates and the output at the same time.

Note: definitions of the operations. The results are registered interventions with 10 seeds each; the diagnostics are post hoc.

\section{Terms and symbols}
\label{app:B}
Table~\ref{tab:terms} of the main text gives the terms. Symbols: $m_i$ preactivation mean; $s_i$ scale (standard deviation); $r_i=m_i/s_i$ standardized mean; $\varphi'$ activation derivative; $\kappa$ readout alignment; $S_i$ switch gradient; $S^{\rm rand}$ common error (labels of the new task replaced by class frequencies); $L_i$ label term; $K_n=\mu^\top x_n+1$ coefficient of the input geometry (distinct from the number of classes $C$); $g_{in}=K_n\varphi'(z_{in})$ backpropagation factor (the ``gated activity'' of App.~C.7; not the activation value $\varphi(z)$); $\bar g_{i,c}$ mean of $g_{in}$ over the images of class $c$; $\pi_n$ prediction before the switch; $e_n=\pi_n-y_n$ output error; $\bar e=\bar\pi-\bar y$ mean output error (minibatch mean).

\textbf{The term removed by error-mean removal (identity B.1; §4.5).} For parameters $\theta$ below the output layer (weights, biases or embeddings), let $J_n=\partial\ell_n/\partial\theta$ be the Jacobian of the logits $\ell_n$ of input $n$ and $e_n=\pi_n-y_n$ the output error; the one-step gradient of the mean cross-entropy is $\frac1N\sum_nJ_n^\top e_n$. Error-mean removal turns this into $\frac1N\sum_nJ_n^\top(e_n-\bar e)$, so the removed amount is $\bar J^\top\bar e$ ($\bar J=\frac1N\sum_nJ_n$). On the other hand, the common-error gradient, in which the correct labels of the new task are replaced by the class frequencies $\bar y$ of the minibatch, splits identically, using $\pi_n-\bar y=(\pi_n-\bar\pi)+(\bar\pi-\bar y)$, as
\[\frac1N\sum_nJ_n^\top(\pi_n-\bar y)=\bar J^\top\bar e+\frac1N\sum_n(J_n-\bar J)^\top(\pi_n-\bar\pi).\]
Error-mean removal removes only the first term; the second term, the covariance of the per-input Jacobian and the fluctuation of the prediction (which also does not depend on the label assignment), remains. The label term is $\frac1N\sum_nJ_n^\top(\bar y-y_n)$, and the actual gradient = common-error gradient + label term.
\begin{itemize}
\item \textbf{Weights of the last hidden layer}: for the input weights $w_i$ of unit $i$, $J_n^\top e=\varphi'(z_{in})\langle v_i,e\rangle x_n$ ($v_i$ is the readout weight from unit $i$ to the output and $x_n$ the layer input), so the first term is $\langle v_i,\bar e\rangle\,\overline{\varphi'_ix}$ ($\overline{\varphi'_ix}=\frac1N\sum_n\varphi'(z_{in})x_n$) and the second term is $\operatorname{Cov}_n\!\big(\varphi'(z_{in})x_n,\langle v_i,\pi_n\rangle\big)$ (covariance with $1/N$). In networks whose bias learns, 1 is appended to $x_n$.
\item \textbf{Relation to the form of the preactivation mean}: with the layer input fixed and the weights and bias moved at the same learning rate, the change of the preactivation mean is $\Delta m_i=\Delta w_i^\top\mu+\Delta b_i$, so the form of Eq.~\ref{eq:S} with $K_n=\mu^\top x_n+1$ ($S_i^{\rm rand}=\bar g_i\langle v_i,\bar e\rangle+\operatorname{Cov}_n(g_{in},\langle v_i,\pi_n\rangle)$, $g_{in}=K_n\varphi'(z_{in})$) is the projection of the above weight and bias gradients onto $(\mu,1)$. This projected form is used only when looking at the preactivation mean in networks whose bias learns. In networks with the hidden biases fixed at 0 (the network of §4.5), the per-parameter form holds as is.
\item \textbf{Size}: the sizes of the first and second terms in the network of §4.5 (real-label 5+1 MLP) were not measured (the image data were available to us only in pickle format, which we do not load). The values for the language model and for the Continual ImageNet CNN, where the same split was measured on saved weights, are in K.4.
\end{itemize}

\textbf{Families of intervals and tests (by appendix).} Each family of comparisons was fixed when that comparison was registered and was not regrouped after seeing the results. When the same difference appears in two families (e.g., cap12 for ReLU on real-label 5+1 = D.2 and D.3), both are reported.

\begin{center}\footnotesize\setlength{\tabcolsep}{3pt}
\begin{tabular}{@{}P{3.0cm}P{4.0cm}P{6.3cm}P{3.3cm}@{}}
\toprule
\textbf{Appendix} & \textbf{Comparison} & \textbf{Interval, test} & \textbf{Family} \\
\midrule
D.1, D.4, H.1, H.2 & checks on saved states & descriptive (no interval) & — \\
D.2, D.3 (RC/ELU) & main comparison of the scale cap & 97.5\% paired t interval (95\% for one-layer caps of D.2) & 2 main comparisons \\
D.3 (additional 2×2), D.3 (cost on networks trained from scratch) & scale cap & Bonferroni simultaneous 95\% interval & 8 comparisons, 6 comparisons \\
E.1 & derivative swap (restore, give) & 97.5\% paired t interval & 2 main comparisons per setting \\
E.2 & swap between the cap12′ and control networks & 95\% paired t interval (97.5\% for the main comparison) & Q2 kept separate from the main family \\
E.3 & learning-rate controls & Bonferroni simultaneous 95\% t interval & 15 comparisons (3 settings × 5) \\
F.1 & comparison with Snake of fixed scale & 97.5\% paired t interval (95\% for the secondary) & 2 main comparisons \\
F.3 & 19 conditions and existing methods & median of paired differences and Holm p; vs L2 Init, 95.9\% interval of the median and equivalence band ±0.5 points & 18 comparisons, 3 comparisons \\
F.4 & 12 conditions of the real-label CNN & registered intervals (bootstrap, A.4); kunekune vs ELU at 7,800 updates: unadjusted bootstrap 95\% (post hoc) & 8 main and 8 secondary comparisons kept separate \\
H.5 & budget extension & Bonferroni bootstrap simultaneous 95\% interval & 8 comparisons per family \\
H.3 & CH′+B vs CH′ (run without TF32) & 95\% paired t interval & 1 comparison \\
H.4; cap of H.5 & derivative swap; adding B later; scale cap & two-sided exact sign test with Holm (H.4 also seed-bootstrap 95\% intervals) & 2 main comparisons \\
G & error-mean removal & verdicts fixed before the runs (as defined in the verdict files) & per verdict \\
J (including J.14) & pairwise comparisons on the standard sequences & 95\% t interval of same-seed paired differences and the criterion (intervention better = 10/10 and mean ≥ +0.5 points; other better = ≥ 8/10 and mean ≤ −0.5 points; otherwise undecided) & per comparison (bar raised across runs = J.5) \\
\bottomrule
\end{tabular}\end{center}

\section{Theory and proofs}
\label{app:C}
Main text: §3.1, §3.3.

\subsection{Cross-entropy and a lower bound on the alignment coefficient of the readout}
Let the logits be $\ell_n=V\phi_n+b^c$, with $V^c$ and $b^c$ centered along the class direction. Set $F=\sum_n(y_n-\mathbf1/C)\phi_n^\top$, $A=\langle V^c,F\rangle$ and $\kappa=A/\|F\|_F^2$ ($F\ne0$).

By Jensen's inequality, $\log\sum_c e^{\ell_{nc}}\ge\log C+\bar\ell_n$. With $L$ the mean CE,
\[A=\sum_n(\ell_{n,y_n}-\bar\ell_n-b^c_{y_n})\ge N(\log C-L)-\Gamma,\qquad \Gamma=\sum_c(n_c-N/C)b_c^c.\]
Hence $\kappa>0$ if $L<\log C-\Gamma/N$. $\Gamma=0$ if the class counts are balanced or the output layer has no bias. The lower bound follows from an identity and an inequality that hold for any fixed state. It does not guarantee the sign of per-unit gradients, the residual condition of H1, or the future rebuilding of the alignment. In states with $F=0$, $\kappa$ is not defined, and 0 is not substituted.

Note: derivation by an identity and Jensen's inequality.

\subsection{H1 and the derivation of Eq.~3}
For unit $i$ let the preactivation mean be $m_i=w_i^\top\mu+b_i$ and $K_n=\mu^\top x_n+1$. One SGD step on the mean loss with the upstream fixed, averaged over new uniform labels, gives
\[\mathbb E_{\rm new}[\Delta m_i]=-\frac\eta N S_i,\quad S_i=\sum_nK_n\varphi_i'(z_{in})\langle v_i^c,\pi_n-\mathbf1/C\rangle.\]
With the class-centered $v^c$, the $1/C$ inside the inner product can be dropped. $K$ is a coefficient of the input geometry, distinct from the number of classes $C$. Adam, finite windows and upstream updates are not included in this formula itself.

\textbf{H1.} (i) The old labels $y_n$ are independent and uniform across inputs; (ii) the new labels are independent and uniform, independent of the old labels and of the network; (iii) the features $\varphi$, the derivatives $\varphi'$ and $K$ are fixed within the average over old labels; (iv) $\kappa>0$ is fixed for that average; (v) writing the readout as $v_i=v_i^0+\kappa\sum_m(y_m-1/C)\varphi_{im}$, the correlation of $v_i^0$ with the old labels vanishes; and (vi) the limit of confident correct predictions on the old task, $\pi_n\to y_n$, is taken. This is a fixed-feature model, not an assumption about real states in general, where $\varphi$ depends on the labels through learning.

Averaging over the old labels,
\[\mathbb E_{\rm old}[\langle v_i^c,y_n-\mathbf1/C\rangle]=\kappa\sum_m\varphi_{im}\mathbb E[\langle y_m-\mathbf1/C,y_n-\mathbf1/C\rangle]=\kappa(1-1/C)\varphi_{in},\]
because the expectation of the inner product is 0 for $m\ne n$ and $1-1/C$ for $m=n$. Therefore
\[\mathbb E_{\rm old,new}[S_i]=\kappa(1-1/C)\sum_n K_n\varphi_i'(z_{in})\varphi_i(z_{in}).\]
Substituting a measured positive $\kappa$ does not make the off-diagonal terms, the projection residuals and the finite confidence vanish. D.1 checks real states including these residuals.

For ReLU the product is $\varphi\varphi'=z_+$. If $K_n\ge0$, the sum of Eq.~3 is nonnegative, and it is positive if $K_n>0$ for at least one open input. All $K_n>0$ is a sufficient condition, but the distribution-free positivity is not transferred to cases with inputs where $K<0$, such as std. Finite displacements in which the upstream also moves are booked with the formula of A.3 including the cross term.

Note: a derivation averaging over new and old labels under conditions; not a statement about finite-time reach, stopping points or convergence of real training.

\subsection{The sign boundary of leaky ReLU}
Let $z=m+s\xi$, $\xi\sim N(0,1)$, $r=m/s$, $s>0$ and $K_n=\bar K>0$, and let $\phi,\Phi$ be the density and distribution function of the standard normal. For leaky ReLU with negative slope $0<a<1$, $\varphi\varphi'$ is $z$ on the positive side and $a^2z$ on the negative side, so from the first moment of the truncated normal
\[G(r)=\frac{\mathbb E[\varphi\varphi']}{s}=\phi(r)+r\Phi(r)+a^2\{r\Phi(-r)-\phi(r)\}.\]
Here only, $\phi$ is the normal density, distinct from the activation $\varphi$. Since $\phi'(r)=-r\phi(r)$,
\[G'(r)=\Phi(r)+a^2\Phi(-r)>0.\]
Since $G(0)=(1-a^2)/\sqrt{2\pi}>0$ and $G(r)\sim a^2r\to-\infty$ as $r\to-\infty$, there is exactly one zero, on the negative side.

\begin{center}\footnotesize\setlength{\tabcolsep}{3pt}
\begin{tabular}{@{}lll@{}}
\toprule
\textbf{a} & \textbf{relative zero r* (numerical)} & \textbf{open fraction Φ(r*)} \\
\midrule
.1 & about −1.72 & about .043 \\
.3 & about −.94 & about .174 \\
.5 & about −.55 & about .291 \\
\bottomrule
\end{tabular}\end{center}

$r^*$ does not depend on $s$, and the absolute zero $m^*=sr^*$ deepens with the scale. It is not a stopping point of real trajectories.

Note: the closed form and the uniqueness are exact under Gaussian $z$ and constant positive $K$; the table of zeros is numerical and not subject to seeds or confidence intervals.

\subsection{A pointwise inequality for SiLU and GELU}
Set $A(z)=\mathrm{sigmoid}(z)$ for SiLU and $A(z)=\Phi(z)$ for GELU; in both, $\varphi(z)=zA(z)$ and $\varphi'(z)=A(z)+zA'(z)$. $A$ is strictly increasing, $A(0)=1/2$, $A'>0$.
\[\begin{array}{ll}
z>0:& A>1/2,\ A+zA'>A\quad\Rightarrow\quad zA(A+zA')>z/4,\\
z<0:& 0<A<1/2,\ A+zA'<A\quad\Rightarrow\quad A(A+zA')<A^2<1/4\\
&\hspace{42mm}\Rightarrow\quad zA(A+zA')>z/4,\\
z=0:& \varphi\varphi'=z/4=0.
\end{array}\]
Hence $\varphi\varphi'\ge z/4$, strictly for $z\ne0$. Under constant positive $K$, if $m=\mathbb E[z]\ge0$ and the distribution is non-degenerate, the sum of Eq.~3 is positive. $\sum K_n\varphi\varphi'\ge\sum K_nz_n/4$ also transfers to per-input $K_n\ge0$, but the sign of the right-hand side cannot be determined from $m$ alone. This proof is not applied to negative $K_n$, to the covariance of the second phase after residuals are taken, or to gate terms involving $\varphi''$.

Note: a distribution-free pointwise algebraic inequality.

\subsection{A model of the second phase (after the sum of Eq.~3 weakens) and H2--H4}
In the second phase, we remove from the features of the target unit the components that other units and the output bias can reproduce. ``Overlap'' is the linear overlap between feature columns on the same inputs; we do not assume, from the identity of a single ReLU alone, that the surroundings carry that linear component.

\textbf{H2 (model).} Freeze the features and local Jacobians at a saved state of the network and fit the old labels with squared error. The old labels are independent across inputs, with mean 0 and equal variance $\sigma_Y^2$; the new labels are independent. Take the ridge of the readout $\lambda\to0$ and require the columns of the augmented feature matrix $\tilde F=[F,1]$ to be linearly independent. The version that links the slope of the capacity to the expected step by a local expansion in all parameters keeps terms up to second order in the label amplitude and separately requires the fit-completion limit $JJ^\top\succ0$. This is not the same model as the real trajectories of CE and Adam.

Let the feature column of unit $i$ be $a_i=(\varphi(z_{in}))_n$, the derivative along the mean direction $u_i=(K_n\varphi'(z_{in}))_n$, and $P_\perp$ the projection onto the orthogonal complement of the other feature columns and the constant column. The least-squares readout is $v=(\tilde F^\top\tilde F)^{-1}\tilde F^\top Y$. After averaging over the new labels, the gradient of the preactivation mean is $v_i\langle u_i,f\rangle$, and averaging further over the old labels gives $\sigma_Y^2\langle u_i,\omega_i\rangle$. Here $\omega_i=\tilde F(\tilde F^\top\tilde F)^{-1}e_i$ is orthogonal to the other columns with $\langle\omega_i,a_i\rangle=1$, so
\[\omega_i=\frac{P_\perp a_i}{\|P_\perp a_i\|^2},\qquad
\mathbb E[\partial_{m_i}L_{\rm new}]=\sigma_Y^2\frac{\langle P_\perp u_i,P_\perp a_i\rangle}{\|P_\perp a_i\|^2}.\]
A complete overlap with a zero denominator lies outside the linear-independence condition. A positive numerator lowers the preactivation mean under gradient descent.

\textbf{H3 (an idealization for reading the sign).} $K_n=\bar K>0$, and $z$ is Gaussian where needed. If the surroundings carry only the constant component, the numerator is $N\bar K\operatorname{Cov}(\varphi',\varphi)$. If $\varphi$ is increasing and $\varphi'$ nondecreasing, then for an independent $Z'$
\[\operatorname{Cov}(\varphi',\varphi)=\tfrac12\mathbb E[(\varphi'(Z)-\varphi'(Z'))(\varphi(Z)-\varphi(Z'))]\ge0.\]
It is positive for ReLU, leaky ReLU and ELU ($\alpha\le1$) unless the derivative is constant on the support. For ReLU it equals $(1-p)\mathbb E[z_+]$. Because the mean term of $\mathbb E[\varphi\varphi']=\operatorname{Cov}(\varphi',\varphi)+\mathbb E[\varphi']\mathbb E[\varphi]$, the sum of Eq.~3, is removed, a positive part can remain even deeper than the zero of the sum of Eq.~3.

If the surroundings carry not only the constant but also feature columns, those columns enter $P_\perp$ above. The sign of the covariance in the constant-only case must not be transferred to arbitrary feature configurations.

In a local model in which the gate also moves, differentiating the capacity $\tfrac12\log\det(\mathcal K+\lambda I)$, with the kernel $\mathcal K$ and $E=(\mathcal K+\lambda I)^{-1}$, gives the gate gradient $A^g_i=\sum_nc_{in}(d_{in}-\hat d_{in})$. Here $c_{in}\ge0$ is the condition of positive $K$ and a monotone smooth gate, and $\hat d$ is the valley floor formed by the other units. Because $\hat d$ changes with the actual configuration of the inputs, H4 below is stated separately.

\textbf{H4 (condition on the actual configuration).} In a one-output model, writing the contributions of the other units to the valley floor as $L_k$ and the amount needed for reversal as $N_i$, $A^g_i=N_i-\sum_{k\ne i}L_k$. We require $\sum L_k<N_i$ in each target state. This is the condition that the gate gradient does not reverse in the actual configuration, not a general theorem. If $N_i>0$ and $A^g_i\le0$, at least $N_i/\max_kL_k$ units with positive contributions are needed (pigeonhole principle). Closed other units of ReLU contribute 0, but for leaky ReLU the negative-side slope remains, so the condition is not replaced simply by ``the number of open units''.

We distinguish the sign of the covariance of permutation averages from the sign in actual configurations satisfying H4. For SiLU/GELU $\varphi''$ changes sign, and the sign region of the gate covariance of ELU is also limited, so the measurements of H4 for leaky ReLU are not transferred to them.

Note: a conditional model with freezing, linearization, squared error and the ridge limit. It contains no claim of finite-time reach or of convergence to stable points. The detailed derivation is in our derivation record and is not included in this appendix.

\subsection{Scale and the zero of the sum of Eq.~3: ELU, SiLU, GELU}
With constant $K$ and $Z\sim N(m,s^2)$, compute $H_s(m)=\mathbb E[\varphi(Z)\varphi'(Z)]$. For ELU ($0<\alpha\le1$), $h(z)=\varphi\varphi'$ has the same sign as $z$. At a point with $H_s(m)=0$, differentiating the density in $m$ gives $H_s'(m)=\mathbb E[h(Z)Z]/s^2>0$. Since $H_s(0)>0$ and $H_s(m)<0$ for sufficiently negative $m$, there is only one zero, on the negative side. Non-degeneracy and the range of $\alpha$ are not relaxed.

\begin{center}\footnotesize\setlength{\tabcolsep}{3pt}
\begin{tabular}{@{}P{6.0cm}P{10.0cm}@{}}
\toprule
\textbf{Activation, condition} & \textbf{numerical zero of the Eq. 3 sum / range of existence} \\
\midrule
ELU α=1, s=1 & r*≈−.72 \\
ELU α=1, s=4 & r*≈−2.75 \\
ELU α=1, s=16 & r*≈−10.7 \\
SiLU & a zero exists for scales below the critical s≈.7254 (numerical boundary, tangent at the critical value) \\
GELU & likewise, below s≈.4249 \\
kunekune, adaptive Snake (α\_s=0.6/s; scale follows the preactivation scale) & r*≈−.99, independent of the scale (homogeneity); mean derivative there .55; .83 at r=−.3; 1.00 at r=0 \\
\bottomrule
\end{tabular}\end{center}

The zero solves $\int\varphi(sz+m)\varphi'(sz+m)\phi(z)\,dz=0$. The search for SiLU/GELU used 1,121 points of $r\in[-40,0]$ (steps .25 on $[-40,-20)$, .05 on $[-20,-8)$ and .01 on $[-8,0]$) and 1,024 logarithmic points of $s\in[.3,10^4]$; float64 composite Gauss--Legendre quadrature with 16 points, a window within 80 nats of the maximum of the envelope, and interval width $\min(.5s, 4/|\text{envelope slope}|)$, plus steps of 1 over the activation kernel $[-40,40]$. From the sign changes, 46 bisection steps in $\log s$; the maximum over continuous $r$ was refined by golden-section search and confirmed with mpmath at 34 digits, 24 points and a 115-nat window. Independent checks with 50 digits, 32 points and a 150-nat window, adaptive quadrature, and the closed form of GELU are also saved.

We do not call the numerical boundaries analytic thresholds. ELU with $\alpha=3.6$ in the CNN lies outside the sign condition $\alpha\le1$.

Note: the uniqueness for ELU is a proof conditional on Gaussian $z$ and constant $K$; the table of zeros and the critical values come from the saved numerical search, which was not rerun.

\subsection{The label term of the switch gradient (lemma)}
Under cross-entropy and a linear readout, the expected gradient of one step right after a task switch, through the own weight and bias paths of unit $i$, written with the inputs $\{x_n\}$ and one-hot labels $\{y_n\}$ of the new task, the predictions $\pi_n$ before the switch, and the gated activity $g_{in}=K_n\varphi'_{in}$, is
\[S_i^{\rm real}=\frac1N\sum_n g_{in}\,\langle v_i,\ \pi_n-y_n\rangle .\]
With the empirical label distribution of the new task $\bar y=(1/N)\sum_n y_n$, identically
\[S_i^{\rm real}=S_i^{\rm rand}-L_i,\qquad S_i^{\rm rand}=\frac1N\sum_n g_{in}\langle v_i,\pi_n-\bar y\rangle,\qquad L_i=\frac1N\sum_n g_{in}\langle v_i,\ y_n-\bar y\rangle .\]
With $\bar g_{i,c}$ the mean of $g$ over the images of class $c$, $\bar y_c$ the frequency of $c$, $\bar v_i=\sum_c \bar y_c v_{i,c}$ and $\bar g_i=\sum_c \bar y_c \bar g_{i,c}$,
\[L_i=\sum_c\bar y_c\,(v_{i,c}-\bar v_i)(\bar g_{i,c}-\bar g_i)=\operatorname{Cov}_{c\sim\bar y}\!\big(v_{i,c},\ \bar g_{i,c}\big)\]
[identity]. $S^{\rm rand}$ does not depend on the label assignment; it is determined only by the state before the switch ($\pi$, $v$, gate) and by the inputs and class frequencies of the new task. $L_i$ is the covariance, over the classes of the new task, between ``the readout weight from unit $i$ to class $c$'' and ``the mean activity of unit $i$ on the images of $c$''.

Corollary: if the readout rows of the new classes are constant for unit $i$ ($v_{i,c}$ independent of $c$), then $L_i=0$. In settings where classes are not revisited, the rows of unseen classes are updated before the switch only as negatives ($\Delta v_{i,c}\propto-\sum_n \pi_{n,c}\varphi_{in}$, with $\pi_{n,c}$ the small and similar predicted probabilities of unseen classes), so they settle to nearly common values per unit, and $L_i$ is small [reading]. In measurements (H.2; real-label 5+1 CNN; 4 conditions × 3 budgets × 10 seeds), the median of $|L|/|S^{\rm rand}|$ is 0.07--0.12 and the sign agreement 0.95--0.98. Random labels (labels independent of the inputs) are the limit in which the expectation of $L_i$ is exactly 0, and the ``independence of the new labels'' of H1 can be replaced by this corollary. In settings that return to learned classes (revisits), we predict that $L$ becomes larger (not measured).

Note: a one-step expected gradient, without the Adam denominator, through the own weight path only.

\section{Checks in the MLP}
\label{app:D}
Main text: §4.1--4.3.

\subsection{The readout alignment and the sum of Eq.~3 against the switch gradient}
In saved states of the 1-hidden-layer MNIST network, we compare the expected gradient averaged over the new labels, $S_{\rm exact}$, with the sum of Eq.~3, $S_{\rm self}$. The sign is measured as the agreement rate over units, and the size as the median of the per-unit ratio (not a mediation ratio). Live units have $p>0$. The actual displacement is over 200 Adam updates after the switch.

In runs that vary the strength of the alignment step by step, from 50\% training accuracy on, the quartiles of the expected gradient/$(\log C-L)$ are .18--.22. Below are $\kappa$ at the ends of tasks and the per-unit gradient tables; these are descriptive aggregates over 2 seeds × 5 activations, without a family of intervals.

74--95\% of the rebuilding of $\kappa$ was carried by changes of the features (ends of tasks 2--29).

\begin{center}\footnotesize\setlength{\tabcolsep}{3pt}
\begin{tabular}{@{}llll@{}}
\toprule
\textbf{Activation} & \textbf{κ, task 1} & \textbf{κ, tasks 19–29} & \textbf{min / task 1} \\
\midrule
leaky 0.1 & 0.0102 & 0.0200 & 0.86–1.00 \\
ReLU & 0.0099 & all dead & 1.00 \\
ELU & 0.0123 & 0.0170 & 0.83–1.00 \\
SiLU & 0.0075 & 0.7560 & 1.00 \\
GELU & 0.0085 & 0.6683 & 1.00 \\
\bottomrule
\end{tabular}\end{center}

\begin{center}\footnotesize\setlength{\tabcolsep}{3pt}
\begin{tabular}{@{}P{1.6cm}P{1.4cm}P{2.6cm}P{3.0cm}P{3.0cm}P{3.0cm}@{}}
\toprule
\textbf{Activation} & \textbf{Tasks} & \textbf{\% lowering in expectation} & \textbf{\% sign agreement of the Eq. 3 sum} & \textbf{\% shifted negative in 200 updates} & \textbf{Eq. 3 sum / full expectation} \\
\midrule
leaky 0.1 & 2–3 & 94\% & 91\% & 100\% & 0.34 \\
leaky 0.1 & 4–8 & 84\% & 52\% & 92\% & 0.03 \\
leaky 0.1 & 9–19 & 69\% & 41\% & 84\% & −0.14 \\
leaky 0.1 & 20–30 & 67\% & 40\% & 82\% & −0.19 \\
ReLU & 2–3 & 93\% & 93\% & 99\% & 0.32 \\
ReLU & 4–8 & 100\% & 100\% & 98\% & 0.27 \\
ReLU & 9–14 & 100\% & 100\% & 100\% & 0.51 \\
ELU & 2–3 & 93\% & 90\% & 100\% & 0.21 \\
ELU & 4–8 & 99\% & 97\% & 100\% & 0.18 \\
ELU & 9–19 & 99\% & 98\% & 97\% & 0.13 \\
ELU & 20–30 & 98\% & 98\% & 93\% & 0.15 \\
SiLU & 2–3 & 93\% & 93\% & 100\% & 0.46 \\
SiLU & 4–8 & 95\% & 95\% & 95\% & 0.30 \\
SiLU & 9–19 & 95\% & 95\% & 88\% & 0.50 \\
SiLU & 20–30 & 95\% & 95\% & 86\% & 0.57 \\
GELU & 2–3 & 94\% & 94\% & 99\% & 0.42 \\
GELU & 4–8 & 98\% & 98\% & 92\% & 0.41 \\
GELU & 9–19 & 96\% & 96\% & 88\% & 0.59 \\
GELU & 20–30 & 98\% & 98\% & 83\% & 0.60 \\
\bottomrule
\end{tabular}\end{center}

Note: post hoc analysis of saved states, 2 seeds × 5 activations, comparing expected gradients with actual displacements. The runs that vary the strength of the alignment are registered runs.

For leaky ReLU, 67--84\% of units are on the lowering side in expectation from task 4 on (94\% for tasks 2--3). The sign agreement of 90--100\% and the magnitude of 13--60\% of the sum of Eq.~3 are not transferred to leaky ReLU over the whole period.

All 1,000 units (100 units × 2 seeds for each activation) entered $p<1/2$ by the end of tasks 1--2; only 4 ELU units returned to $p\ge1/2$ at a later task end (1 in seed 0, 3 in seed 1). The other 4 activations: 0/200 each. The 4 units that returned were again at $p<1/2$ at the end of the next task. This is a post hoc description; the number of independent replications remains 2 seeds.

\subsection{Scale cap for ReLU on 5+1: the original family of intervals}
The cap of A.5 on 5+1 ReLU. $\delta_1$ = cap − control in late online (accuracy rate). $n=10$, df 9; 97.5\% paired $t$ interval for the main cap12 and 95\% for one layer. A family separate from D.3.

The difference cap12 − control in late online is −3.60 pt, 97.5\% interval [−6.70, −0.50].

States are means over the late hard tasks. We show the preactivation mean and the scale (standard deviation), distinct from the standardized mean $m/s$.

\begin{center}\footnotesize\setlength{\tabcolsep}{3pt}
\begin{tabular}{@{}P{2.2cm}P{3.5cm}P{3.0cm}P{4.5cm}@{}}
\toprule
\textbf{Condition} & \textbf{late preactivation mean} & \textbf{late scale} & \textbf{late fraction of dead units, layer 2} \\
\midrule
control & −8.53 & 5.46 & .342 \\
cap12 & −5.61 & 3.67 & .399 \\
\bottomrule
\end{tabular}\end{center}

Note: registered intervention, 10 seeds.

\subsection{One-layer caps, the additional 2×2, and the cost on networks trained from scratch}
The operations are those of A.5. The original RC/ELU run uses the seed mean of online over tasks 31--50, 10 seeds. The 2 main comparisons cap12 − control use 97.5\% paired $t$ intervals; the one-layer caps are location controls.

\begin{center}\footnotesize\setlength{\tabcolsep}{3pt}
\begin{tabular}{@{}ll@{}}
\toprule
\textbf{RC/ELU condition} & \textbf{late online} \\
\midrule
control & 0.10088 \\
cap1 & 0.11914 \\
cap2 & 0.25096 \\
cap12 & 0.85977 \\
cap12\_bfix & 0.78854 \\
\bottomrule
\end{tabular}\end{center}

cap12 − control is +75.8889 pt, 97.5\% interval [+75.1465, +76.6312]. The norm of the second-layer input mean went from a common 52.568 at the end of task 1 to 1157.213 for the control and 40.195 for cap12 at the end of task 50 (seed mean, descriptive).

The additional 2×2 uses Bonferroni simultaneous 95\% intervals (pt) for 4 cells × 2 items = 8 comparisons, a family different from the original run of D.2.

\begin{center}\footnotesize\setlength{\tabcolsep}{3pt}
\begin{tabular}{@{}P{2.6cm}P{2.4cm}P{4.6cm}P{6.0cm}@{}}
\toprule
\textbf{Cell} & \textbf{cap difference} & \textbf{simultaneous 95\% interval (8 comparisons)} & \textbf{registered verdict} \\
\midrule
RC/ELU & +75.889 pt & [+74.908, +76.869] & HELP \\
RC/ReLU & about 0.000 pt & [−0.0000175, +0.0000133] pt & direction verdict UNRESOLVED; the separate verdict: equivalent within ±0.5 pt \\
5+1 MLP/ELU & −1.999 pt & [−2.495, −1.504] & NO\_RESCUE, HARM present \\
5+1 MLP/ReLU & −3.601 pt & [−7.696, +0.494] & UNRESOLVED \\
\bottomrule
\end{tabular}\end{center}

The post hoc fresh controls use hard task 29 with 10 seeds each; simultaneous intervals for 6 comparisons (2 activations × 3 differences). Differences in pt.

\begin{center}\footnotesize\setlength{\tabcolsep}{3pt}
\begin{tabular}{@{}P{2.0cm}P{4.4cm}P{4.4cm}P{4.8cm}@{}}
\toprule
\textbf{Activation} & \textbf{cap difference, fresh} & \textbf{cap difference, continued} & \textbf{interaction (continued difference − fresh difference)} \\
\midrule
ELU & −0.056 pt [−0.084, −0.028] & −1.999 pt [−2.469, −1.529] & −1.943 pt [−2.416, −1.471] \\
ReLU & −0.036 pt [−0.073, +0.002] & −3.601 pt [−7.486, +0.283] & −3.565 pt [−7.450, +0.319] \\
\bottomrule
\end{tabular}\end{center}

Note: the additional 2×2 has verdicts fixed in advance with 10 seeds; the fresh cost is a post hoc control with 10 seeds.

Interaction = cap difference in continued networks − cap difference in fresh networks. The 97.5\% intervals of the original run, the additional 8 comparisons and the fresh 6 comparisons are separate families of intervals.

\subsection{Checks of the second phase and H4 on actual states}
We compute the required amount $N_i$ of C.5 and the lift by other units $\sum L_k$ in saved states of the 1-hidden-layer MNIST network. A ratio of 1 is the boundary at which the direction of the gate reverses. Replications and denominators are as in the table below; unit--task pairs are not counted as independent seeds. These are descriptive aggregates without interval estimates.

\begin{center}\footnotesize\setlength{\tabcolsep}{3pt}
\begin{tabular}{@{}P{3.6cm}P{12.4cm}@{}}
\toprule
\textbf{Check} & \textbf{Target, result} \\
\midrule
fluctuation & units shifting negative within one task in the second phase 46–58\%; rank correlation with the capacity gradient .39–.61 \\
bridge coverage & leaky, 2 seeds; reversals 0/573; coverage .11–.18 (required .50–.64); at most 6 per target \\
bridge lifetime & fraction vanishing within one task 87.8–95\% \\
supply / requirement & 3 states of seed 0; medians .37/.68/.86; maxima .61/.79/.92 \\
number of other open units & target-states at or above the lower bound of the required number: 0/52,083 \\
retention & 1/53,232 over 2 seeds; the only case (a=.5, task 26) had p=.62 and resolved one task later \\
\bottomrule
\end{tabular}\end{center}

The measurements of the coverage and persistence of H4 are for leaky ReLU and are not transferred to the CNN. H4 is violated in arbitrary configurations. Nor does the table show finite-time reach.

Note: descriptions of saved real states; the numbers of seeds above are distinguished.

\subsection{Centering over 50 and 200 tasks}
\label{app:D5}
The operations C/H/CH/CHB are defined in A.5. The 50 tasks of RC with ReLU use seeds 0--9; the 200-task confirmation run uses the unused seeds 10--19. The late windows, tasks 31--50 and 181--200, are averaged within seed before aggregation. The upper (50-task) table gives seed medians and the lower (200-task) table seed means; differences are not formed directly as if they were the same statistic. Layer-wise states are descriptive.

\begin{center}\footnotesize\setlength{\tabcolsep}{3pt}
\begin{tabular}{@{}P{4.6cm}P{1.6cm}P{1.4cm}P{3.0cm}P{1.6cm}P{3.0cm}@{}}
\toprule
\textbf{Condition} & \textbf{window t31–50} & \textbf{t1 online} & \textbf{dead fraction, layer 1, end of t1} & \textbf{$\bar z_2$ at t50} & \textbf{dead fraction, layer 2, t50} \\
\midrule
control (raw, plain ReLU) & 0.1132 & 0.133 & 0.985 & −0.04 & 1.000 \\
H (hidden-layer centering only, S2) & 0.1133 & 0.626 & 0.935 & −3.2 & 0.990 \\
C (input centering only) & 0.1133 & 0.981 & 0.000 & −168 & 1.000 \\
CH (input + hidden layers) & 0.9867 & 0.984 & 0.000 & −6.8 & 0.000 \\
CHB (CH + bias decay) & 0.9854 & 0.984 & 0.000 & −0.07 & 0.000 \\
\bottomrule
\end{tabular}\end{center}

\begin{center}\footnotesize\setlength{\tabcolsep}{3pt}
\begin{tabular}{@{}P{8.0cm}ll@{}}
\toprule
\textbf{Condition (200-task run, mean of seeds 10–19)} & \textbf{tasks 31–50} & \textbf{tasks 181–200} \\
\midrule
CH: centering of the input and hidden layers & 98.63\% & 62.98\% \\
CHB: CH + zeroing of the first-layer bias + decay of the second-layer bias & 98.54\% & 99.03\% \\
\bottomrule
\end{tabular}\end{center}

Note: intervention comparisons; 50 tasks with 10 seeds and 200 tasks with 10 independent seeds.

\section{Derivative swap}
\label{app:E}
Main text: §4.4.

\subsection{Two-way swaps with a fixed offset in three settings}
\label{app:E1}
Let the recipient's preactivation at the branch be $z^0(x)$ and the donor's $z^D(x)$, and fix the per-image $d=z^D-z^0$. The activity of the second hidden layer is replaced by
\[\widetilde a(x)=\varphi(z^0(x))+\{\varphi(z(x)+d(x))-\varphi(z^0(x)+d(x))\}.\]
A frozen reference network provides $z^0$ on the current minibatch. At the branch, the activities and predictions coincide with those of the original recipient, and only the local derivative changes to $\varphi'(z+d)$. What is fixed is the reference and the offset, not the gate values during learning. The control and the intervention in the same direction share the weights, the new task and the minibatch sequence.

\begin{center}\footnotesize\setlength{\tabcolsep}{3pt}
\begin{tabular}{@{}P{2.4cm}P{3.0cm}P{3.4cm}P{4.2cm}P{3.0cm}@{}}
\toprule
\textbf{Setting / activation} & \textbf{recipient ← donor (end of task)} & \textbf{next task / budget} & \textbf{Adam} & \textbf{main aggregation} \\
\midrule
RC / ELU, std & 10←1, reverse 1←10 & 11 / 2, 30,000 updates each & continued on the restore side; initialized in both conditions on the give side & online on the next task, mean of 10 seeds \\
RM / ELU & 10←2, reverse 2←10 & 11 / 3, 6,000 updates each × 2 tasks & continued on the restore side; initialized in both conditions on the give side & online on the first task, mean of 10 seeds \\
5+1 / ReLU & 28←2, reverse 2←28 & hard 29 / 3, 780 updates each & continued on the restore side; initialized in both conditions on the give side & online on the next hard task, mean of 10 seeds \\
\bottomrule
\end{tabular}\end{center}

The 2 main comparisons of each setting (restore and give) use 97.5\% paired $t$ intervals. The table below gives, for both directions, swap − control differences; deterioration in the reverse direction is shown as negative. The restore and give sides use different tasks, so the level differences of the two directions are not compared directly.

Levels and 97.5\% intervals of the restore and give sides (the numbers of §4.4 and Figure~\ref{fig:mlp}c):

\begin{center}\footnotesize\setlength{\tabcolsep}{3pt}
\begin{tabular}{@{}P{5.4cm}P{5.4cm}P{5.4cm}@{}}
\toprule
\textbf{Setting (branch points)} & \textbf{restore the early derivative to the degraded network} & \textbf{give the degraded derivative to the early network} \\
\midrule
random-label CIFAR, ELU/std (ends of tasks 1 and 10) & 10.2\% → 57.0\% (+46.7[42.9–50.6]) & 97.1\% → 29.2\% (−67.9[−69.6 to −66.2]) \\
random-label MNIST, ELU/raw (ends of tasks 2 and 10) & +76.0[74.6–77.4] & −25.9[−31.7 to −20.2] \\
real-label 5+1 CIFAR, ReLU/std (ends of tasks 2 and 28) & 40.4\% → 61.7\% (+21.3[18.4–24.3]) & −6.0[−7.3 to −4.7] \\
\bottomrule
\end{tabular}\end{center}

Online accuracy on the next task after the swap. Point differences are paired differences of the means over 10 seeds, and [ ] are 97.5\% intervals; the tasks and budgets of each setting are in the table above. In all three settings, each direction went the predicted way in 10/10 seeds.

Note: registered branch interventions, 10 seeds each.

\subsection{Swapping the derivative field between the cap12′ network and the control network}
cap12′ of RM/ELU caps, for each first-layer row $w=qe+v$ at the end of task 1 ($e$ the unit direction of the fixed input mean, $v\perp e$), both $|q|$ and $\|v\|$, and at the same time caps the second-layer row norms. It is marked ′ to distinguish it from the usual row-norm cap12. Projection after Adam; the bias and the Adam state are not changed directly. The control and cap12′ share task 1 and continue to task 12; the end of task 10 is the main branch point.

The fixed offset is as in E.1. For the moving field, the donor's shadow continues separately without intervention, and $d(s)=z_{\rm shadow}(s)-z^0_{\rm recipient}$. The shadow continues the donor's Adam and, if the donor is capped, its cap. Recipients were Adam-initialized in all 14 conditions. Below are the 4 conditions relevant to the text of §4.4; own refers to the cap of the capped network at the end of task 1. The main quantity is online over 6,000 updates of the same new task, mean of 10 seeds each.

\begin{center}\footnotesize\setlength{\tabcolsep}{3pt}
\begin{tabular}{@{}lllll@{}}
\toprule
\textbf{Condition} & \textbf{Network} & \textbf{Field} & \textbf{Cap} & \textbf{E (continuation 1)} \\
\midrule
NC\_r & cap12 & — & own & 0.808 \\
\textbf{SC\_dyn\_r} & cap12 & control (moving) & own & \textbf{0.197} \\
\textbf{NR\_r} & control & — & none & \textbf{0.169} \\
\textbf{RR\_dyn\_r} & control & cap12 (moving) & none & \textbf{0.846} \\
\bottomrule
\end{tabular}\end{center}

Difference between the controls TE = NC\_r − NR\_r = .640, 95\% [.624, .655]. Amount lost L\_S = NC\_r − SC\_dyn\_r = .611 [.596, .626]; amount gained G\_R = RR\_dyn\_r − NR\_r = .677 [.661, .693] (all positive in 10/10; 95\% paired $t$ intervals of Q2). L\_S/TE = .96 and G\_R/TE = 1.06 are ratios of recovery and loss under an artificial intervention, not mediation ratios of the cap's benefit itself. The remaining comparisons of the main Q1 use 97.5\% and are not mixed with the Q2 family above. The initial predictions match, but the subsequent trajectories of the features, the donor and the recipient do not.

Note: registered intervention, 10 seeds. Not called a ``mediation ratio''.

\subsection{Learning-rate controls calibrated to gradient norms and Adam displacements}
We use the first 256 updates of N (the control network) and R (the network with the derivative restored) on calibration seeds 100--109. The W and bias of a layer form one group, $G^2=\|\nabla W\|^2+\|\nabla b\|^2$ and $U^2=\|\Delta W\|^2+\|\Delta b\|^2$; $U$ is the actual displacement obtained by subtracting the float32 values before and after in float64. The seed-wise RMS ratio R/N, medianed over seeds, fixes $q_G/q_U$, and the learning rate is $\eta q$ without changing the field. No clipping, constant over the whole period, Adam initial state shared with N.

Evaluation seeds 0--9. Paired online differences at RC t11/30,000 updates, RM t11/6,000 and 5+1 MLP t29/780. Of the 15 comparisons (3 settings × 5) with Bonferroni simultaneous 95\% $t$ intervals, the 6 comparisons corresponding to the claims of the main text are shown. Although only some are shown, the correction family remains the 15 comparisons.

\begin{center}\footnotesize\setlength{\tabcolsep}{3pt}
\begin{tabular}{@{}lll@{}}
\toprule
\textbf{Setting} & \textbf{Comparison} & \textbf{mean paired difference pt [simultaneous 95\% interval, 15 comparisons]} \\
\midrule
RC & LR\_G−R & -46.9939 [-52.6495,-41.3383] \\
RC & LR\_U−R & -46.9593 [-52.6150,-41.3035] \\
RM & LR\_G−R & -79.6216 [-81.3140,-77.9291] \\
RM & LR\_U−R & -78.3155 [-79.3910,-77.2400] \\
5+1 MLP & LR\_G−R & -14.8658 [-18.2728,-11.4588] \\
5+1 MLP & LR\_U−R & -13.5789 [-17.5619,-9.5960] \\
\bottomrule
\end{tabular}\end{center}

Verdict NOT\_LR\_ALL\_THREE. These are results for constant, layer-wise calibrated multipliers and do not rule out arbitrary time-varying or coordinate-wise learning rates.

Note: an intervention with verdicts fixed in advance; the 10 calibration seeds and the 10 evaluation seeds are separate.

\section{Activations: definitions and performance}
\label{app:F}
Main text: §3.3, §4.3, §5 (comparisons of levels are placed only in this appendix).

\subsection{Definitions of kunekune, adaptive Snake and fixed scales}
\label{app:F1}
Let $\alpha>0$ and $\theta=2\alpha z$. Adaptive Snake and its derivative are
\[S_\alpha(z)=z+\sin^2(\alpha z)/\alpha,\qquad S'_\alpha(z)=1+\sin(2\alpha z).\]
kunekune of the main text is KKT1:
\[\phi_\alpha(z)=\begin{cases}z&\theta<-2\pi,\\S_\alpha(z)&-2\pi\le \theta\le \pi,\\z+1/\alpha&\theta>\pi.\end{cases}\]
It is a $C^1$ function whose value and first derivative match at both ends, with tail slope 1; it is distinct from Snake, which keeps the periodic term everywhere. KKA is a $C^2$ version that joins straight lines of slope 2 outside $\theta\in[-3\pi/2,\pi/2]$, and KKA23 is 2/3 of it. In tail comparisons the slope and the joint change at the same time, so this is not a control of the tail slope alone.

The per-unit population variance $V$ starts at 1, $V\leftarrow.99V+.01\,\mathrm{Var}_{\rm batch}(z)$, and $\alpha=\mathrm{clip}(.6/\sqrt V,.005,3)$. No gradient passes through the statistics, and no learnable multiplier or shift is added. The MLP takes the variance over the minibatch, the CNN convolutions over minibatch × space, and the fully connected layers over the minibatch. The forward pass uses the old $V$, and the next $V$ is formed from the pre-update $z$ of the same forward pass. $\alpha$ fixed at the initial .6, $\alpha$ fixed at 3, and fixing after adaptation at the end of task 1 (frz1) are separate conditions. The scale-following $\alpha$ does not directly center the preactivation mean.

NEO is a loss-gated Snake. During task 1 the EMA is normal; the median CE of the last 78 updates is fixed as $L_{\rm ref}$ (lower bound $10^{-4}$), after which $g=\min(1,L_{\rm ref}/\max(L,10^{-4}))$ and the EMA coefficient is $.01g$. Common to the 4 layers; statistics detached. NEOfrz1 is fixed at the end of task 1.

\begin{center}\footnotesize\setlength{\tabcolsep}{3pt}
\begin{tabular}{@{}P{4.0cm}P{3.6cm}P{3.0cm}P{4.6cm}@{}}
\toprule
\textbf{Scale control (5+1 MLP)} & \textbf{paired difference of late online (pt)} & \textbf{Interval} & \textbf{Status} \\
\midrule
SNA − α fixed at the initial .6 & +6.1 & 97.5\% [5.5,6.8] & 10/10, registered main comparison \\
KKT1 − α fixed at the initial .6 & +4.6 & 97.5\% [3.8,5.4] & 10/10, registered main comparison \\
SNAfrz1−SNA & +.7 & 95\% [.1,1.3] & secondary comparison, report only \\
KKT1frz1−KKT1 & +1.1 & 95\% [.5,1.7] & secondary comparison, report only \\
\bottomrule
\end{tabular}\end{center}

Note: definitions of the functions and registered scale controls with 10 seeds. The definitions of NEO and of the additional fixed Snakes are in the registration records.

\subsection{All activation conditions with random labels, and scale controls}
On the same RC MLP with seeds 0--9, 13 activations are compared for both raw and std. Pre-update online over tasks 31--50 is averaged within seed, and the median over seeds is shown. raw and std are not pooled afterwards.

\begin{center}\footnotesize\setlength{\tabcolsep}{3pt}
\begin{tabular}{@{}lll@{}}
\toprule
\textbf{Condition} & \textbf{raw} & \textbf{std} \\
\midrule
KKT1 (kunekune) & \textbf{0.9565} & 0.9871 \\
SNA (adaptive Snake) & 0.9549 & 0.9870 \\
KKA & 0.9535 & 0.9871 \\
KKA23 & 0.9484 & \textbf{0.9878} \\
RSL & 0.8653 & 0.9404 \\
LK03 (leaky .3) & 0.8447 & 0.9789 \\
LR (leaky .1) & 0.7264 & 0.9655 \\
SL (Smooth-Leaky) & 0.7229 & 0.9645 \\
LK001 (leaky .01) & 0.3881 & 0.8967 \\
SiLU & 0.1233 & 0.1132 \\
GELU & 0.1160 & 0.1132 \\
ReLU & 0.1132 & 0.1132 \\
ELU & 0.1011 & 0.1005 \\
\bottomrule
\end{tabular}\end{center}

Note: registered comparison, 10 seeds, raw and std separate.

\subsection{The 19 conditions of the real-label MLP and existing methods}
\label{app:F3}
The late hard tasks of the 5+1 MLP are tasks 21, 23, 25, 27, 29. Online is averaged within each seed, and the table gives the median over 10 seeds. The median of paired differences need not equal the difference of the medians of levels. The comparisons with SNA form a Holm family of 18 comparisons.

\begin{center}\footnotesize\setlength{\tabcolsep}{3pt}
\begin{tabular}{@{}lll@{}}
\toprule
\textbf{Condition} & \textbf{Late} & \textbf{median difference vs SNA / Holm p} \\
\midrule
SNA & 0.6710 & — \\
KKA & 0.6691 & +0.0003/1.0000 \\
SNAfrz1 & 0.6688 & +0.0058/0.3281 \\
KKT1 & 0.6667 & +0.0013/1.0000 \\
ELU & 0.6493 & -0.0199/0.0352 \\
LK03 & 0.6324 & -0.0373/0.0352 \\
RSL & 0.6271 & -0.0400/0.0352 \\
KKA23 & 0.6246 & -0.0437/0.0352 \\
SL & 0.6243 & -0.0413/0.0352 \\
LR & 0.6195 & -0.0468/0.0352 \\
SILU & 0.6142 & -0.0526/0.0352 \\
SN06 & 0.6043 & -0.0632/0.0352 \\
GELU & 0.5959 & -0.0686/0.0352 \\
LK001 & 0.5915 & -0.0765/0.0352 \\
LK07 & 0.5883 & -0.0747/0.0352 \\
SN3 & 0.5405 & -0.1207/0.0352 \\
R & 0.4241 & -0.2258/0.0352 \\
DF & 0.3818 & -0.2829/0.0352 \\
CR & 0.2544 & -0.4192/0.0352 \\
\bottomrule
\end{tabular}\end{center}

The comparisons with L2 Init use 20 seeds: the existing seeds 0--9 plus the registered follow-up seeds 10--19. The 95.9\% interval of the median and a registered equivalence band of ±.005 (accuracy rate; ±.5 pt) are used.

\begin{center}\footnotesize\setlength{\tabcolsep}{3pt}
\begin{tabular}{@{}lll@{}}
\toprule
\textbf{Comparison} & \textbf{Median} & \textbf{95.9\% interval (all within ±.005)} \\
\midrule
E1 KKT1 − SNA & +0.0027 & [−0.0001, +0.0042] \\
E2 SNA − l2init(λ 1e−3) & +0.0009 & [−0.0020, +0.0042] \\
E2 KKT1 − l2init(λ 1e−3) & +0.0023 & [−0.0005, +0.0037] \\
\bottomrule
\end{tabular}\end{center}

S\&P, CBP and ReDo use the same ReLU network with lr = $10^{-4}$; the setting with the largest mean late online on calibration seeds 100--109 was fixed first and applied to the evaluation seeds 0--19.

\begin{center}\footnotesize\setlength{\tabcolsep}{3pt}
\begin{tabular}{@{}P{1.6cm}P{7.6cm}P{3.0cm}P{3.0cm}@{}}
\toprule
\textbf{Method} & \textbf{Calibration grid} & \textbf{Selected value} & \textbf{late online (\%)} \\
\midrule
S\&P & ε ∈ \{10⁻⁵,10⁻⁴,10⁻³\} × σ ∈ \{10⁻⁴,10⁻³,10⁻²\} & ε=10⁻⁵, σ=10⁻² & 51.94 \\
CBP & ρ ∈ \{10⁻⁵,10⁻⁴,10⁻³\}; maturity 100 and utility decay 0.99 fixed & ρ=10⁻⁵ & 49.90 \\
ReDo & τ ∈ \{0,0.025,0.1\} × check period F ∈ \{100,780,1560\} updates & τ=0.1, F=1560 & 53.43 \\
\bottomrule
\end{tabular}\end{center}

On the same 20 seeds, ReLU/KKT1/SNA reached 43.24/66.37/66.06\%. In the 6 comparisons of KKT1 and SNA against the 3 methods, the former was higher in 20/20, with median differences of 12.33--17.12 pt. The selections were at grid edges, and the comparison holds within this calibration range. The implementation definitions are in the registration records.

The fraction of dead units of ReDo is the 20-seed median of the within-seed means of the values at the end of the 5 late hard tasks, before reinitialization: ReLU 33.6\% → ReDo 0.2\%. The recovery ratio of performance is median(ReDo − ReLU)/median(SNA − ReLU) = 0.4298. The fraction of dead units and the recovery ratio are descriptive and do not represent a causal effect of the fraction of dead units alone.

Note: registered comparison with 10 seeds; L2 Init and the standard methods in a separate series with 20 seeds.

\textbf{Reselecting the learning rate by the same rule (§4.3).} To the comparison above (MLP on real-label 5+1 CIFAR, registered comparison at learning rate $10^{-4}$, 10 seeds) we added a comparison in which each condition selects its learning rate from $\{10^{-4}, 3\times10^{-4}, 10^{-3}\}$ by the same rule (largest late online accuracy over 10 calibration seeds), with verdicts fixed before the runs and 20 evaluation seeds. Excerpt of the verdict output (K4b and K4c label the two runs):

\begin{Verbatim}[fontsize=\scriptsize,breaklines=true,breakanywhere=true]
== registered_P2_scale_rule
                  row  median      lo      hi  a_pos  n_nonzero       p  p_holm label_holm
 SNA(K4b) - SN06(K4c) +0.1141 +0.1106 +0.1158     20         20 +0.0000 +0.0000     A_WINS
KKT1(K4b) - SN06(K4c) +0.1154 +0.1118 +0.1183     20         20 +0.0000 +0.0000     A_WINS
== registered_P3_l2init_lambda
                    row  median      lo      hi  a_pos  n_nonzero       p  p_holm label_holm
KKT1(K4b) - L2Init(K4c) +0.0603 +0.0576 +0.0639     20         20 +0.0000 +0.0000     A_WINS
 SNA(K4b) - L2Init(K4c) +0.0572 +0.0556 +0.0599     20         20 +0.0000 +0.0000     A_WINS
\end{Verbatim}

\begin{itemize}
\item The median differences of 11.4--11.5 points (20/20) of §4.3, when each condition selects its learning rate from \{0.0001, 0.0003, 0.001\} by the same rule, are SNA(K4b) − SN06(K4c) = +0.1141 and KKT1(K4b) − SN06(K4c) = +0.1154 in the excerpt above (20/20; A\_WINS after Holm). The differences are medians of seed pairs (intervals by seed bootstrap). Snake with fixed scale (SN06) selected $3\times10^{-4}$ (inside the grid).
\item Comparison 3 of the same excerpt (kunekune and adaptive Snake vs L2 Init: +0.0603, +0.0572, 20/20) is not stated in the main text (a comparison on one sequence, the real-label 5+1 MLP, outside the claims of the main text).
\end{itemize}

\subsection{The 12 conditions of the real-label CNN; calibration and paired differences at the long budget}
\label{app:F4}
780 updates per task on the 5+1 CNN, evaluation seeds 2000--2009. Selection on calibration seeds 100--102; the main quantity is online on the late hard tasks (tasks 21, 23, 25, 27, 29). The calibrated selections and EDGE are given in the table below. Not all evaluated conditions use the same learning rate, and the differences of the selected values are part of this comparison.

\begin{center}\footnotesize\setlength{\tabcolsep}{3pt}
\begin{tabular}{@{}P{1.8cm}P{5.2cm}P{3.2cm}ll@{}}
\toprule
\textbf{Condition} & \textbf{Selected value} & \textbf{EDGE} & \textbf{late \%} & \textbf{test\%} \\
\midrule
ELU & α 3.6, lr 3e−4 & α (grid of 2 values) & 83.99 & 77.13 \\
SNAfrz1 & lr 3e−4 & — & 80.79 & 78.22 \\
SNA & lr 3e−4 & — & 80.01 & 77.98 \\
NEO & lr 1e−3 & lr & 79.42 & 76.80 \\
SN06 & lr 3e−4 & — & 79.88 & 78.05 \\
RSL & setting of Lillo \& Cheney, lr 1e−3 & lr & 78.23 & 76.65 \\
KKT1 & lr 3e−4 & — & 78.92 & 77.36 \\
LK & slope 0.8, lr 1e−3 & lr, slope & 78.39 & 77.17 \\
SN3 & lr 1e−3 & lr & 75.98 & 76.59 \\
L2I & λ 1e−2, lr 1e−3 & lr & 71.94 & 77.28 \\
R & lr 3e−4 & — & 48.38 & 65.66 \\
ReDo & threshold 0, period 1560, lr 3e−4 & — & 47.12 & 62.86 \\
\bottomrule
\end{tabular}\end{center}

The secondary family of 8 comparisons against SNAfrz1 is shown (§5 states only that, in level, ELU at 780 updates ranks highest, and does not cite the numbers of this table). It is a family separate from the 8 main comparisons of NEO. Registered intervals, unit pt.

\begin{center}\footnotesize\setlength{\tabcolsep}{3pt}
\begin{tabular}{@{}llll@{}}
\toprule
\textbf{Comparison with SNAfrz1} & \textbf{mean pt} & \textbf{interval pt} & \textbf{positive seeds} \\
\midrule
SNAfrz1 − ELU & −3.20 & [−4.01, −2.39] & 0/10 \\
SNAfrz1 − SN06 & +0.91 & [+0.10, +1.67] & 8/10 \\
SNAfrz1 − KKT1 & +1.87 & [+1.26, +2.63] & 10/10 \\
SNAfrz1 − LK & +2.40 & [+1.51, +3.19] & 10/10 \\
SNAfrz1 − RSL & +2.56 & [+1.60, +3.49] & 10/10 \\
SNAfrz1 − SN3 & +4.81 & [+4.12, +5.30] & 10/10 \\
SNAfrz1 − L2I & +8.85 & [+7.79, +9.94] & 10/10 \\
SNAfrz1 − ReDo & +33.67 & [+30.54, +37.05] & 10/10 \\
\bottomrule
\end{tabular}\end{center}

KKT1 − ELU at 7,800 updates (post hoc): mean paired difference over 10 seeds; 20,000 bootstrap resamples, resampling seed 1023; unadjusted 95\% interval.

\begin{center}\footnotesize\setlength{\tabcolsep}{3pt}
\begin{tabular}{@{}P{4.4cm}lllll@{}}
\toprule
\textbf{Measure} & \textbf{KKT1} & \textbf{ELU} & \textbf{mean paired difference} & \textbf{positive seeds} & \textbf{bootstrap 95\%} \\
\midrule
online, first 780 updates (\%) & 83.54 & 82.82 & +0.73 & 8/10 & [+0.21, +1.32] \\
online, whole task (\%) & 98.27 & 98.18 & +0.10 & 8/10 & [+0.02, +0.18] \\
test (\%) & 77.10 & 75.12 & +1.98 & 10/10 & [+1.41, +2.51] \\
\bottomrule
\end{tabular}\end{center}

This long-budget difference is not added to the confirmatory family of NEO at 780 updates. The verdict fixed in advance for the budget extension itself is in H.5.

Note: registered intervention comparison with 3 calibration seeds and 10 evaluation seeds. The paired difference at 7,800 updates is a post hoc analysis.

\section{Error-mean removal (all conditions of §4.5)}
\label{app:G}
\begin{itemize}
\item Network: the MLP on real-label 5+1 CIFAR (3072-100-100-100, std, Adam $10^{-4}$) in which the EMAs of the outputs of both hidden layers are subtracted, the biases are suppressed, and the learning rate of the output bias is multiplied by 100 (symbol CHB0). It is delayed temporarily only at the switch from task 2 to 3. The verdicts and rules were fixed before the runs (the sha256 of the verdict script is on the first line of the verdict file).
\item Condition symbols: K = nothing; New = network trained from scratch; Ob = oracle that aligns the output bias at the boundary; KF300 = the hidden layers frozen for only 300 updates; T1 = the output bias aligned across all classes at the boundary; T1F200 = T1 with a 200-update freeze; T2a/T2b = changed speed of the output bias; T3 = the output error passed back to the hidden layers is always $e_n-\bar e$ (all of $e_n$ flows to the readout and the output bias; error-mean removal); T3e = the boundary-only version of T3; C4 = reducing the same amount as the freeze in a different form; Bfirst300/Wfirst300 = freezing only the biases / weights for the first 300 updates. d\_on = difference in online on the next task (from New); r = fraction of the delay prevented.
\end{itemize}

Excerpt of the verdict output:

\begin{Verbatim}[fontsize=\scriptsize,breaklines=true,breakanywhere=true]
3.1 (CHB0 vs New; r den med[M_mid(Ob)-M_mid(K)] +0.1660)
   K          d_on -0.0746 pos  0  d_Mmid -0.1551  T0.9 3590/1980  r +0.000 (pos 0)  class NOT_EQUAL
   Ob         d_on -0.0048 pos  2  d_Mmid -0.0059  T0.9 1840/1980  r +1.000 (pos 10)  class EQUAL
   KF300      d_on -0.0094 pos  0  d_Mmid -0.0186  T0.9 1600/1980  r +0.879 (pos 10)  class NOT_EQUAL
   T1         d_on +0.0198 pos 10  d_Mmid +0.0586  T0.9 1220/1980  r +1.346 (pos 10)  class ABOVE
   T1F200     d_on +0.0177 pos 10  d_Mmid +0.0568  T0.9 1130/1980  r +1.263 (pos 10)  class ABOVE
   T2a        d_on -0.0109 pos  0  d_Mmid -0.0229  T0.9 1640/1980  r +0.887 (pos 10)  class NOT_EQUAL
   T2b        d_on -0.0131 pos  0  d_Mmid -0.0371  T0.9 1450/1980  r +0.748 (pos 10)  class NOT_EQUAL
   T3         d_on +0.0079 pos 10  d_Mmid +0.0348  T0.9 1270/1980  r +1.158 (pos 10)  class ABOVE
   T3e        d_on +0.0073 pos 10  d_Mmid +0.0300  T0.9 1270/1980  r +1.155 (pos 10)  class ABOVE
   C4         d_on -0.0579 pos  0  d_Mmid -0.1221  T0.9 2930/1980  r +0.285 (pos 10)  class NOT_EQUAL
   Bfirst300  d_on -0.0134 pos  0  d_Mmid -0.0324  T0.9 1460/1980  r +0.759 (pos 10)  class NOT_EQUAL
   Wfirst300  d_on -0.0823 pos  0  d_Mmid -0.1979  T0.9 3060/1980  r -0.255 (pos 2)  class NOT_EQUAL
3.2 den med[W3(KF300)-W3(K)] +0.2012 pos 10 holds=True f_B +0.964 (pos 10) f_W -0.012 (pos 4) class BIAS_SUFFICES | aligning the premise (bias) first suffices; the direction is not needed
3.3 T3 r +1.158 (pos 10) class PREVENT | without freezing, merely not passing the common component to the hidden layers largely prevents the wound => the wound is made by the push of the premise error that the hidden layers receive while the offset remains
3.3 T3e r +1.155 (pos 10) class PREVENT | without freezing, merely not passing the common component to the hidden layers largely prevents the wound => the wound is made by the push of the premise error that the hidden layers receive while the offset remains
3.4 r_C4 +0.285 (pos 10) r_KF300 +0.879 (pos 10) class FORM_NEEDED | reducing the same amount does not prevent it unless it takes the form of freezing
3.5 R T1      d_on vs K +0.0033 (pos 10)  class NO_HARM  | desc vs New +0.0092
3.5 R T1F200  d_on vs K -0.0046 (pos 3)  class NO_HARM  | desc vs New +0.0027
3.5 R T2b     d_on vs K +0.0028 (pos 7)  class NO_HARM  | desc vs New +0.0097
3.5 R T3      d_on vs K -0.0142 (pos 0)  class HARM  | desc vs New -0.0055
3.5 R C4      d_on vs K -0.0286 (pos 0)  class HARM  | desc vs New -0.0208
\end{Verbatim}

\begin{itemize}
\item The phrases ``not passing the common component to the hidden layers'' and ``push of the premise error'' in the 3.3 lines of the excerpt are the original words of the verdict file and correspond to ``removing, from the error backpropagated to the hidden layers, only the part equal to the minibatch mean of the output error'' of §4.5. What T3 removes is not all of the common error $S^{\rm rand}$ but only its first term $\bar J^\top\bar e$ (identity B.1 of Appendix~B; for the weights of the last hidden layer, $\langle v_i,\bar e\rangle\,\overline{\varphi'_ix}$).
\item \textbf{Correspondence with the classical method}: \citet{schraudolph1998centering} (the chapter in \emph{Neural Networks: Tricks of the Trade}; we read the text of the 1998 revision of technical report IDSIA-19-97), §2.2, eq.~(4), moves the weights other than the bias by the centered error $\check\delta_j=\delta_j-\langle\delta_j\rangle$, and eq.~(5) states that ``using the same update for the bias weight would stop the bias'', so the bias moves as usual (eq.~(3)); this form, which passes the mean error only to the bias weights, is called d.c.\ error shunting. §2.3, eq.~(8), states that the backpropagated centered error has mean 0 by itself. T3 applies this to backpropagation from the output error into the hidden layers, with two differences: (i) all errors $e_n$ also flow to the readout weights $W_3$, and (ii) the mean is the minibatch mean. In CHB0 the hidden biases are fixed at 0, so the treatment of the bias in eq.~(5) does not apply.
\item {[Inference]} One reading of the harm of T3 in the unmodified network R: T3 in R also passes the centered error to the hidden biases $b_1,b_2$ (checked in our records), close to the form that Schraudolph's eq.~(5) forbids. The amount reaching the hidden biases is, however, $\sum_n\varphi'(z_n)\odot W_3^\top(e_n-\bar e)$, which is not 0 because the slope differs from input to input, so the effect is weaker than the ``stop'' of eq.~(5). A comparison that passes the centered error only to the hidden weights was not run.
\item T1 and T1F200 in the unmodified network (output-bias learning rate as usual; 30 tasks; symbol R) (L = gap to scratch; dead2130 = fraction of dead units in layer 2 over tasks 21--30; medians of [arm − record]). Excerpt:
\end{itemize}

\begin{Verbatim}[fontsize=\scriptsize,breaklines=true,breakanywhere=true]
T1: class STILL | neither moves G1 10/10 fresh rows 10 removed none diverged none
   L: n 10 k(n) 9 median(arm-rec) 0.0048 same_sign 6 gate 0.0300 moved False
   dead2130: n 10 k(n) 9 median(arm-rec) -0.0060 same_sign 7 gate 0.0500 moved False
   width30: n 10 k(n) 9 median(arm-rec) -1.6499 same_sign 9 gate 6.0302 moved False
   width moved False
   rho: L n 10 0.0048/-0.0482 = -0.099; dead2130 n 10 -0.0060/-0.1245 = 0.048
   switch med online(arm-rec) t=3..29: 0.0002 -0.0007 -0.0013 -0.0015 0.0008 0.0016 0.0053 0.0037 0.0067 0.0105 -0.0003 0.0057 0.0044 -0.0048
   b3 gap before align (median std / range over classes) t=2..: 2:0.059/0.231 3:0.007/0.063 4:0.020/0.156 5:0.007/0.077 6:0.023/0.183 7:0.008/0.088 8:0.025/0.187 9:0.009/0.094 10:0.027/0.204 11:0.009/0.091 12:0.030/0.236 13:0.009/0.094 14:0.031/0.224 15:0.010/0.094 16:0.031/0.235 17:0.010/0.102 18:0.032/0.250 19:0.011/0.100 20:0.034/0.261 21:0.011/0.103 22:0.037/0.261 23:0.011/0.105 24:0.038/0.301 25:0.012/0.117 26:0.040/0.298 27:0.013/0.116 28:0.045/0.319 29:0.013/0.113 30:0.051/0.353
   n(train>=0.95) t=1..30: 10 10 10 10 10 10 10 10 10 10 10 10 10 10 10 10 10 10 10 10 10 10 10 10 10 10 10 10 10 10
T1F200: class MOVES | the cure reduces the ever-growing loss and the death G1 10/10 fresh rows 10 removed none diverged none
   L: n 10 k(n) 9 median(arm-rec) -0.0785 same_sign 10 gate 0.0300 moved True
   dead2130: n 10 k(n) 9 median(arm-rec) -0.1505 same_sign 10 gate 0.0500 moved True
   width30: n 10 k(n) 9 median(arm-rec) -13.2861 same_sign 10 gate 6.0302 moved True
   width moved True
rho: L n 10 -0.0785/-0.0482 = 1.628; dead2130 n 10 -0.1505/-0.1245 = 1.209 -> most of the effect of a faster output bias is obtained by the alignment at the boundary alone
\end{Verbatim}

\begin{itemize}
\item Comparison of the network with both paths blocked (CHB0; output-bias learning rate ×100) and the unmodified network [post hoc, descriptive]: on the same machine and with the same setting, the blocked network was 6--12 points below the unmodified network on every hard task from task 3 to 29, and at task 29 it trailed its own network trained from scratch by 6--17 points in all 10 seeds. In shape, it drops by one step at task 3 and is nearly flat afterwards, whereas the unmodified network keeps declining. This is the source of the limiting sentence of §4.2.
\item T1F200 of the excerpt above (in the unmodified network, the output bias aligned at the boundary and the hidden layers frozen for 200 updates) has L −0.0785 and dead2130 −0.1505 (4-digit display), corresponding to a 7.9-point smaller gap to scratch and about 15 points fewer dead units in layer 2 (rounded from the full digits −0.078539663 and −0.1504999996 of median\_diff in the T1F200 rows of the verdict json; 15.05 lies exactly on the rounding boundary, hence ``about 15''). The main text does not use these numbers.
\item Reading [§4.5]: error-mean removal prevents the temporary delay of the centered network (r 1.16, 10/10, PREVENT) but is harmful in the unmodified network (T3 in R −1.4 points, 0/10, HARM). On a 20-task sequence that adds 5 classes in turn (ci5, 20 seeds), the removal of the mean part works on tasks 2--6, and the unmodified network catches up on tasks 16--20. On the standard sequences it is harmful (J.10, §4.5).
\end{itemize}

\section{CNN}
\label{app:H}
Main text: §5.

\subsection{Readout alignment and layer-wise switch gradients of the random-label CNN}
$\kappa$ at the saved task ends of 10 seeds of leaky ReLU, and the expected gradient toward independent new labels. own = own weights, full = all paths; positive is the lowering side; the values are seed medians of fractions of live units. The three values of the layer table are tasks 1/10/30; the absolute share of the upstream for f2 is .97--1.00. The 200 updates are actual Adam updates; all tables are descriptive without intervals.

The ratio of $\kappa$ to its value at the end of task 1 went 1.46 → 7.05 from task 2 to the last task, and $\cos^2$ went .75 → .63 from task 1 to the last task.

\begin{center}\footnotesize\setlength{\tabcolsep}{3pt}
\begin{tabular}{@{}llllll@{}}
\toprule
\textbf{Condition} & \textbf{f2 own} & \textbf{f2 full} & \textbf{f1 full} & \textbf{c2 full} & \textbf{c1 full} \\
\midrule
LR & 1.00 / 1.00 / 0.99 & 0.83 / 0.82 / 0.75 & 0.72 / 0.92 / 0.97 & 0.74 / 1.00 / 1.00 & 0.50 / 0.10 / 0.40 \\
\bottomrule
\end{tabular}\end{center}

\begin{center}\footnotesize\setlength{\tabcolsep}{3pt}
\begin{tabular}{@{}lll@{}}
\toprule
\textbf{LR task} & \textbf{shifted negative in 200 updates} & \textbf{within-task return ratio} \\
\midrule
1 & 0.92 & 0.32 \\
3 & 0.89 & 0.60 \\
5 & 0.84 & 0.90 \\
10 & 0.98 & 1.02 \\
30 & 1.00 & 1.25 \\
\bottomrule
\end{tabular}\end{center}

The gradient of c1 is not determined by the sum of Eq.~3 alone. Splitting the signal $H$ returning from the next layers into a constant mode and the rest, $G_j=\sum a_{nq} H_{nq,j}=(\sum a_{nq})\bar H_j+\sum a_{nq}(H_{nq,j}-\bar H_j)$ ($a_{nq}=\gamma K1$). This is a decomposition of the downstream signal into its mean and its variation, not a matter of the input mean only. At tasks 5, 10 and 20 of leaky ReLU, the absolute share of the constant mode was .70--.73, and its sign agreement with the full gradient .89--.95. While the constant mode lifted in fractions .62--.99, the remainder was on the lowering side in .86--1.00. The cancellation differs by unit and window, and no uniform sign of the negative shift is imposed on c1.

Note: diagnostics of saved trajectories with 10 seeds; predictions recorded before the computation. c1 is a decomposition of fixed states.

\subsection{Real-label CNN: alignment with the current task and what remains of it}
Saved states of 4 conditions × 3 budgets, 10 seeds each; 467 is a number of states, not of independent seeds. Restricting the output layer to the rows of the 5 classes of the current hard task, $\kappa$ and the other quantities of C.1 are measured on the images and labels of the same task, and the label permutation and the lower bound of fixed states are checked. The tables are descriptive without intervals.

\begin{center}\footnotesize\setlength{\tabcolsep}{3pt}
\begin{tabular}{@{}llll@{}}
\toprule
\textbf{Condition} & \textbf{Updates} & \textbf{states with κ>0} & \textbf{cos², b1→29} \\
\midrule
ReLU & 780 & 40/40 & 0.22 → 0.06 \\
ReLU & 2340 & 37/37 & 0.34 → 0.11 \\
ReLU & 7800 & 30/30 & 0.40 → 0.10 \\
ELU & 780 & 40/40 & 0.60 → 0.38 \\
ELU & 2340 & 40/40 & 0.74 → 0.34 \\
ELU & 7800 & 40/40 & 0.78 → 0.30 \\
KKT1 & 780 & 40/40 & 0.27 → 0.25 \\
KKT1 & 2340 & 40/40 & 0.28 → 0.29 \\
KKT1 & 7800 & 40/40 & 0.26 → 0.26 \\
SNAfrz1 & 780 & 40/40 & 0.28 → 0.31 \\
SNAfrz1 & 2340 & 40/40 & 0.24 → 0.28 \\
SNAfrz1 & 7800 & 40/40 & 0.24 → 0.26 \\
\bottomrule
\end{tabular}\end{center}

In all 12 conditions, the number of cases with the permutation maximum equals the number with $\kappa>0$, and the violations of the lower bound are 0 in all conditions.

\begin{center}\footnotesize\setlength{\tabcolsep}{3pt}
\begin{tabular}{@{}P{2.8cm}P{3.2cm}P{2.8cm}P{4.0cm}P{3.0cm}@{}}
\toprule
\textbf{Quantity} & \textbf{before the task (boundary b−1)} & \textbf{end of the task (boundary b)} & \textbf{previous hard task (s → end of next hard)} & \textbf{8 or more tasks later} \\
\midrule
mean cos² & 0.000–0.004 & 0.06–0.78 & 0.001–0.011 & 0.000–0.004 \\
seeds with κ > 0 & half (177/355) & all & 0.77 (265/345) & half \\
permutation maximum & 13/470 (chance 3.9) & 467/467 & 0–4/10 & 0–2/10 \\
|Q\_pre|/Q\_post & 3 live conditions ≤ 0.07, ReLU ≤ 0.30 & — & median κ ratio 0.06 & 0.00–0.06 \\
\bottomrule
\end{tabular}\end{center}

Label term of the switch gradient (same saved states; verdicts and predictions registered before the implementation). On all images of the next task, the one-step expected gradient through the own weight and bias paths of FC2 was split as $S^{\rm real}=S^{\rm rand}-L$ ($S^{\rm rand}$ is the model with the labels replaced by the empirical distribution of the task, $L$ is the label term, and $S>0$ is the lowering side). Ranges of 10-seed medians of within-seed boundary averages for hard switches (boundaries 1, 9, 19 → task $b+2$).

\begin{center}\footnotesize\setlength{\tabcolsep}{3pt}
\begin{tabular}{@{}P{3.0cm}P{2.2cm}P{3.0cm}P{1.8cm}P{3.2cm}P{1.6cm}@{}}
\toprule
\textbf{Condition} & \textbf{sign agreement (agree)} & \textbf{ratio\_L = median |L|/|S\textasciicircum{}rand|} & \textbf{fraction L<0} & \textbf{lowering side real / rand} & \textbf{live units} \\
\midrule
ReLU (780, 2340, 7800) & 0.95–0.96 & 0.07–0.09 & 0.48–0.51 & 0.82–0.83 / 0.82–0.83 & 47–62 \\
ELU α3.6 & 0.97–0.98 & 0.07–0.08 & 0.50–0.52 & 0.50–0.55 / 0.50–0.55 & 100 \\
KKT1 & 0.96–0.97 & 0.09–0.10 & 0.50 & 0.49–0.53 / 0.50–0.53 & 100 \\
SNAfrz1 & 0.96–0.97 & 0.10–0.12 & 0.47–0.51 & 0.51–0.53 / 0.52 & 100 \\
\bottomrule
\end{tabular}\end{center}

Verdicts: J1 RAND\_PREDICTS\_SIGN (agree $\ge0.85$ in 9 of the 9 cells other than ReLU), J2 LABEL\_TERM\_SMALL (ratio\_L $\le0.30$ in 9/9). At easy switches (1 class, $L\equiv0$), the direction of the term of the unseen-class rows is split (class\_sink 0.48--0.54; ReLU 0.55--0.71), J3 MIXED, and the size of that term is comparable to the term of the uniform model (ratio 0.75--1.25). Checks: reproduction of the forward pass (480 states, relative error 4e−7), the identity (7.8e−15), agreement with autograd (7.6e−8) and invariance under centering of the output layer PASS; the label-permuted control satisfies $L\approx0$ in all 356 states, but the improvement of agreement is partial, 323/356. The excluded states are the same 13 as in the earlier analysis (collapsed ReLU). The predictions (by the analyst: J1 0.60, J2 0.50, J3 0.60) hit for J1 and J2 and missed for J3. This is the one-step expected gradient of fixed states, not the actual Adam update, the upstream terms, or the gradient within a task.

Note: post hoc analysis of saved states, 10 seeds each, descriptive; the verdict rules and the predictions were recorded before the analysis.

\subsection{Prevention by centering and bias suppression (B) in the CNN}
C subtracts the input mean per color channel; H′ subtracts the EMA mean per channel × position after pooling (coefficient .01, no gradient) and the per-unit means of the fully connected layers; CH′ uses both. The convolutional part of H/CH subtracts only channel means and leaves the means of spatial positions. B, after the Adam update, projects the biases of c1/c2/f1/f2 into ±.2 of the standard deviation of the bias-free preactivation of the same minibatch. Weights, the output layer and the Adam moments are not touched directly. RC CNN, seeds 10--19; the main quantity is online over the late tasks 21--30.

CH/CH′ of the first run with TF32=true are seed medians, and CH′/CH′+B of the separate run with TF32=false are seed means. Differences across settings are not taken as one paired difference.

\begin{center}\footnotesize\setlength{\tabcolsep}{3pt}
\begin{tabular}{@{}llll@{}}
\toprule
\textbf{Condition} & \textbf{TF32} & \textbf{late online (\%), aggregate} & \textbf{seeds at ≥50\%} \\
\midrule
CH & true & 11.5 (10.3–26.8), median & 0/10 \\
CH′ & true & 17.8 (11.1–91.7), median & 1/10 \\
CH′ & false & 49.518, mean & 6/10 \\
CH′+B & false & 94.155, mean & 10/10 \\
\bottomrule
\end{tabular}\end{center}

CH′+B − CH′ with TF32=false is +44.637 pt, 95\% paired $t$ interval [23.095, 66.179], positive in 10/10 seeds.

In the TF32=true run, the bias share of CH′ is 1.00 in all 4 layers. $\Delta b$ / fraction of dead units: c1 −1.04/.78, c2 −10.5/.88, f1 −6.5/.77, f2 −6.0/.86 (seed medians of unit medians). In the TF32=false run, CH′ collapsed in 4/10. Prevention from the start and adding B later (H.4) are separate.

Note: registered intervention, 10 seeds. The TF32=true condition and the layer-wise post hoc diagnostics are in our records.

\subsection{Swapping the derivatives of all layers of the CNN, and adding B later}
CH′/CH′+B at the end of task 30 with TF32=false serve as recipients/donors, and the compensated fixed offset of E.1 is inserted into all 4 layers. Self-donation CH\_self/CHB\_self and swaps CH\_swap/CHB\_swap (CH\_ and CHB\_ denote the CH′ and CH′+B networks), all 4 conditions, 10 seeds each, 30,000 updates of the next task. The activities and predictions at the branch are matched, and the main quantity is online over the first 3000 updates. With rescue = swap − self and reverse = self − swap as positive effects, the 2 main comparisons use two-sided exact sign tests with Holm correction, and the intervals are seed-bootstrap 95\%.

\begin{center}\footnotesize\setlength{\tabcolsep}{3pt}
\begin{tabular}{@{}lllll@{}}
\toprule
\textbf{Comparison} & \textbf{mean difference for the verdict (pt)} & \textbf{95\% seed bootstrap CI(pt)} & \textbf{pos/neg/tie} & \textbf{Holm p} \\
\midrule
rescue & +43.58292 & +35.38182, +50.87208 & 10/0/0 & 0.00390625 \\
reverse & +11.46833 & +3.13392, +16.80855 & 9/1/0 & 0.021484375 \\
\bottomrule
\end{tabular}\end{center}

The main and final windows are shown. The final window is descriptive and does not replace the main verdict. Improvement of accuracy and worsening of CE are distinguished.

\begin{center}\footnotesize\setlength{\tabcolsep}{3pt}
\begin{tabular}{@{}lllll@{}}
\toprule
\textbf{Condition} & \textbf{first3000 accuracy (\%)} & \textbf{last3000 accuracy (\%)} & \textbf{first3000 CE} & \textbf{last3000 CE} \\
\midrule
CH\_self & 14.63562 & 37.44771 & 2.386382 & 1.644639 \\
CH\_swap & 58.21854 & 99.20042 & 732.470452 & 8.699800 \\
CHB\_self & 71.41917 & 98.58313 & 7563.076352 & 168.687711 \\
CHB\_swap & 59.95083 & 98.74083 & 14223.842597 & 231.909957 \\
\bottomrule
\end{tabular}\end{center}

The effective gate is measured at $z+d$ during learning; as an example, the course of f2:

\begin{center}\footnotesize\setlength{\tabcolsep}{3pt}
\begin{tabular}{@{}lllll@{}}
\toprule
\textbf{Updates} & \textbf{CH\_self} & \textbf{CH\_swap} & \textbf{CHB\_self} & \textbf{CHB\_swap} \\
\midrule
0 & 0.0434775 & 0.48021 & 0.48021 & 0.0434775 \\
100 & 0.0170183 & 0.475371 & 0.404443 & 0.64046 \\
30000 & 0.03804 & 0.472648 & 0.481019 & 0.504674 \\
\bottomrule
\end{tabular}\end{center}

The comparison that adds B later branches CH\_continue/CH\_addB from the full state of the same CH′ at the end of task 30. No artificial field; the first B comes after the first Adam/EMA. 10 seeds × 30,000 updates each. The 2 registered comparisons of first-3000 accuracy/CE are Holm-corrected; the other windows below are descriptive.

\begin{center}\footnotesize\setlength{\tabcolsep}{3pt}
\begin{tabular}{@{}lll@{}}
\toprule
\textbf{Window} & \textbf{accuracy control→B (\%) or difference pt} & \textbf{CE control→B} \\
\midrule
first3000 & 14.636→17.567, +2.931 & 2.386→6082.710 \\
last3000 & 37.448→29.548, −7.899 & 1.645→12922.019 \\
\bottomrule
\end{tabular}\end{center}

Accuracy difference over the first 3000: 95\% bootstrap [−1.178, +6.825] pt, 7 improved / 3 worse, Holm $p=.34375$, INCONCLUSIVE. CE worse in 10/10, $p=.00390625$, WORSENS. B was active: the mean derivatives at update 1 of c1/c2/f1/f2 = .441/.876/.971/.911, and the f2 scale went 6.26 → 114934. We do not identify the scale alone as the cause of the failure or the derivative alone as the mediator.

Note: registered branch intervention, 10 seeds. Adding B later is a separate registration.

\subsection{Budget extension and scale cap in the real-label CNN}
\label{app:H5}
We compare 780/2,340/7,800 updates on the same task sequence with the learning rate selected per condition. Evaluation seeds 2000--2009. Shown are the speed over the first 780 updates (family 1, pt) and the FC2 scale (family 4, difference of log scales), which the main text uses; Bonferroni bootstrap simultaneous 95\% with 8 comparisons per family. The separate families for online over the full budget and for the gap to fresh are in our records. State quantities are descriptive.

\begin{center}\footnotesize\setlength{\tabcolsep}{3pt}
\begin{tabular}{@{}P{4.6cm}lllll@{}}
\toprule
\textbf{Family} & \textbf{Condition} & \textbf{2340 − 780} & \textbf{Verdict} & \textbf{7800 − 780} & \textbf{Verdict} \\
\midrule
family 1: online, first 780 updates & ReLU & −13.12 & \texttt{LOWER} & −29.50 & \texttt{LOWER} \\
 & ELU & +0.57 & \texttt{INCONCLUSIVE} & −1.17 & \texttt{LOWER} \\
 & KKT1 & +5.50 & \texttt{HIGHER} & +4.62 & \texttt{HIGHER} \\
 & SNAfrz1 & +3.37 & \texttt{HIGHER} & +1.01 & \texttt{INCONCLUSIVE} \\
family 4: FC2 scale at boundary 30 & all 4 conditions & +0.75 to +1.22 & \texttt{HIGHER} & +1.55 to +2.14 & \texttt{HIGHER} \\
\bottomrule
\end{tabular}\end{center}

\begin{center}\footnotesize\setlength{\tabcolsep}{3pt}
\begin{tabular}{@{}llllll@{}}
\toprule
\textbf{Condition} & \textbf{Budget} & \textbf{scale, boundary 2} & \textbf{scale, boundary 30} & \textbf{φ′, boundary 30} & \textbf{death, boundary 30} \\
\midrule
ReLU & 780 & 1.88 & 10.12 & 0.00 & 0.85 \\
ReLU & 7800 & 2.70 & 86.97 & 0.00 & 0.94 \\
ELU & 780 & 1.09 & 6.69 & 0.65 & 0 \\
ELU & 2340 & 1.50 & 19.35 & 0.39 & 0 \\
ELU & 7800 & 2.06 & 31.59 & 0.27 & 0 \\
KKT1 & 780 & 1.94 & 5.17 & 0.84 & 0 \\
KKT1 & 7800 & 6.62 & 25.07 & 0.92 & 0 \\
SNAfrz1 & 780 & 1.65 & 4.40 & 0.96 & 0 \\
SNAfrz1 & 7800 & 31.90 & 35.26 & 0.94 & 0 \\
\bottomrule
\end{tabular}\end{center}

The scale cap pairs ReLU/ELU with a new control (natural) at \textbf{7,800 updates/task}, 10 seeds each. The per-unit preactivation scale at the end of task 1 is the cap, and after Adam/EMA the W and bias of the rows exceeding it are shrunk by the same factor. The output layer and Adam are not changed directly. The cap uses the EMA of training minibatches and is not an exact cap of the scale on a fixed probe. The main quantity is the \textbf{first 780 updates} of the late hard tasks; this is not a cap on the short 780-update run. Exact sign tests with Holm correction for the 2 main comparisons.

\begin{center}\footnotesize\setlength{\tabcolsep}{3pt}
\begin{tabular}{@{}lllllll@{}}
\toprule
\textbf{Activation} & \textbf{natural} & \textbf{cap} & \textbf{cap−natural} & \textbf{pos/neg/0} & \textbf{Holm p} & \textbf{Verdict} \\
\midrule
ReLU & 22.184 & 33.823 & 11.639 & 5/4/1 & 1.000000 & INCONCLUSIVE \\
ELU α=3.6 & 83.657 & 78.547 & -5.111 & 0/10/0 & 0.003906 & WORSENS \\
\bottomrule
\end{tabular}\end{center}

f2 at the end of task 30 for all 4 conditions. The scale, standardized mean and derivative are seed means of unit medians; death is the seed mean of unit fractions, with the definition of death in A.3.

\begin{center}\footnotesize\setlength{\tabcolsep}{3pt}
\begin{tabular}{@{}lllllll@{}}
\toprule
\textbf{Activation} & \textbf{Condition} & \textbf{scale s} & \textbf{standardized mean r} & \textbf{mean φ′} & \textbf{dead} & \textbf{low-derivative} \\
\midrule
ReLU & natural & 61.0502 & undefined & 0.0000 & 0.9560 & 0.9910 \\
ReLU & cap & 1.4066 & undefined & 0.0003 & 0.7620 & 0.8440 \\
ELU α=3.6 & natural & 33.9526 & -0.6880 & 0.2914 & 0.0000 & 0.1950 \\
ELU α=3.6 & cap & 1.3859 & 0.1136 & 0.9988 & 0.0000 & 0.1480 \\
\bottomrule
\end{tabular}\end{center}

For ELU, late end-of-task test accuracy went 75.372 → 75.616\% (+.244 pt), and train accuracy was 100\% for both (descriptive). The new control (natural) and the earlier budget extension are separate runs.

Note: registered budget comparison, 10 seeds each. The cap is a separately registered intervention with 10 seeds.

\setcounter{section}{9}
\section{Standard task sequences}
\label{app:J}
\subsection{Setup and differences from the description of Kumar et al.}
\label{app:J1}
\begin{itemize}
\item \textbf{Continual ImageNet}: the setting of §4.2 and Table~1 of App.~A.1 of \citet{kumar2024maintaining} (1 task = 2 classes × 600 images = 1,200 images; minibatch 100; 10 epochs = 120 updates; 500 tasks; each class appears in only one task; a 2-output output layer shared across all tasks; Adam defaults; PyTorch default initialization). The data are the 32×32 images released by \citet{dohare2024loss} (1,000 classes × 700 images; the first 600 images of each class for training and 100 for testing; distributed as classes.zip on Google Drive, 8.5~GB, as pointed to by lop/imagenet/README.md of the repository of Dohare et al.; the code is MIT-licensed, the distribution states no terms of use for the data, and we read the ImageNet terms [non-commercial research and education] as applying; we use the data only within this study and do not redistribute them). The float64 continuous values were mapped to float16 (the maximum rounding difference is recorded). Loading uses only np.load(allow\_pickle=False).
\item \textbf{Differences from the description of Kumar et al.}: (i) the network is the CNN of §5 (2 convolutional layers of 5×5 [16 channels], max pooling after each, fully connected 1024→100→100) with size-preserving padding 2. Kumar et al.\ state the same numbers of layers and widths but not the padding (with padding 0 the input to the fully connected layer would be 400 and the number of parameters different [inference]). (ii) The inputs were further standardized by per-channel constants of all training images (the same for all methods). (iii) This is a setting different from the Continual ImageNet of Dohare et al.\ (SGD; 3 convolutional + 3 fully connected layers; 2,000--5,000 tasks; per-task test accuracy).
\item \textbf{Permuted MNIST}: 1 task = 10,000 images drawn from the MNIST training set preserving the class proportions, with a random permutation of the pixels; minibatch 16; 1 epoch = 625 updates; 500 tasks; MLP 784-100-100-10.
\item \textbf{Split CIFAR-100}: a reduced ResNet-18 (with batch normalization); 20 tasks × 5 classes; 780 updates; 10 seeds (seeds 0--4 plus the added 5--9).
\item \textbf{Definitions of the added methods}: L2 = add $2\lambda\theta$ to the gradient. Layer Norm = normalize the preactivation of each hidden layer per example before the activation (100 units for fully connected layers; for convolutions, the whole channel × position = GroupNorm with 1 group; gain and bias learned). Kumar et al.\ do not state the placement, so we placed it after the linear map and before the activation. Concat ReLU = remove units (channels for convolutions) of each hidden layer in the ratio of Kumar et al.\ (9\% on Permuted MNIST and 27\% on Continual ImageNet removed = width 100 → 91, 100 → 73; convolutions 16 → 12) and pass [relu($z$), relu($-z$)] to the next layer (parameters at least those of the base network). Scale cap (Continual ImageNet) = with the per-unit preactivation standard deviation at the end of task 1 as the cap, after each Adam update the incoming rows and bias of the units exceeding it are shrunk by the same factor (App.~H.5). The input centering of centering + bias suppression subtracts a per-feature EMA of the input (coefficient 0.01) instead of a fixed mean, because the input mean changes with every task (unlike the fixed-mean operation C of App.~A.5).
\item \textbf{Differences of the grids} (from those of Kumar et al.): $\lambda$ of L2 and L2 Init in $\{10^{-2},10^{-3},10^{-4}\}$ (Kumar et al.\ also $10^{-5}$); S\&P 3 × 3 (Kumar et al.\ 4 × 4); replacement rate of CBP in $\{10^{-1},\ldots,10^{-5}\}$ (no $10^{-6}$); threshold of ReDo in \{0, 0.025, 0.1\} (Kumar et al.\ \{0, 0.01, 0.1\}). \textbf{The learning rate is \{0.0001, 0.001, 0.003, 0.01\} for all methods} (the Adam grid of Kumar et al., $\{10^{-3},10^{-4}\}$ \citep[§5]{kumar2024maintaining}, plus the 2 upper points; $3\times10^{-2}$ is added once only for methods that select $10^{-2}$). The strengths of our interventions are 3 points, the current value and ratio 2 on either side: width of the bias projection $c\in\{0.1, 0.2, 0.4\}$ (centering + bias suppression on Continual ImageNet), shrink rate of the second-layer bias $\lambda\in\{0.08565, 0.1713, 0.3426\}$ (centering + bias suppression on Permuted MNIST), a0 of kunekune in \{0.3, 0.6, 1.2\}, and multiplier of the scale cap $k\in\{0.5, 1, 2\}$. The strengths were combined with all learning rates.
\end{itemize}

\subsection{Rules fixed before the runs}
\label{app:J2}
\begin{itemize}
\item Comparison at learning rates \{0.001, 0.0001\} (runs A and B): the measures, the loss criterion (mean gap to scratch of at least +0.5 points with at least 9 seeds positive), the pairwise criterion, the handling of divergence and the replication control were written in the design record, whose sha256 was recorded before run B started.
\item Reselection of 4 learning rates and strengths (run C): the same measures and criteria (primary = gap to scratch; secondary = total average online), the selection rule, the rule for adding $3\times10^{-2}$, the handling of divergence, the replication control and the wording rules for the sentences of §6.3 were written in the design record, and the sha256 of the design and scripts was recorded before run C started. The sentences of §6.3 are the output of the rules (the decisions section of the verdict output), and no part was left undecided by the rules. Regardless of the results, no ranking sentence is placed in the abstract (decided before the run).
\end{itemize}

\subsection{Tables of all methods (evaluation seeds 0--9)}
\label{app:J3}
The numbers in the tables were computed from the raw records of each seed and rounded once. All places where the rounded display differs by 0.1 from the rounded 4-digit values of the verdict record are listed in our build log.

Settings changed by the reselection (run C) and settings selected at grid edges (§6.3): on Continual ImageNet, the strength of the scale cap changed to twice the current value (the upper edge of the strength grid), and the other methods kept their previous selections. Selected at grid edges were the learning rates of the unmodified network and Layer Norm (0.0001, lower edge), the noise size of S\&P (0.01, upper edge), the period of ReDo (120 updates, short edge) and the strength of the scale cap. On Permuted MNIST, centering + bias suppression changed from learning rate 0.001 to 0.003 and its second-layer bias decay to half the current value (lower edge), and the other methods kept their previous selections. Selected at grid edges were the noise size of S\&P (0.01, upper edge), the threshold (0.1, upper edge) and period (625 updates, short edge) of ReDo, and the strength of centering + bias suppression. No method selected learning rate 0.01, so $3\times10^{-2}$ was never added.

The numbers at learning rate 0.001 in §6.2 (unmodified network 28.2 points; centering + bias suppression −7.8 and 36.0 points above the unmodified network; kunekune −2.1 and 30.3 points; all with the primary measure of §6.1 = mean of tasks 400 and 500) are values of the first run (run A) (the gap of the unmodified network at task 500 alone is +30.7 in the same rows), and they differ from the ``rerun'' row of centering + bias suppression in the table below (run C) only by the replication difference (App.~J.5). The comparison of §6.3 relies on the table below (the selected setting prefers the newer run).

The Loss column is decided by the criterion of J.2 (mean gap to scratch of at least +0.5 points with at least 9 of 10 seeds positive), not by the size of the mean alone. For example, the scale cap on Continual ImageNet at the reselected strength (twice the current value, run C) has a mean of +10.6 but only 8/10 positive seeds and is therefore ``no'', while at the current strength (run B; the +6.8 of §6.2) 10/10 are positive and it is ``yes''.

\textbf{Continual ImageNet}

\begin{center}\scriptsize\setlength{\tabcolsep}{3pt}
\begin{tabular}{@{}P{2.0cm}P{2.7cm}P{1.1cm}P{1.1cm}P{1.0cm}P{2.5cm}P{0.9cm}P{0.8cm}P{2.3cm}P{0.6cm}@{}}
\toprule
\textbf{Method} & \textbf{Selected setting} & \textbf{Grid edge} & \textbf{Run} & \textbf{Total avg. online} & \textbf{Gap to scratch (tasks 400, 500; points; [95\% t]; pos/neg seeds)} & \textbf{Late window} & \textbf{task 1} & \textbf{Gap at tasks 100, 200, 300, 400, 500} & \textbf{Loss} \\
\midrule
L2 Init & \texttt{l2i|\allowbreak{}lr=0.001,\allowbreak{}l2init=0.001} & interior & run C (rerun) & 89.3\% & −7.6 [−9.2, −6.1] +0/−10 & 90.2\% & 84.3\% & −6.5 −5.8 −7.9 −6.9 −8.4 & no \\
L2 & \texttt{l2|\allowbreak{}lr=0.001,\allowbreak{}l2zero=0.001} & interior & run B & 87.2\% & −6.9 [−8.8, −5.0] +0/−10 & 88.7\% & 84.2\% & −2.8 −3.8 −5.8 −5.8 −8.0 & no \\
S\&P & \texttt{snp|\allowbreak{}lr=0.001,\allowbreak{}snp\_eps=0.0001,\allowbreak{}snp\_sigma=0.01} & edge: snp\_sigma & run A & 83.1\% & −2.5 [−3.9, −1.2] +2/−8 & 83.4\% & 84.3\% & −0.2 +0.3 −1.1 −2.1 −3.0 & no \\
CBP & \texttt{cbp|\allowbreak{}lr=0.001,\allowbreak{}cbp\_rho=0.001} & interior & run A & 85.4\% & −3.9 [−5.9, −1.8] +1/−9 & 85.6\% & 84.4\% & −1.8 −1.3 −3.1 −2.7 −5.0 & no \\
ReDo & \texttt{redo|\allowbreak{}lr=0.001,\allowbreak{}redo\_tau=0.025,\allowbreak{}redo\_period=120} & edge: redo\_period & run A & 86.1\% & −4.4 [−6.0, −2.8] +0/−10 & 85.8\% & 84.4\% & −3.3 −2.5 −3.6 −4.8 −4.0 & no \\
Layer Norm & \texttt{ln|\allowbreak{}lr=0.0001} & edge: lr & run B & 68.5\% & +17.1 [+13.5, +20.7] +10/−0 & 63.9\% & 83.4\% & +9.9 +11.2 +19.9 +16.3 +17.8 & yes \\
Concat ReLU & \texttt{crelu|\allowbreak{}lr=0.001} & interior & run B & 88.4\% & −5.3 [−6.4, −4.2] +0/−10 & 88.5\% & 85.3\% & −4.6 −3.6 −4.9 −5.5 −5.1 & no \\
unmodified (selected) & \texttt{none|\allowbreak{}lr=0.0001} & edge: lr & run A & 75.0\% & −4.1 [−6.9, −1.2] +2/−8 & 75.1\% & 74.4\% & −1.3 −1.4 −2.6 −4.8 −3.3 & no \\
centering + bias suppression & \texttt{block|\allowbreak{}lr=0.001} & interior & run C (rerun) & 91.7\% & −7.7 [−9.1, −6.3] +0/−10 & 92.1\% & 85.8\% & −6.0 −7.0 −6.8 −7.7 −7.7 & no \\
kunekune & \texttt{kkt1|\allowbreak{}lr=0.001} & interior & run A & 90.3\% & −2.1 [−3.1, −1.2] +0/−10 & 90.2\% & 90.0\% & −1.9 −1.1 −2.0 −2.3 −2.0 & no \\
scale cap & \texttt{wcap|\allowbreak{}lr=0.001,\allowbreak{}wcap\_k=2} & edge: wcap\_k & run C & 77.6\% & +10.6 [+2.2, +19.0] +8/−2 & 74.8\% & 84.4\% & +4.0 +5.4 +1.9 +7.8 +13.4 & no \\
centering + bias suppression + L2 Init & \texttt{combo|\allowbreak{}lr=0.001,\allowbreak{}l2init=0.001} & — & run B & 93.7\% & −9.1 [−10.8, −7.3] +0/−10 & 93.8\% & 85.9\% & −9.8 −9.1 −8.6 −9.1 −9.1 & no \\
unmodified (0.001) & \texttt{none|\allowbreak{}lr=0.001} & — & run A & 66.3\% & +28.2 [+21.2, +35.2] +10/−0 & 54.9\% & 84.4\% & +1.2 +11.6 +27.3 +25.7 +30.7 & yes \\
centering + bias suppression (0.001, current strength) & \texttt{block|\allowbreak{}lr=0.001} & — & run C (rerun) & 91.7\% & −7.7 [−9.1, −6.3] +0/−10 & 92.1\% & 85.8\% & −6.0 −7.0 −6.8 −7.7 −7.7 & no \\
kunekune (0.001, current strength) & \texttt{kkt1|\allowbreak{}lr=0.001} & — & run A & 90.3\% & −2.1 [−3.1, −1.2] +0/−10 & 90.2\% & 90.0\% & −1.9 −1.1 −2.0 −2.3 −2.0 & no \\
scale cap (0.001, current strength) & \texttt{wcap|\allowbreak{}lr=0.001} & — & run B & 77.4\% & +6.8 [−0.0, +13.7] +10/−0 & 74.1\% & 84.4\% & +3.0 +4.9 +7.5 +4.9 +8.8 & yes \\
\bottomrule
\end{tabular}\end{center}

\textbf{Permuted MNIST}

\begin{center}\scriptsize\setlength{\tabcolsep}{3pt}
\begin{tabular}{@{}P{2.0cm}P{2.7cm}P{1.1cm}P{1.1cm}P{1.0cm}P{2.5cm}P{0.9cm}P{0.8cm}P{2.3cm}P{0.6cm}@{}}
\toprule
\textbf{Method} & \textbf{Selected setting} & \textbf{Grid edge} & \textbf{Run} & \textbf{Total avg. online} & \textbf{Gap to scratch (tasks 400, 500; points; [95\% t]; pos/neg seeds)} & \textbf{Late window} & \textbf{task 1} & \textbf{Gap at tasks 100, 200, 300, 400, 500} & \textbf{Loss} \\
\midrule
L2 Init & \texttt{l2i|\allowbreak{}lr=0.001,\allowbreak{}l2init=0.001} & interior & run C (rerun) & 88.4\% & −4.6 [−4.9, −4.3] +0/−10 & 88.4\% & 83.6\% & −4.9 −4.7 −4.6 −4.6 −4.6 & no \\
L2 & \texttt{l2|\allowbreak{}lr=0.001,\allowbreak{}l2zero=0.001} & interior & run B & 87.7\% & −4.3 [−4.6, −4.0] +0/−10 & 87.8\% & 83.3\% & −4.4 −4.4 −4.3 −4.3 −4.3 & no \\
S\&P & \texttt{snp|\allowbreak{}lr=0.001,\allowbreak{}snp\_eps=0.0001,\allowbreak{}snp\_sigma=0.01} & edge: snp\_sigma & run A & 86.5\% & −2.7 [−3.0, −2.3] +0/−10 & 86.5\% & 83.8\% & −2.9 −2.5 −2.6 −2.7 −2.7 & no \\
CBP & \texttt{cbp|\allowbreak{}lr=0.001,\allowbreak{}cbp\_rho=0.001} & interior & run A & 87.4\% & −3.6 [−3.9, −3.4] +0/−10 & 87.6\% & 83.7\% & −3.5 −3.6 −3.7 −3.7 −3.6 & no \\
ReDo & \texttt{redo|\allowbreak{}lr=0.001,\allowbreak{}redo\_tau=0.1,\allowbreak{}redo\_period=625} & edge: redo\_tau, redo\_period & run A & 87.8\% & −3.7 [−3.9, −3.5] +0/−10 & 87.9\% & 83.8\% & −3.9 −4.0 −4.0 −3.8 −3.6 & no \\
Layer Norm & \texttt{ln|\allowbreak{}lr=0.001} & interior & run B & 80.5\% & +10.5 [+9.9, +11.0] +10/−0 & 76.9\% & 87.9\% & +4.2 +7.4 +9.0 +9.7 +11.3 & yes \\
Concat ReLU & \texttt{crelu|\allowbreak{}lr=0.001} & interior & run B & 85.0\% & +0.3 [−0.0, +0.6] +8/−2 & 84.4\% & 84.5\% & −0.6 −0.4 +0.1 +0.1 +0.4 & no \\
unmodified (selected) & \texttt{none|\allowbreak{}lr=0.001} & interior & run A & 78.8\% & +8.1 [+7.6, +8.6] +10/−0 & 75.3\% & 83.8\% & +2.1 +4.7 +6.6 +7.9 +8.3 & yes \\
centering + bias suppression & \texttt{block|\allowbreak{}lr=0.003,\allowbreak{}B\_lam=0.08565} & edge: B\_lam & run C & 85.0\% & +3.5 [+3.3, +3.7] +10/−0 & 83.9\% & 87.4\% & +1.8 +2.4 +3.5 +3.4 +3.6 & yes \\
kunekune & \texttt{kkt1|\allowbreak{}lr=0.001} & interior & run A & 85.0\% & +2.0 [+1.8, +2.3] +10/−0 & 84.4\% & 86.2\% & +0.9 +1.4 +1.5 +2.1 +2.0 & yes \\
centering + bias suppression + L2 Init & \texttt{combo|\allowbreak{}lr=0.003,\allowbreak{}l2init=0.001,\allowbreak{}B\_lam=0.08565} & — & run C & 89.4\% & −2.3 [−2.5, −2.1] +0/−10 & 89.5\% & 87.1\% & −2.3 −2.4 −2.2 −2.4 −2.2 & no \\
unmodified (0.001) & \texttt{none|\allowbreak{}lr=0.001} & — & run A & 78.8\% & +8.1 [+7.6, +8.6] +10/−0 & 75.3\% & 83.8\% & +2.1 +4.7 +6.6 +7.9 +8.3 & yes \\
centering + bias suppression (0.001, current strength) & \texttt{block|\allowbreak{}lr=0.001} & — & run A & 84.7\% & +1.7 [+1.4, +1.9] +10/−0 & 83.4\% & 84.8\% & −1.1 +0.2 +0.9 +1.7 +1.6 & yes \\
kunekune (0.001, current strength) & \texttt{kkt1|\allowbreak{}lr=0.001} & — & run A & 85.0\% & +2.0 [+1.8, +2.3] +10/−0 & 84.4\% & 86.2\% & +0.9 +1.4 +1.5 +2.1 +2.0 & yes \\
\bottomrule
\end{tabular}\end{center}

\begin{itemize}
\item Note: ``edge: lr'' means that the selected learning rate is at an edge of the grid \{0.0001, 0.001, 0.003, 0.01\}; ``edge: B\_clip'' and the like mean an edge of the strength grid. The Run column gives the source of the evaluation run (run A = the first run with 2 learning rates; run B = the run that added the 7 methods; run C = the run that reselected 4 learning rates and strengths; ``rerun'' = the same setting run again in the process of run C, used for comparisons within the same process). The scale cap was added only to Continual ImageNet (the row-norm cap cap12 on Permuted MNIST is from run A; its gap to scratch is −3.4 points for the mean of tasks 400 and 500 [§6.2, J.12], −3.5 points for task 500 alone, and its total average online accuracy is 87.4\%).
\item The loss of L2 Init on Permuted MNIST is −4.6 in this table (rerun in run C) and −4.7 in the table of J.12 (selection at learning rates \{0.001, 0.0001\} = value of run A); the runs differ (both agree with the records).
\item Gaps of centering without bias suppression (CHp; learning rate 0.001; Continual ImageNet; run A) and of centering + bias suppression relative to the unmodified network at learning rate 0.001 (§6.2). Excerpt:
\end{itemize}

\begin{Verbatim}[fontsize=\scriptsize,breaklines=true,breakanywhere=true]
CHp|lr=0.001 late +0.3722 [+0.3135, +0.4308] pos 10/10 improved | gap500 +0.3855 [+0.3176, +0.4534] pos 10/10 improved (sign: + = smaller loss than A0)
block|lr=0.001 late +0.3748 [+0.3144, +0.4351] pos 10/10 improved | gap500 +0.3856 [+0.3187, +0.4525] pos 10/10 improved (sign: + = smaller loss than A0)
\end{Verbatim}

\textbf{Pairwise comparisons (Continual ImageNet)}

Centering + bias suppression − each method (same-seed pairs; points; [95\% $t$]; positive/negative seeds; verdict X>C = intervention better, C>X = other better, ≈ = undecided):

\begin{center}\footnotesize\setlength{\tabcolsep}{3pt}
\begin{tabular}{@{}P{3.0cm}P{4.4cm}P{4.4cm}P{4.4cm}@{}}
\toprule
\textbf{Other method} & \textbf{Gap to scratch (positive = the intervention preserved more)} & \textbf{Total avg. online} & \textbf{Late window} \\
\midrule
L2 Init & ≈ +0.1 [−0.4, +0.5] +5/−5 & ≈ +2.4 [+1.2, +3.6] +9/−1 & ≈ +1.9 [+1.2, +2.6] +9/−1 \\
L2 & ≈ +0.8 [+0.1, +1.5] +8/−2 (bar raised) & ≈ +4.4 [+3.1, +5.7] +9/−1 (bar raised) & X>C +3.4 [+2.8, +3.9] +10/−0 (bar raised) \\
S\&P & X>C +5.1 [+4.0, +6.3] +10/−0 (bar raised) & X>C +8.5 [+7.1, +9.9] +10/−0 (bar raised) & X>C +8.7 [+7.8, +9.6] +10/−0 (bar raised) \\
CBP & X>C +3.8 [+3.0, +4.7] +10/−0 (bar raised) & X>C +6.3 [+5.0, +7.6] +10/−0 (bar raised) & X>C +6.5 [+5.9, +7.1] +10/−0 (bar raised) \\
ReDo & X>C +3.3 [+1.9, +4.7] +10/−0 (bar raised) & X>C +5.6 [+4.3, +6.9] +10/−0 (bar raised) & X>C +6.3 [+5.7, +6.9] +10/−0 (bar raised) \\
Layer Norm & X>C +24.8 [+21.3, +28.3] +10/−0 (bar raised) & X>C +23.2 [+22.0, +24.3] +10/−0 (bar raised) & X>C +28.2 [+25.8, +30.5] +10/−0 (bar raised) \\
Concat ReLU & X>C +2.4 [+1.8, +3.0] +10/−0 (bar raised) & ≈ +3.3 [+2.0, +4.6] +9/−1 (bar raised) & X>C +3.6 [+2.9, +4.3] +10/−0 (bar raised) \\
unmodified (selected) & ≈ +3.6 [+0.6, +6.6] +8/−2 (bar raised) & X>C +16.7 [+15.4, +18.0] +10/−0 (bar raised) & X>C +17.0 [+16.0, +18.0] +10/−0 (bar raised) \\
\bottomrule
\end{tabular}\end{center}

kunekune − each method (same notation):

\begin{center}\footnotesize\setlength{\tabcolsep}{3pt}
\begin{tabular}{@{}P{3.0cm}P{4.4cm}P{4.4cm}P{4.4cm}@{}}
\toprule
\textbf{Other method} & \textbf{Gap to scratch (positive = the intervention preserved more)} & \textbf{Total avg. online} & \textbf{Late window} \\
\midrule
L2 Init & C>X −5.5 [−6.4, −4.6] +0/−10 (bar raised) & ≈ +1.0 [+0.9, +1.2] +10/−0 (bar raised) & ≈ −0.1 [−0.4, +0.3] +3/−7 (bar raised) \\
L2 & C>X −4.8 [−6.0, −3.5] +0/−10 (bar raised) & X>C +3.1 [+2.8, +3.4] +10/−0 (bar raised) & X>C +1.4 [+1.1, +1.8] +10/−0 (bar raised) \\
S\&P & ≈ −0.4 [−1.3, +0.4] +4/−6 & X>C +7.2 [+6.9, +7.5] +10/−0 & X>C +6.7 [+5.9, +7.5] +10/−0 \\
CBP & ≈ −1.7 [−3.1, −0.3] +3/−7 & X>C +5.0 [+4.8, +5.1] +10/−0 & X>C +4.6 [+4.3, +4.9] +10/−0 \\
ReDo & C>X −2.3 [−3.9, −0.6] +2/−8 & X>C +4.2 [+4.1, +4.4] +10/−0 & X>C +4.4 [+4.1, +4.6] +10/−0 \\
Layer Norm & X>C +19.2 [+15.6, +22.8] +10/−0 (bar raised) & X>C +21.8 [+21.2, +22.5] +10/−0 (bar raised) & X>C +26.2 [+23.9, +28.6] +10/−0 (bar raised) \\
Concat ReLU & C>X −3.2 [−4.1, −2.2] +0/−10 (bar raised) & X>C +1.9 [+1.8, +2.0] +10/−0 (bar raised) & X>C +1.7 [+1.4, +1.9] +10/−0 (bar raised) \\
unmodified (selected) & ≈ −1.9 [−4.7, +0.8] +4/−6 & X>C +15.3 [+15.2, +15.5] +10/−0 & X>C +15.1 [+14.4, +15.8] +10/−0 \\
\bottomrule
\end{tabular}\end{center}

Scale cap − each method (same notation):

\begin{center}\footnotesize\setlength{\tabcolsep}{3pt}
\begin{tabular}{@{}P{3.0cm}P{4.4cm}P{4.4cm}P{4.4cm}@{}}
\toprule
\textbf{Other method} & \textbf{Gap to scratch (positive = the intervention preserved more)} & \textbf{Total avg. online} & \textbf{Late window} \\
\midrule
L2 Init & C>X −18.2 [−26.7, −9.7] +0/−10 & C>X −11.6 [−14.5, −8.7] +0/−10 & C>X −15.4 [−22.3, −8.5] +0/−10 \\
L2 & C>X −17.5 [−26.1, −8.8] +0/−10 (bar raised) & C>X −9.6 [−12.3, −6.9] +0/−10 (bar raised) & C>X −13.9 [−20.9, −7.0] +0/−10 (bar raised) \\
S\&P & C>X −13.1 [−21.1, −5.1] +0/−10 (bar raised) & C>X −5.5 [−8.3, −2.8] +0/−10 (bar raised) & C>X −8.6 [−15.7, −1.6] +0/−10 (bar raised) \\
CBP & C>X −14.4 [−22.5, −6.3] +0/−10 (bar raised) & C>X −7.7 [−10.6, −4.8] +0/−10 (bar raised) & C>X −10.8 [−17.6, −4.0] +0/−10 (bar raised) \\
ReDo & C>X −15.0 [−23.1, −6.8] +0/−10 (bar raised) & C>X −8.4 [−11.3, −5.5] +0/−10 (bar raised) & C>X −11.0 [−17.9, −4.1] +0/−10 (bar raised) \\
Layer Norm & ≈ +6.5 [−3.3, +16.3] +7/−3 (bar raised) & ≈ +9.1 [+5.9, +12.4] +9/−1 (bar raised) & ≈ +10.9 [+3.1, +18.7] +9/−1 (bar raised) \\
Concat ReLU & C>X −15.9 [−24.0, −7.7] +0/−10 (bar raised) & C>X −10.7 [−13.6, −7.9] +0/−10 (bar raised) & C>X −13.7 [−20.6, −6.8] +0/−10 (bar raised) \\
unmodified (selected) & C>X −14.7 [−24.9, −4.4] +0/−10 (bar raised) & ≈ +2.6 [−0.3, +5.6] +9/−1 (bar raised) & ≈ −0.3 [−7.4, +6.8] +8/−2 (bar raised) \\
\bottomrule
\end{tabular}\end{center}

Centering + bias suppression + L2 Init − each method (same notation):

\begin{center}\footnotesize\setlength{\tabcolsep}{3pt}
\begin{tabular}{@{}P{3.0cm}P{4.4cm}P{4.4cm}P{4.4cm}@{}}
\toprule
\textbf{Other method} & \textbf{Gap to scratch (positive = the intervention preserved more)} & \textbf{Total avg. online} & \textbf{Late window} \\
\midrule
L2 Init & X>C +1.4 [+0.7, +2.2] +10/−0 (bar raised) & X>C +4.4 [+4.2, +4.6] +10/−0 (bar raised) & X>C +3.5 [+3.1, +4.0] +10/−0 (bar raised) \\
L2 & X>C +2.2 [+1.3, +3.0] +10/−0 & X>C +6.4 [+6.1, +6.8] +10/−0 & X>C +5.0 [+4.5, +5.5] +10/−0 \\
S\&P & X>C +6.5 [+5.0, +8.0] +10/−0 (bar raised) & X>C +10.5 [+10.3, +10.8] +10/−0 (bar raised) & X>C +10.3 [+9.5, +11.2] +10/−0 (bar raised) \\
CBP & X>C +5.2 [+4.0, +6.5] +10/−0 (bar raised) & X>C +8.3 [+8.2, +8.5] +10/−0 (bar raised) & X>C +8.2 [+7.7, +8.7] +10/−0 (bar raised) \\
ReDo & X>C +4.7 [+2.9, +6.5] +10/−0 (bar raised) & X>C +7.6 [+7.5, +7.7] +10/−0 (bar raised) & X>C +8.0 [+7.5, +8.4] +10/−0 (bar raised) \\
Layer Norm & X>C +26.1 [+22.2, +30.1] +10/−0 & X>C +25.2 [+24.5, +25.8] +10/−0 & X>C +29.8 [+27.2, +32.4] +10/−0 \\
Concat ReLU & X>C +3.8 [+2.7, +4.9] +10/−0 & X>C +5.3 [+5.2, +5.3] +10/−0 & X>C +5.2 [+4.9, +5.6] +10/−0 \\
unmodified (selected) & ≈ +5.0 [+1.6, +8.4] +9/−1 (bar raised) & X>C +18.7 [+18.5, +18.9] +10/−0 (bar raised) & X>C +18.7 [+17.7, +19.6] +10/−0 (bar raised) \\
\bottomrule
\end{tabular}\end{center}

\textbf{Pairwise comparisons (Permuted MNIST)}

Centering + bias suppression − each method (same notation):

\begin{center}\footnotesize\setlength{\tabcolsep}{3pt}
\begin{tabular}{@{}P{3.0cm}P{4.4cm}P{4.4cm}P{4.4cm}@{}}
\toprule
\textbf{Other method} & \textbf{Gap to scratch (positive = the intervention preserved more)} & \textbf{Total avg. online} & \textbf{Late window} \\
\midrule
L2 Init & C>X −8.2 [−8.5, −7.8] +0/−10 & C>X −3.4 [−3.4, −3.4] +0/−10 & C>X −4.5 [−4.5, −4.4] +0/−10 \\
L2 & C>X −7.8 [−8.2, −7.5] +0/−10 & C>X −2.8 [−2.8, −2.7] +0/−10 & C>X −3.8 [−3.9, −3.7] +0/−10 \\
S\&P & C>X −6.2 [−6.5, −5.8] +0/−10 & C>X −1.5 [−1.6, −1.5] +0/−10 & C>X −2.6 [−2.7, −2.5] +0/−10 \\
CBP & C>X −7.2 [−7.5, −6.8] +0/−10 & C>X −2.4 [−2.5, −2.4] +0/−10 & C>X −3.6 [−3.7, −3.5] +0/−10 \\
ReDo & C>X −7.3 [−7.5, −7.0] +0/−10 & C>X −2.8 [−2.9, −2.8] +0/−10 & C>X −4.0 [−4.1, −3.9] +0/−10 \\
Layer Norm & X>C +6.9 [+6.4, +7.4] +10/−0 & X>C +4.5 [+4.4, +4.5] +10/−0 & X>C +7.1 [+6.9, +7.2] +10/−0 \\
Concat ReLU & C>X −3.3 [−3.6, −2.9] +0/−10 & ≈ −0.0 [−0.1, −0.0] +0/−10 & ≈ −0.5 [−0.6, −0.4] +0/−10 \\
unmodified (selected) & X>C +4.6 [+4.0, +5.1] +10/−0 & X>C +6.2 [+6.0, +6.3] +10/−0 & X>C +8.6 [+8.3, +8.8] +10/−0 \\
\bottomrule
\end{tabular}\end{center}

kunekune − each method (same notation):

\begin{center}\footnotesize\setlength{\tabcolsep}{3pt}
\begin{tabular}{@{}P{3.0cm}P{4.4cm}P{4.4cm}P{4.4cm}@{}}
\toprule
\textbf{Other method} & \textbf{Gap to scratch (positive = the intervention preserved more)} & \textbf{Total avg. online} & \textbf{Late window} \\
\midrule
L2 Init & C>X −6.6 [−7.0, −6.3] +0/−10 & C>X −3.4 [−3.4, −3.4] +0/−10 & C>X −4.0 [−4.1, −3.9] +0/−10 \\
L2 & C>X −6.3 [−6.7, −6.0] +0/−10 & C>X −2.7 [−2.8, −2.7] +0/−10 & C>X −3.4 [−3.4, −3.3] +0/−10 \\
S\&P & C>X −4.7 [−5.1, −4.3] +0/−10 & C>X −1.5 [−1.5, −1.5] +0/−10 & C>X −2.1 [−2.2, −2.1] +0/−10 \\
CBP & C>X −5.7 [−6.1, −5.3] +0/−10 & C>X −2.4 [−2.4, −2.4] +0/−10 & C>X −3.2 [−3.2, −3.1] +0/−10 \\
ReDo & C>X −5.8 [−6.1, −5.4] +0/−10 & C>X −2.8 [−2.8, −2.8] +0/−10 & C>X −3.5 [−3.6, −3.4] +0/−10 \\
Layer Norm & X>C +8.4 [+7.9, +9.0] +10/−0 & X>C +4.5 [+4.4, +4.6] +10/−0 & X>C +7.5 [+7.4, +7.7] +10/−0 \\
Concat ReLU & C>X −1.7 [−2.2, −1.3] +0/−10 & ≈ −0.0 [−0.0, +0.0] +3/−7 & ≈ −0.0 [−0.1, +0.0] +4/−6 \\
unmodified (selected) & X>C +6.1 [+5.4, +6.7] +10/−0 & X>C +6.2 [+6.1, +6.4] +10/−0 & X>C +9.1 [+8.8, +9.3] +10/−0 \\
\bottomrule
\end{tabular}\end{center}

Centering + bias suppression + L2 Init − each method (same notation):

\begin{center}\footnotesize\setlength{\tabcolsep}{3pt}
\begin{tabular}{@{}P{3.0cm}P{4.4cm}P{4.4cm}P{4.4cm}@{}}
\toprule
\textbf{Other method} & \textbf{Gap to scratch (positive = the intervention preserved more)} & \textbf{Total avg. online} & \textbf{Late window} \\
\midrule
L2 Init & C>X −2.3 [−2.7, −2.0] +0/−10 & X>C +1.0 [+1.0, +1.1] +10/−0 & X>C +1.1 [+1.0, +1.1] +10/−0 \\
L2 & C>X −2.0 [−2.3, −1.7] +0/−10 & X>C +1.7 [+1.7, +1.7] +10/−0 & X>C +1.7 [+1.6, +1.8] +10/−0 \\
S\&P & ≈ −0.4 [−0.7, −0.1] +1/−9 & X>C +2.9 [+2.9, +2.9] +10/−0 & X>C +2.9 [+2.9, +3.0] +10/−0 \\
CBP & C>X −1.4 [−1.6, −1.1] +0/−10 & X>C +2.0 [+2.0, +2.1] +10/−0 & X>C +1.9 [+1.8, +2.0] +10/−0 \\
ReDo & C>X −1.4 [−1.7, −1.2] +0/−10 & X>C +1.6 [+1.6, +1.6] +10/−0 & X>C +1.6 [+1.5, +1.6] +10/−0 \\
Layer Norm & X>C +12.8 [+12.3, +13.2] +10/−0 & X>C +8.9 [+8.9, +9.0] +10/−0 & X>C +12.6 [+12.4, +12.8] +10/−0 \\
Concat ReLU & X>C +2.6 [+2.2, +2.9] +10/−0 & X>C +4.4 [+4.4, +4.4] +10/−0 & X>C +5.0 [+5.0, +5.1] +10/−0 \\
unmodified (selected) & X>C +10.4 [+9.9, +10.9] +10/−0 & X>C +10.6 [+10.5, +10.8] +10/−0 & X>C +14.1 [+13.8, +14.4] +10/−0 \\
\bottomrule
\end{tabular}\end{center}

\subsection{Split CIFAR-100 (reduced ResNet-18, 10 seeds)}
\label{app:J4}
gap20 = network trained from scratch − continued network at task 20; W = difference in the learning speed of a probe with a new output layer; p = two-sided exact sign test. SPL = split loss; MASK = restriction without the output offset; PA = oracle that aligns the output bias at the boundary; L2I = L2 Init. Excerpt of the analysis output:

\begin{Verbatim}[fontsize=\scriptsize,breaklines=true,breakanywhere=true]
# readout_resnet: seeds per arm {'none': [0, 1, 2, 3, 4, 5, 6, 7, 8, 9], 'SPL': [0, 1, 2, 3, 4, 5, 6, 7, 8, 9], 'MASK': [0, 1, 2, 3, 4, 5, 6, 7, 8, 9], 'PA': [0, 1, 2, 3, 4, 5, 6, 7, 8, 9], 'none_L2I': [0, 1, 2, 3, 4, 5, 6, 7, 8, 9], 'SPL_L2I': [0, 1, 2, 3, 4, 5, 6, 7, 8, 9]}
none      late 0.6578 | gap20 +0.1559 [+0.0999, +0.2119] pos 10/10 p 0.0020 | W -0.0816 [-0.0973, -0.0660] pos 0/10 p 0.0020
SPL       late 0.9224 | gap20 -0.1341 [-0.1582, -0.1099] pos 0/10 p 0.0020 | W -0.1746 [-0.1973, -0.1519] pos 0/10 p 0.0020
MASK      late 0.5872 | gap20 +0.1891 [+0.1338, +0.2443] pos 10/10 p 0.0020 | W -0.1726 [-0.1940, -0.1512] pos 0/10 p 0.0020
PA        late 0.9403 | gap20 -0.1584 [-0.1882, -0.1286] pos 0/10 p 0.0020 | W -0.1675 [-0.1870, -0.1481] pos 0/10 p 0.0020
none_L2I  late 0.8167 | gap20 -0.0389 [-0.0549, -0.0228] pos 0/10 p 0.0020 | W -0.1362 [-0.1524, -0.1200] pos 0/10 p 0.0020
SPL_L2I   late 0.8933 | gap20 -0.1089 [-0.1304, -0.0873] pos 0/10 p 0.0020 | W -0.1705 [-0.1949, -0.1460] pos 0/10 p 0.0020
SPL - none: late +0.2646 [+0.2473, +0.2820] pos 10/10 p 0.0020 | W -0.0930 [-0.1041, -0.0820] pos 0/10 p 0.0020 | gap20 -0.2900 [-0.3326, -0.2474] pos 0/10 p 0.0020
MASK - none: late -0.0706 [-0.0959, -0.0453] pos 0/10 p 0.0020 | W -0.0910 [-0.1024, -0.0796] pos 0/10 p 0.0020 | gap20 +0.0331 [-0.0447, +0.1110] pos 8/10 p 0.1094
PA - none: late +0.2825 [+0.2649, +0.3001] pos 10/10 p 0.0020 | W -0.0859 [-0.0926, -0.0793] pos 0/10 p 0.0020 | gap20 -0.3143 [-0.3573, -0.2714] pos 0/10 p 0.0020
none_L2I - none: late +0.1589 [+0.1340, +0.1839] pos 10/10 p 0.0020 | W -0.0546 [-0.0634, -0.0458] pos 0/10 p 0.0020 | gap20 -0.1948 [-0.2480, -0.1415] pos 0/10 p 0.0020
SPL_L2I - none: late +0.2355 [+0.2179, +0.2530] pos 10/10 p 0.0020 | W -0.0888 [-0.1003, -0.0773] pos 0/10 p 0.0020 | gap20 -0.2648 [-0.3096, -0.2199] pos 0/10 p 0.0020
SPL_L2I - SPL: late -0.0292 [-0.0315, -0.0269] pos 0/10 p 0.0020 | W +0.0042 [-0.0017, +0.0101] pos 7/10 p 0.3438
PA - SPL: late +0.0178 [+0.0162, +0.0195] pos 10/10 p 0.0020 | W +0.0071 [-0.0004, +0.0146] pos 8/10 p 0.1094
MASK - SPL: late -0.3352 [-0.3541, -0.3163] pos 0/10 p 0.0020 | W +0.0020 [-0.0054, +0.0094] pos 5/10 p 1.0000
\end{Verbatim}

Restricting without removing the output offset lowered late online accuracy by 7.1 points relative to the unmodified network (0/10) = \texttt{MASK - none: late -0.0706 \ldots\ pos 0/10} in the excerpt above. The loss of MASK (softmax restricted to the classes present in the minibatch) has the same form as the incoming-minibatch loss of ER-ACE (\citealp[§4.3]{caccia2022new}; denominator restricted to the classes that have appeared) and the same idea as SS-IL (\citealp[§5]{ahn2021ssil}; separate softmax for new and old classes). Both are combined with replay memory, whereas this sequence uses no replay.

\subsection{Replication controls and post hoc checks}
\label{app:J5}
Run B: difference when the same L2 Init setting was rerun in the process of run B (on Continual ImageNet, G and the late window did not agree, and per the rules the bar for comparisons across runs was raised; on Permuted MNIST they agreed). Excerpts (Continual ImageNet, then Permuted MNIST):

\begin{Verbatim}[fontsize=\scriptsize,breaklines=true,breakanywhere=true]
## replication control: the same L2 Init setting run again in the run B process minus the run A (same seed)
  M1: +0.0013 [-0.0015, +0.0041] +6/-4 (n 10) -> reproduced
  G: -0.0058 [-0.0129, +0.0013] +2/-8 (n 10) -> NOT REPRODUCED (flag)
  M3: +0.0091 [+0.0042, +0.0140] +10/-0 (n 10) -> NOT REPRODUCED (flag)
\end{Verbatim}

\begin{Verbatim}[fontsize=\scriptsize,breaklines=true,breakanywhere=true]
## replication control: the same L2 Init setting run again in the run B process minus the run A (same seed)
  M1: +0.0000 [-0.0002, +0.0002] +5/-5 (n 10) -> reproduced
  G: +0.0009 [-0.0005, +0.0022] +7/-3 (n 10) -> reproduced
  M3: -0.0003 [-0.0006, -0.0000] +2/-8 (n 10) -> reproduced
\end{Verbatim}

[Post hoc] (run B; outside the rules): counted without raising the bar, the labels for total average online and for preservation do not change; the fraction by which the loss on Permuted MNIST was reduced, computed for task 500 alone:

\begin{Verbatim}[fontsize=\scriptsize,breaklines=true,breakanywhere=true]
(1) CI block vs rep M1: X>C +0.0301 [+0.0285, +0.0316] +10/-0 (n 10)
(1) CI block vs rep G: ~ -0.0047 [-0.0098, +0.0004] +3/-7 (n 10)
(1) CI block vs l2i M1: X>C +0.0314 [+0.0286, +0.0341] +10/-0 (n 10)
(1) CI block vs l2i G: ~ +0.0011 [-0.0057, +0.0080] +5/-5 (n 10)
(1) CI block vs l2 M1: X>C +0.0495 [+0.0460, +0.0531] +10/-0 (n 10)
(1) CI block vs l2 G: ~ +0.0088 [+0.0028, +0.0148] +9/-1 (n 10)
(1) CI block vs crelu M1: X>C +0.0381 [+0.0364, +0.0399] +10/-0 (n 10)
(1) CI block vs crelu G: ~ +0.0249 [+0.0153, +0.0344] +9/-1 (n 10)
(1) CI combo vs rep M1: X>C +0.0449 [+0.0432, +0.0466] +10/-0 (n 10)
(1) CI combo vs rep G: ~ +0.0082 [+0.0015, +0.0149] +8/-2 (n 10)
(1) CI combo vs l2i M1: X>C +0.0462 [+0.0435, +0.0490] +10/-0 (n 10)
(1) CI combo vs l2i G: ~ +0.0140 [+0.0050, +0.0231] +9/-1 (n 10)
(1) CI combo vs l2 M1: X>C +0.0644 [+0.0612, +0.0675] +10/-0 (n 10)
(1) CI combo vs l2 G: X>C +0.0217 [+0.0134, +0.0300] +10/-0 (n 10)
(1) CI combo vs crelu M1: X>C +0.0530 [+0.0525, +0.0535] +10/-0 (n 10)
(1) CI combo vs crelu G: X>C +0.0378 [+0.0270, +0.0485] +10/-0 (n 10)
(2) PM block: reduction with G(400,500) = 0.7940 | task 500 only = 0.8038
(2) PM kkt1: reduction with G(400,500) = 0.7497 | task 500 only = 0.7640
(3) changed: kkt1 vs l2 M3: judge ~ -> plain line X>C
(3) changed: kkt1 vs crelu M3: judge ~ -> plain line X>C
(3) labels changed by the raised line: 2
\end{Verbatim}

Run C: replication controls (the same L2 Init setting and the selected setting of centering + bias suppression rerun in the process of run C) and, within the same process, centering + bias suppression vs L2 Init (descriptive; bar not raised).

\textbf{Continual ImageNet} (excerpt of the verdict output)

\begin{Verbatim}[fontsize=\scriptsize,breaklines=true,breakanywhere=true]
## replication control: rep:l2i|lr=0.001,l2init=0.001 (run C process) minus l2i|lr=0.001,l2init=0.001 (run A), same seeds
  M1: +0.0022 [-0.0005, +0.0049] +7/-3 (n 10) -> reproduced
  G: +0.0005 [-0.0063, +0.0072] +5/-5 (n 10) -> reproduced
  M3: +0.0034 [-0.0011, +0.0079] +7/-3 (n 10) -> reproduced
## replication control: rep:block|lr=0.001 (run C process) minus block|lr=0.001 (run A), same seeds
  M1: -0.0053 [-0.0171, +0.0065] +4/-6 (n 10) -> NOT REPRODUCED (flag)
  G: +0.0009 [-0.0035, +0.0052] +5/-5 (n 10) -> reproduced
  M3: -0.0030 [-0.0086, +0.0027] +3/-7 (n 10) -> reproduced
## same run C process (descriptive): rep:block|lr=0.001 vs rep:l2i|lr=0.001,l2init=0.001
  G: ~ +0.0007 [-0.0038, +0.0053] +5/-5 (n 10)
  M1: ~ +0.0239 [+0.0122, +0.0356] +9/-1 (n 10)
  M3: ~ +0.0188 [+0.0117, +0.0259] +9/-1 (n 10)
\end{Verbatim}

\textbf{Permuted MNIST} (excerpt of the verdict output)

\begin{Verbatim}[fontsize=\scriptsize,breaklines=true,breakanywhere=true]
## replication control: rep:l2i|lr=0.001,l2init=0.001 (run C process) minus l2i|lr=0.001,l2init=0.001 (run A), same seeds
  M1: +0.0000 [-0.0002, +0.0002] +5/-5 (n 10) -> reproduced
  G: +0.0009 [-0.0005, +0.0022] +7/-3 (n 10) -> reproduced
  M3: -0.0003 [-0.0006, -0.0000] +2/-8 (n 10) -> reproduced
## same run C process (descriptive): block|lr=0.003,B_lam=0.08565 vs rep:l2i|lr=0.001,l2init=0.001
  G: C>X -0.0815 [-0.0850, -0.0781] +0/-10 (n 10)
  M1: C>X -0.0342 [-0.0344, -0.0339] +0/-10 (n 10)
  M3: C>X -0.0446 [-0.0454, -0.0438] +0/-10 (n 10)
\end{Verbatim}

\subsection{Learning-rate table and strength table (calibration seeds 100--102; descriptive; early window = tasks 11--60; late window = tasks 451--500)}
\label{app:J6}
Each cell = total average online / early window − late window (positive = declined late).

\begin{center}\footnotesize\setlength{\tabcolsep}{3pt}
\begin{tabular}{@{}P{2.6cm}P{4.0cm}P{2.2cm}P{2.2cm}P{2.2cm}P{2.2cm}@{}}
\toprule
\textbf{Sequence} & \textbf{Method (other keys at their selected values)} & \textbf{Learning rate 0.0001} & \textbf{Learning rate 0.001} & \textbf{Learning rate 0.003} & \textbf{Learning rate 0.01} \\
\midrule
Continual ImageNet & unmodified (selected) & 75.0\% / −2.1 & 64.3\% / +31.6 & 57.2\% / +24.7 & 49.5\% / +0.4 \\
Continual ImageNet & centering + bias suppression & 77.7\% / −3.1 & 92.3\% / −2.2 & 66.7\% / −2.6 & 66.3\% / −5.1 \\
Continual ImageNet & kunekune & 80.7\% / −0.6 & 90.5\% / −1.2 & 87.5\% / +1.6 & 77.9\% / +1.5 \\
Continual ImageNet & scale cap & 76.3\% / −2.0 & 80.7\% / +1.3 & 63.5\% / +20.1 & 50.1\% / +5.4 \\
Continual ImageNet & Layer Norm & 68.5\% / +11.0 & 68.4\% / +6.0 & 61.2\% / +22.0 & 59.2\% / +18.5 \\
Continual ImageNet & Concat ReLU & 79.4\% / −4.2 & 88.6\% / −2.5 & 88.5\% / −2.0 & 77.8\% / +1.9 \\
Continual ImageNet & L2 Init & 75.8\% / −2.4 & 89.4\% / −5.7 & 84.6\% / −2.2 & 74.5\% / −2.2 \\
Continual ImageNet & L2 & 72.8\% / −0.4 & 87.5\% / −4.7 & 81.9\% / −2.0 & 61.1\% / +1.1 \\
Continual ImageNet & S\&P & 71.6\% / +1.4 & 83.1\% / −2.7 & 70.7\% / −0.7 & 49.6\% / +1.5 \\
Continual ImageNet & CBP & 71.0\% / +1.4 & 85.6\% / −2.3 & 83.7\% / −1.2 & 73.4\% / −0.8 \\
Continual ImageNet & ReDo & 75.7\% / −0.9 & 86.2\% / −2.0 & 81.9\% / −1.4 & 50.7\% / +0.6 \\
Permuted MNIST & unmodified (selected) & 74.2\% / +1.5 & 78.8\% / +8.2 & 75.4\% / +8.2 & 51.3\% / +20.9 \\
Permuted MNIST & centering + bias suppression & 75.6\% / +4.5 & 84.7\% / +3.8 & 85.0\% / +2.9 & 79.3\% / +1.1 \\
Permuted MNIST & kunekune & 76.3\% / +3.7 & 85.0\% / +1.9 & 84.0\% / +0.9 & 74.7\% / +0.4 \\
Permuted MNIST & Layer Norm & 78.4\% / +8.9 & 80.5\% / +9.0 & 79.4\% / +10.4 & 75.7\% / +13.3 \\
Permuted MNIST & Concat ReLU & 76.9\% / +2.4 & 85.0\% / +2.1 & 83.4\% / +1.3 & 78.9\% / −0.3 \\
Permuted MNIST & L2 Init & 79.7\% / −3.7 & 88.4\% / −0.0 & 87.7\% / +0.1 & 82.8\% / −0.3 \\
Permuted MNIST & L2 & 79.3\% / −3.6 & 87.7\% / +0.0 & 87.0\% / −0.0 & 82.1\% / −0.1 \\
Permuted MNIST & S\&P & 76.2\% / +0.2 & 86.5\% / +0.1 & 84.0\% / +0.1 & 72.7\% / +2.5 \\
Permuted MNIST & CBP & 75.0\% / −1.7 & 87.4\% / −0.4 & 86.4\% / −0.1 & 81.7\% / −0.1 \\
Permuted MNIST & ReDo & 77.2\% / −2.8 & 87.8\% / −0.3 & 86.9\% / +0.0 & 80.2\% / −0.8 \\
\bottomrule
\end{tabular}\end{center}

\textbf{Strength table} (strength of the mechanism interventions × learning rate; total average online):

\begin{center}\footnotesize\setlength{\tabcolsep}{3pt}
\begin{tabular}{@{}P{2.4cm}P{3.0cm}P{3.0cm}P{1.8cm}P{1.8cm}P{1.8cm}P{1.8cm}@{}}
\toprule
\textbf{Sequence} & \textbf{Method} & \textbf{Strength} & \textbf{Learning rate 0.0001} & \textbf{Learning rate 0.001} & \textbf{Learning rate 0.003} & \textbf{Learning rate 0.01} \\
\midrule
Continual ImageNet & centering + bias suppression & B\_clip = 0.1 & 77.48\% & 86.10\% & 67.19\% & 67.29\% \\
Continual ImageNet & centering + bias suppression & B\_clip = 0.2 (current value) & 77.70\% & 92.33\% & 66.71\% & 66.30\% \\
Continual ImageNet & centering + bias suppression & B\_clip = 0.4 & 77.68\% & 92.05\% & 67.33\% & 67.06\% \\
Continual ImageNet & kunekune & kkt\_a0 = 0.3 & 79.09\% & 89.18\% & 88.89\% & 76.50\% \\
Continual ImageNet & kunekune & kkt\_a0 = 0.6 (current value) & 80.68\% & 90.50\% & 87.50\% & 77.90\% \\
Continual ImageNet & kunekune & kkt\_a0 = 1.2 & 80.02\% & 90.17\% & 88.21\% & 78.67\% \\
Continual ImageNet & scale cap & wcap\_k = 0.5 & 75.85\% & 78.81\% & 66.86\% & 49.80\% \\
Continual ImageNet & scale cap & wcap\_k = 1 (current value) & 76.11\% & 78.68\% & 62.05\% & 49.53\% \\
Continual ImageNet & scale cap & wcap\_k = 2 & 76.31\% & 80.72\% & 63.51\% & 50.10\% \\
Permuted MNIST & centering + bias suppression & B\_lam = 0.08565 & 75.65\% & 84.69\% & 84.96\% & 79.32\% \\
Permuted MNIST & centering + bias suppression & B\_lam = 0.1713 (current value) & 75.50\% & 84.67\% & 84.91\% & 79.33\% \\
Permuted MNIST & centering + bias suppression & B\_lam = 0.3426 & 75.40\% & 84.72\% & 84.89\% & 79.27\% \\
Permuted MNIST & kunekune & kkt\_a0 = 0.3 & 76.42\% & 83.89\% & 81.43\% & 76.00\% \\
Permuted MNIST & kunekune & kkt\_a0 = 0.6 (current value) & 76.35\% & 84.97\% & 84.01\% & 74.72\% \\
Permuted MNIST & kunekune & kkt\_a0 = 1.2 & 75.86\% & 84.02\% & 82.63\% & 73.62\% \\
\bottomrule
\end{tabular}\end{center}

(Total average online accuracy on calibration seeds 100--102: the 4-digit values of the verdict output multiplied by 100 and written with 2 decimals; the selection used these values.)

\subsection{All points of the calibration grids}
\label{app:J7}
\textbf{Continual ImageNet (total average online on calibration seeds 100--102; 164 points; 0 diverged; 0 missing; no method with $3\times10^{-2}$ added)}

{\scriptsize\setlength{\tabcolsep}{3pt}
\begin{longtable}{@{}P{8.0cm}P{3.0cm}P{4.0cm}@{}}
\toprule
\textbf{Setting} & \textbf{Total avg. online} & \textbf{Selection} \\
\midrule
\endhead
\texttt{none|\allowbreak{}lr=0.001} & 64.26\% &  \\
\texttt{none|\allowbreak{}lr=0.0001} & 74.98\% & \textbf{selected} (edge: lr) \\
\texttt{l2i|\allowbreak{}lr=0.001,\allowbreak{}l2init=0.01} & 85.96\% &  \\
\texttt{l2i|\allowbreak{}lr=0.001,\allowbreak{}l2init=0.001} & 89.39\% & \textbf{selected} (interior) \\
\texttt{l2i|\allowbreak{}lr=0.001,\allowbreak{}l2init=0.0001} & 84.26\% &  \\
\texttt{snp|\allowbreak{}lr=0.001,\allowbreak{}snp\_eps=0.001,\allowbreak{}snp\_sigma=0.01} & 79.06\% &  \\
\texttt{snp|\allowbreak{}lr=0.001,\allowbreak{}snp\_eps=0.001,\allowbreak{}snp\_sigma=0.001} & 78.49\% &  \\
\texttt{snp|\allowbreak{}lr=0.001,\allowbreak{}snp\_eps=0.001,\allowbreak{}snp\_sigma=0.0001} & 77.97\% &  \\
\texttt{snp|\allowbreak{}lr=0.001,\allowbreak{}snp\_eps=0.0001,\allowbreak{}snp\_sigma=0.01} & 83.06\% & \textbf{selected} (edge: snp\_sigma) \\
\texttt{snp|\allowbreak{}lr=0.001,\allowbreak{}snp\_eps=0.0001,\allowbreak{}snp\_sigma=0.001} & 82.90\% &  \\
\texttt{snp|\allowbreak{}lr=0.001,\allowbreak{}snp\_eps=0.0001,\allowbreak{}snp\_sigma=0.0001} & 82.57\% &  \\
\texttt{snp|\allowbreak{}lr=0.001,\allowbreak{}snp\_eps=1e-05,\allowbreak{}snp\_sigma=0.01} & 68.42\% &  \\
\texttt{snp|\allowbreak{}lr=0.001,\allowbreak{}snp\_eps=1e-05,\allowbreak{}snp\_sigma=0.001} & 66.01\% &  \\
\texttt{snp|\allowbreak{}lr=0.001,\allowbreak{}snp\_eps=1e-05,\allowbreak{}snp\_sigma=0.0001} & 66.26\% &  \\
\texttt{cbp|\allowbreak{}lr=0.001,\allowbreak{}cbp\_rho=0.001} & 85.58\% & \textbf{selected} (interior) \\
\texttt{cbp|\allowbreak{}lr=0.001,\allowbreak{}cbp\_rho=0.0001} & 83.96\% &  \\
\texttt{cbp|\allowbreak{}lr=0.001,\allowbreak{}cbp\_rho=1e-05} & 73.17\% &  \\
\texttt{redo|\allowbreak{}lr=0.001,\allowbreak{}redo\_tau=0,\allowbreak{}redo\_period=120} & 81.31\% &  \\
\texttt{redo|\allowbreak{}lr=0.001,\allowbreak{}redo\_tau=0,\allowbreak{}redo\_period=240} & 80.46\% &  \\
\texttt{redo|\allowbreak{}lr=0.001,\allowbreak{}redo\_tau=0,\allowbreak{}redo\_period=600} & 80.78\% &  \\
\texttt{redo|\allowbreak{}lr=0.001,\allowbreak{}redo\_tau=0.025,\allowbreak{}redo\_period=120} & 86.21\% & \textbf{selected} (edge: redo\_period) \\
\texttt{redo|\allowbreak{}lr=0.001,\allowbreak{}redo\_tau=0.025,\allowbreak{}redo\_period=240} & 85.45\% &  \\
\texttt{redo|\allowbreak{}lr=0.001,\allowbreak{}redo\_tau=0.025,\allowbreak{}redo\_period=600} & 84.23\% &  \\
\texttt{redo|\allowbreak{}lr=0.001,\allowbreak{}redo\_tau=0.1,\allowbreak{}redo\_period=120} & 85.99\% &  \\
\texttt{redo|\allowbreak{}lr=0.001,\allowbreak{}redo\_tau=0.1,\allowbreak{}redo\_period=240} & 85.24\% &  \\
\texttt{redo|\allowbreak{}lr=0.001,\allowbreak{}redo\_tau=0.1,\allowbreak{}redo\_period=600} & 84.20\% &  \\
\texttt{l2i|\allowbreak{}lr=0.0001,\allowbreak{}l2init=0.01} & 75.72\% &  \\
\texttt{l2i|\allowbreak{}lr=0.0001,\allowbreak{}l2init=0.001} & 75.80\% &  \\
\texttt{l2i|\allowbreak{}lr=0.0001,\allowbreak{}l2init=0.0001} & 75.39\% &  \\
\texttt{l2|\allowbreak{}lr=0.001,\allowbreak{}l2zero=0.01} & 81.19\% &  \\
\texttt{l2|\allowbreak{}lr=0.001,\allowbreak{}l2zero=0.001} & 87.54\% & \textbf{selected} (interior) \\
\texttt{l2|\allowbreak{}lr=0.001,\allowbreak{}l2zero=0.0001} & 85.15\% &  \\
\texttt{l2|\allowbreak{}lr=0.0001,\allowbreak{}l2zero=0.01} & 70.46\% &  \\
\texttt{l2|\allowbreak{}lr=0.0001,\allowbreak{}l2zero=0.001} & 72.78\% &  \\
\texttt{l2|\allowbreak{}lr=0.0001,\allowbreak{}l2zero=0.0001} & 73.93\% &  \\
\texttt{snp|\allowbreak{}lr=0.0001,\allowbreak{}snp\_eps=0.001,\allowbreak{}snp\_sigma=0.01} & 61.10\% &  \\
\texttt{snp|\allowbreak{}lr=0.0001,\allowbreak{}snp\_eps=0.001,\allowbreak{}snp\_sigma=0.001} & 53.13\% &  \\
\texttt{snp|\allowbreak{}lr=0.0001,\allowbreak{}snp\_eps=0.001,\allowbreak{}snp\_sigma=0.0001} & 50.49\% &  \\
\texttt{snp|\allowbreak{}lr=0.0001,\allowbreak{}snp\_eps=0.0001,\allowbreak{}snp\_sigma=0.01} & 71.57\% &  \\
\texttt{snp|\allowbreak{}lr=0.0001,\allowbreak{}snp\_eps=0.0001,\allowbreak{}snp\_sigma=0.001} & 68.95\% &  \\
\texttt{snp|\allowbreak{}lr=0.0001,\allowbreak{}snp\_eps=0.0001,\allowbreak{}snp\_sigma=0.0001} & 68.84\% &  \\
\texttt{snp|\allowbreak{}lr=0.0001,\allowbreak{}snp\_eps=1e-05,\allowbreak{}snp\_sigma=0.01} & 76.46\% &  \\
\texttt{snp|\allowbreak{}lr=0.0001,\allowbreak{}snp\_eps=1e-05,\allowbreak{}snp\_sigma=0.001} & 74.08\% &  \\
\texttt{snp|\allowbreak{}lr=0.0001,\allowbreak{}snp\_eps=1e-05,\allowbreak{}snp\_sigma=0.0001} & 73.90\% &  \\
\texttt{cbp|\allowbreak{}lr=0.001,\allowbreak{}cbp\_rho=0.01} & 83.75\% &  \\
\texttt{cbp|\allowbreak{}lr=0.001,\allowbreak{}cbp\_rho=0.1} & 80.82\% &  \\
\texttt{cbp|\allowbreak{}lr=0.0001,\allowbreak{}cbp\_rho=0.1} & 68.33\% &  \\
\texttt{cbp|\allowbreak{}lr=0.0001,\allowbreak{}cbp\_rho=0.01} & 75.66\% &  \\
\texttt{cbp|\allowbreak{}lr=0.0001,\allowbreak{}cbp\_rho=0.001} & 70.96\% &  \\
\texttt{cbp|\allowbreak{}lr=0.0001,\allowbreak{}cbp\_rho=0.0001} & 72.79\% &  \\
\texttt{cbp|\allowbreak{}lr=0.0001,\allowbreak{}cbp\_rho=1e-05} & 74.92\% &  \\
\texttt{redo|\allowbreak{}lr=0.0001,\allowbreak{}redo\_tau=0,\allowbreak{}redo\_period=120} & 74.41\% &  \\
\texttt{redo|\allowbreak{}lr=0.0001,\allowbreak{}redo\_tau=0,\allowbreak{}redo\_period=240} & 74.20\% &  \\
\texttt{redo|\allowbreak{}lr=0.0001,\allowbreak{}redo\_tau=0,\allowbreak{}redo\_period=600} & 73.88\% &  \\
\texttt{redo|\allowbreak{}lr=0.0001,\allowbreak{}redo\_tau=0.025,\allowbreak{}redo\_period=120} & 75.66\% &  \\
\texttt{redo|\allowbreak{}lr=0.0001,\allowbreak{}redo\_tau=0.025,\allowbreak{}redo\_period=240} & 75.26\% &  \\
\texttt{redo|\allowbreak{}lr=0.0001,\allowbreak{}redo\_tau=0.025,\allowbreak{}redo\_period=600} & 74.15\% &  \\
\texttt{redo|\allowbreak{}lr=0.0001,\allowbreak{}redo\_tau=0.1,\allowbreak{}redo\_period=120} & 76.14\% &  \\
\texttt{redo|\allowbreak{}lr=0.0001,\allowbreak{}redo\_tau=0.1,\allowbreak{}redo\_period=240} & 75.46\% &  \\
\texttt{redo|\allowbreak{}lr=0.0001,\allowbreak{}redo\_tau=0.1,\allowbreak{}redo\_period=600} & 74.12\% &  \\
\texttt{ln|\allowbreak{}lr=0.001} & 68.36\% &  \\
\texttt{ln|\allowbreak{}lr=0.0001} & 68.46\% & \textbf{selected} (edge: lr) \\
\texttt{block|\allowbreak{}lr=0.001} & 92.33\% & \textbf{selected} (interior) \\
\texttt{block|\allowbreak{}lr=0.0001} & 77.70\% &  \\
\texttt{kkt1|\allowbreak{}lr=0.001} & 90.50\% & \textbf{selected} (interior) \\
\texttt{kkt1|\allowbreak{}lr=0.0001} & 80.68\% &  \\
\texttt{wcap|\allowbreak{}lr=0.001} & 78.68\% &  \\
\texttt{wcap|\allowbreak{}lr=0.0001} & 76.11\% &  \\
\texttt{none|\allowbreak{}lr=0.003} & 57.21\% &  \\
\texttt{none|\allowbreak{}lr=0.01} & 49.48\% &  \\
\texttt{l2i|\allowbreak{}lr=0.003,\allowbreak{}l2init=0.01} & 83.67\% &  \\
\texttt{l2i|\allowbreak{}lr=0.003,\allowbreak{}l2init=0.001} & 84.62\% &  \\
\texttt{l2i|\allowbreak{}lr=0.003,\allowbreak{}l2init=0.0001} & 78.81\% &  \\
\texttt{l2i|\allowbreak{}lr=0.01,\allowbreak{}l2init=0.01} & 74.87\% &  \\
\texttt{l2i|\allowbreak{}lr=0.01,\allowbreak{}l2init=0.001} & 74.50\% &  \\
\texttt{l2i|\allowbreak{}lr=0.01,\allowbreak{}l2init=0.0001} & 57.27\% &  \\
\texttt{l2|\allowbreak{}lr=0.003,\allowbreak{}l2zero=0.01} & 78.73\% &  \\
\texttt{l2|\allowbreak{}lr=0.003,\allowbreak{}l2zero=0.001} & 81.92\% &  \\
\texttt{l2|\allowbreak{}lr=0.003,\allowbreak{}l2zero=0.0001} & 77.04\% &  \\
\texttt{l2|\allowbreak{}lr=0.01,\allowbreak{}l2zero=0.01} & 52.58\% &  \\
\texttt{l2|\allowbreak{}lr=0.01,\allowbreak{}l2zero=0.001} & 61.10\% &  \\
\texttt{l2|\allowbreak{}lr=0.01,\allowbreak{}l2zero=0.0001} & 52.25\% &  \\
\texttt{snp|\allowbreak{}lr=0.003,\allowbreak{}snp\_eps=0.001,\allowbreak{}snp\_sigma=0.01} & 78.51\% &  \\
\texttt{snp|\allowbreak{}lr=0.003,\allowbreak{}snp\_eps=0.001,\allowbreak{}snp\_sigma=0.001} & 76.13\% &  \\
\texttt{snp|\allowbreak{}lr=0.003,\allowbreak{}snp\_eps=0.001,\allowbreak{}snp\_sigma=0.0001} & 74.92\% &  \\
\texttt{snp|\allowbreak{}lr=0.003,\allowbreak{}snp\_eps=0.0001,\allowbreak{}snp\_sigma=0.01} & 70.67\% &  \\
\texttt{snp|\allowbreak{}lr=0.003,\allowbreak{}snp\_eps=0.0001,\allowbreak{}snp\_sigma=0.001} & 69.75\% &  \\
\texttt{snp|\allowbreak{}lr=0.003,\allowbreak{}snp\_eps=0.0001,\allowbreak{}snp\_sigma=0.0001} & 67.54\% &  \\
\texttt{snp|\allowbreak{}lr=0.003,\allowbreak{}snp\_eps=1e-05,\allowbreak{}snp\_sigma=0.01} & 62.56\% &  \\
\texttt{snp|\allowbreak{}lr=0.003,\allowbreak{}snp\_eps=1e-05,\allowbreak{}snp\_sigma=0.001} & 60.95\% &  \\
\texttt{snp|\allowbreak{}lr=0.003,\allowbreak{}snp\_eps=1e-05,\allowbreak{}snp\_sigma=0.0001} & 62.31\% &  \\
\texttt{snp|\allowbreak{}lr=0.01,\allowbreak{}snp\_eps=0.001,\allowbreak{}snp\_sigma=0.01} & 52.29\% &  \\
\texttt{snp|\allowbreak{}lr=0.01,\allowbreak{}snp\_eps=0.001,\allowbreak{}snp\_sigma=0.001} & 49.84\% &  \\
\texttt{snp|\allowbreak{}lr=0.01,\allowbreak{}snp\_eps=0.001,\allowbreak{}snp\_sigma=0.0001} & 50.92\% &  \\
\texttt{snp|\allowbreak{}lr=0.01,\allowbreak{}snp\_eps=0.0001,\allowbreak{}snp\_sigma=0.01} & 49.61\% &  \\
\texttt{snp|\allowbreak{}lr=0.01,\allowbreak{}snp\_eps=0.0001,\allowbreak{}snp\_sigma=0.001} & 49.52\% &  \\
\texttt{snp|\allowbreak{}lr=0.01,\allowbreak{}snp\_eps=0.0001,\allowbreak{}snp\_sigma=0.0001} & 49.71\% &  \\
\texttt{snp|\allowbreak{}lr=0.01,\allowbreak{}snp\_eps=1e-05,\allowbreak{}snp\_sigma=0.01} & 49.47\% &  \\
\texttt{snp|\allowbreak{}lr=0.01,\allowbreak{}snp\_eps=1e-05,\allowbreak{}snp\_sigma=0.001} & 49.47\% &  \\
\texttt{snp|\allowbreak{}lr=0.01,\allowbreak{}snp\_eps=1e-05,\allowbreak{}snp\_sigma=0.0001} & 49.56\% &  \\
\texttt{cbp|\allowbreak{}lr=0.003,\allowbreak{}cbp\_rho=0.1} & 79.42\% &  \\
\texttt{cbp|\allowbreak{}lr=0.003,\allowbreak{}cbp\_rho=0.01} & 84.62\% &  \\
\texttt{cbp|\allowbreak{}lr=0.003,\allowbreak{}cbp\_rho=0.001} & 83.68\% &  \\
\texttt{cbp|\allowbreak{}lr=0.003,\allowbreak{}cbp\_rho=0.0001} & 77.24\% &  \\
\texttt{cbp|\allowbreak{}lr=0.003,\allowbreak{}cbp\_rho=1e-05} & 59.80\% &  \\
\texttt{cbp|\allowbreak{}lr=0.01,\allowbreak{}cbp\_rho=0.1} & 77.44\% &  \\
\texttt{cbp|\allowbreak{}lr=0.01,\allowbreak{}cbp\_rho=0.01} & 81.62\% &  \\
\texttt{cbp|\allowbreak{}lr=0.01,\allowbreak{}cbp\_rho=0.001} & 73.43\% &  \\
\texttt{cbp|\allowbreak{}lr=0.01,\allowbreak{}cbp\_rho=0.0001} & 59.03\% &  \\
\texttt{cbp|\allowbreak{}lr=0.01,\allowbreak{}cbp\_rho=1e-05} & 49.79\% &  \\
\texttt{redo|\allowbreak{}lr=0.003,\allowbreak{}redo\_tau=0,\allowbreak{}redo\_period=120} & 78.50\% &  \\
\texttt{redo|\allowbreak{}lr=0.003,\allowbreak{}redo\_tau=0,\allowbreak{}redo\_period=240} & 76.69\% &  \\
\texttt{redo|\allowbreak{}lr=0.003,\allowbreak{}redo\_tau=0,\allowbreak{}redo\_period=600} & 71.55\% &  \\
\texttt{redo|\allowbreak{}lr=0.003,\allowbreak{}redo\_tau=0.025,\allowbreak{}redo\_period=120} & 81.87\% &  \\
\texttt{redo|\allowbreak{}lr=0.003,\allowbreak{}redo\_tau=0.025,\allowbreak{}redo\_period=240} & 77.86\% &  \\
\texttt{redo|\allowbreak{}lr=0.003,\allowbreak{}redo\_tau=0.025,\allowbreak{}redo\_period=600} & 72.16\% &  \\
\texttt{redo|\allowbreak{}lr=0.003,\allowbreak{}redo\_tau=0.1,\allowbreak{}redo\_period=120} & 78.83\% &  \\
\texttt{redo|\allowbreak{}lr=0.003,\allowbreak{}redo\_tau=0.1,\allowbreak{}redo\_period=240} & 74.82\% &  \\
\texttt{redo|\allowbreak{}lr=0.003,\allowbreak{}redo\_tau=0.1,\allowbreak{}redo\_period=600} & 70.62\% &  \\
\texttt{redo|\allowbreak{}lr=0.01,\allowbreak{}redo\_tau=0,\allowbreak{}redo\_period=120} & 53.05\% &  \\
\texttt{redo|\allowbreak{}lr=0.01,\allowbreak{}redo\_tau=0,\allowbreak{}redo\_period=240} & 52.88\% &  \\
\texttt{redo|\allowbreak{}lr=0.01,\allowbreak{}redo\_tau=0,\allowbreak{}redo\_period=600} & 50.53\% &  \\
\texttt{redo|\allowbreak{}lr=0.01,\allowbreak{}redo\_tau=0.025,\allowbreak{}redo\_period=120} & 50.69\% &  \\
\texttt{redo|\allowbreak{}lr=0.01,\allowbreak{}redo\_tau=0.025,\allowbreak{}redo\_period=240} & 50.08\% &  \\
\texttt{redo|\allowbreak{}lr=0.01,\allowbreak{}redo\_tau=0.025,\allowbreak{}redo\_period=600} & 49.69\% &  \\
\texttt{redo|\allowbreak{}lr=0.01,\allowbreak{}redo\_tau=0.1,\allowbreak{}redo\_period=120} & 50.25\% &  \\
\texttt{redo|\allowbreak{}lr=0.01,\allowbreak{}redo\_tau=0.1,\allowbreak{}redo\_period=240} & 49.94\% &  \\
\texttt{redo|\allowbreak{}lr=0.01,\allowbreak{}redo\_tau=0.1,\allowbreak{}redo\_period=600} & 50.09\% &  \\
\texttt{ln|\allowbreak{}lr=0.003} & 61.23\% &  \\
\texttt{ln|\allowbreak{}lr=0.01} & 59.23\% &  \\
\texttt{block|\allowbreak{}lr=0.003,\allowbreak{}B\_clip=0.1} & 67.19\% &  \\
\texttt{block|\allowbreak{}lr=0.003} & 66.71\% &  \\
\texttt{block|\allowbreak{}lr=0.003,\allowbreak{}B\_clip=0.4} & 67.33\% &  \\
\texttt{block|\allowbreak{}lr=0.01,\allowbreak{}B\_clip=0.1} & 67.29\% &  \\
\texttt{block|\allowbreak{}lr=0.01} & 66.30\% &  \\
\texttt{block|\allowbreak{}lr=0.01,\allowbreak{}B\_clip=0.4} & 67.06\% &  \\
\texttt{block|\allowbreak{}lr=0.001,\allowbreak{}B\_clip=0.1} & 86.10\% &  \\
\texttt{block|\allowbreak{}lr=0.001,\allowbreak{}B\_clip=0.4} & 92.05\% &  \\
\texttt{block|\allowbreak{}lr=0.0001,\allowbreak{}B\_clip=0.1} & 77.48\% &  \\
\texttt{block|\allowbreak{}lr=0.0001,\allowbreak{}B\_clip=0.4} & 77.68\% &  \\
\texttt{kkt1|\allowbreak{}lr=0.003,\allowbreak{}kkt\_a0=0.3} & 88.89\% &  \\
\texttt{kkt1|\allowbreak{}lr=0.003} & 87.50\% &  \\
\texttt{kkt1|\allowbreak{}lr=0.003,\allowbreak{}kkt\_a0=1.2} & 88.21\% &  \\
\texttt{kkt1|\allowbreak{}lr=0.01,\allowbreak{}kkt\_a0=0.3} & 76.50\% &  \\
\texttt{kkt1|\allowbreak{}lr=0.01} & 77.90\% &  \\
\texttt{kkt1|\allowbreak{}lr=0.01,\allowbreak{}kkt\_a0=1.2} & 78.67\% &  \\
\texttt{kkt1|\allowbreak{}lr=0.001,\allowbreak{}kkt\_a0=0.3} & 89.18\% &  \\
\texttt{kkt1|\allowbreak{}lr=0.001,\allowbreak{}kkt\_a0=1.2} & 90.17\% &  \\
\texttt{kkt1|\allowbreak{}lr=0.0001,\allowbreak{}kkt\_a0=0.3} & 79.09\% &  \\
\texttt{kkt1|\allowbreak{}lr=0.0001,\allowbreak{}kkt\_a0=1.2} & 80.02\% &  \\
\texttt{wcap|\allowbreak{}lr=0.003,\allowbreak{}wcap\_k=0.5} & 66.86\% &  \\
\texttt{wcap|\allowbreak{}lr=0.003} & 62.05\% &  \\
\texttt{wcap|\allowbreak{}lr=0.003,\allowbreak{}wcap\_k=2} & 63.51\% &  \\
\texttt{wcap|\allowbreak{}lr=0.01,\allowbreak{}wcap\_k=0.5} & 49.80\% &  \\
\texttt{wcap|\allowbreak{}lr=0.01} & 49.53\% &  \\
\texttt{wcap|\allowbreak{}lr=0.01,\allowbreak{}wcap\_k=2} & 50.10\% &  \\
\texttt{wcap|\allowbreak{}lr=0.001,\allowbreak{}wcap\_k=0.5} & 78.81\% &  \\
\texttt{wcap|\allowbreak{}lr=0.001,\allowbreak{}wcap\_k=2} & 80.72\% & \textbf{selected} (edge: wcap\_k) \\
\texttt{wcap|\allowbreak{}lr=0.0001,\allowbreak{}wcap\_k=0.5} & 75.85\% &  \\
\texttt{wcap|\allowbreak{}lr=0.0001,\allowbreak{}wcap\_k=2} & 76.31\% &  \\
\texttt{crelu|\allowbreak{}lr=0.001} & 88.55\% & \textbf{selected} (interior) \\
\texttt{crelu|\allowbreak{}lr=0.0001} & 79.40\% &  \\
\texttt{crelu|\allowbreak{}lr=0.003} & 88.51\% &  \\
\texttt{crelu|\allowbreak{}lr=0.01} & 77.76\% &  \\
\bottomrule
\end{longtable}}

\textbf{Permuted MNIST (total average online on calibration seeds 100--102; 152 points; 0 diverged; 0 missing; no method with $3\times10^{-2}$ added)}

{\scriptsize\setlength{\tabcolsep}{3pt}
\begin{longtable}{@{}P{8.0cm}P{3.0cm}P{4.0cm}@{}}
\toprule
\textbf{Setting} & \textbf{Total avg. online} & \textbf{Selection} \\
\midrule
\endhead
\texttt{none|\allowbreak{}lr=0.001} & 78.82\% & \textbf{selected} (interior) \\
\texttt{none|\allowbreak{}lr=0.0001} & 74.24\% &  \\
\texttt{l2i|\allowbreak{}lr=0.001,\allowbreak{}l2init=0.01} & 86.78\% &  \\
\texttt{l2i|\allowbreak{}lr=0.001,\allowbreak{}l2init=0.001} & 88.37\% & \textbf{selected} (interior) \\
\texttt{l2i|\allowbreak{}lr=0.001,\allowbreak{}l2init=0.0001} & 86.53\% &  \\
\texttt{snp|\allowbreak{}lr=0.001,\allowbreak{}snp\_eps=0.001,\allowbreak{}snp\_sigma=0.01} & 86.17\% &  \\
\texttt{snp|\allowbreak{}lr=0.001,\allowbreak{}snp\_eps=0.001,\allowbreak{}snp\_sigma=0.001} & 85.88\% &  \\
\texttt{snp|\allowbreak{}lr=0.001,\allowbreak{}snp\_eps=0.001,\allowbreak{}snp\_sigma=0.0001} & 85.71\% &  \\
\texttt{snp|\allowbreak{}lr=0.001,\allowbreak{}snp\_eps=0.0001,\allowbreak{}snp\_sigma=0.01} & 86.52\% & \textbf{selected} (edge: snp\_sigma) \\
\texttt{snp|\allowbreak{}lr=0.001,\allowbreak{}snp\_eps=0.0001,\allowbreak{}snp\_sigma=0.001} & 86.09\% &  \\
\texttt{snp|\allowbreak{}lr=0.001,\allowbreak{}snp\_eps=0.0001,\allowbreak{}snp\_sigma=0.0001} & 85.82\% &  \\
\texttt{snp|\allowbreak{}lr=0.001,\allowbreak{}snp\_eps=1e-05,\allowbreak{}snp\_sigma=0.01} & 82.81\% &  \\
\texttt{snp|\allowbreak{}lr=0.001,\allowbreak{}snp\_eps=1e-05,\allowbreak{}snp\_sigma=0.001} & 82.80\% &  \\
\texttt{snp|\allowbreak{}lr=0.001,\allowbreak{}snp\_eps=1e-05,\allowbreak{}snp\_sigma=0.0001} & 82.72\% &  \\
\texttt{cbp|\allowbreak{}lr=0.001,\allowbreak{}cbp\_rho=0.001} & 87.42\% & \textbf{selected} (interior) \\
\texttt{cbp|\allowbreak{}lr=0.001,\allowbreak{}cbp\_rho=0.0001} & 86.09\% &  \\
\texttt{cbp|\allowbreak{}lr=0.001,\allowbreak{}cbp\_rho=1e-05} & 83.04\% &  \\
\texttt{redo|\allowbreak{}lr=0.001,\allowbreak{}redo\_tau=0,\allowbreak{}redo\_period=625} & 87.61\% &  \\
\texttt{redo|\allowbreak{}lr=0.001,\allowbreak{}redo\_tau=0,\allowbreak{}redo\_period=1250} & 87.17\% &  \\
\texttt{redo|\allowbreak{}lr=0.001,\allowbreak{}redo\_tau=0,\allowbreak{}redo\_period=3125} & 86.27\% &  \\
\texttt{redo|\allowbreak{}lr=0.001,\allowbreak{}redo\_tau=0.025,\allowbreak{}redo\_period=625} & 87.68\% &  \\
\texttt{redo|\allowbreak{}lr=0.001,\allowbreak{}redo\_tau=0.025,\allowbreak{}redo\_period=1250} & 87.27\% &  \\
\texttt{redo|\allowbreak{}lr=0.001,\allowbreak{}redo\_tau=0.025,\allowbreak{}redo\_period=3125} & 86.40\% &  \\
\texttt{redo|\allowbreak{}lr=0.001,\allowbreak{}redo\_tau=0.1,\allowbreak{}redo\_period=625} & 87.82\% & \textbf{selected} (edge: redo\_tau, redo\_period) \\
\texttt{redo|\allowbreak{}lr=0.001,\allowbreak{}redo\_tau=0.1,\allowbreak{}redo\_period=1250} & 87.36\% &  \\
\texttt{redo|\allowbreak{}lr=0.001,\allowbreak{}redo\_tau=0.1,\allowbreak{}redo\_period=3125} & 86.41\% &  \\
\texttt{l2i|\allowbreak{}lr=0.0001,\allowbreak{}l2init=0.01} & 80.89\% &  \\
\texttt{l2i|\allowbreak{}lr=0.0001,\allowbreak{}l2init=0.001} & 79.70\% &  \\
\texttt{l2i|\allowbreak{}lr=0.0001,\allowbreak{}l2init=0.0001} & 77.12\% &  \\
\texttt{l2|\allowbreak{}lr=0.001,\allowbreak{}l2zero=0.01} & 85.71\% &  \\
\texttt{l2|\allowbreak{}lr=0.001,\allowbreak{}l2zero=0.001} & 87.70\% & \textbf{selected} (interior) \\
\texttt{l2|\allowbreak{}lr=0.001,\allowbreak{}l2zero=0.0001} & 85.98\% &  \\
\texttt{l2|\allowbreak{}lr=0.0001,\allowbreak{}l2zero=0.01} & 79.13\% &  \\
\texttt{l2|\allowbreak{}lr=0.0001,\allowbreak{}l2zero=0.001} & 79.26\% &  \\
\texttt{l2|\allowbreak{}lr=0.0001,\allowbreak{}l2zero=0.0001} & 77.30\% &  \\
\texttt{snp|\allowbreak{}lr=0.0001,\allowbreak{}snp\_eps=0.001,\allowbreak{}snp\_sigma=0.01} & 62.70\% &  \\
\texttt{snp|\allowbreak{}lr=0.0001,\allowbreak{}snp\_eps=0.001,\allowbreak{}snp\_sigma=0.001} & 62.94\% &  \\
\texttt{snp|\allowbreak{}lr=0.0001,\allowbreak{}snp\_eps=0.001,\allowbreak{}snp\_sigma=0.0001} & 62.95\% &  \\
\texttt{snp|\allowbreak{}lr=0.0001,\allowbreak{}snp\_eps=0.0001,\allowbreak{}snp\_sigma=0.01} & 76.20\% &  \\
\texttt{snp|\allowbreak{}lr=0.0001,\allowbreak{}snp\_eps=0.0001,\allowbreak{}snp\_sigma=0.001} & 75.41\% &  \\
\texttt{snp|\allowbreak{}lr=0.0001,\allowbreak{}snp\_eps=0.0001,\allowbreak{}snp\_sigma=0.0001} & 75.07\% &  \\
\texttt{snp|\allowbreak{}lr=0.0001,\allowbreak{}snp\_eps=1e-05,\allowbreak{}snp\_sigma=0.01} & 77.12\% &  \\
\texttt{snp|\allowbreak{}lr=0.0001,\allowbreak{}snp\_eps=1e-05,\allowbreak{}snp\_sigma=0.001} & 78.40\% &  \\
\texttt{snp|\allowbreak{}lr=0.0001,\allowbreak{}snp\_eps=1e-05,\allowbreak{}snp\_sigma=0.0001} & 78.48\% &  \\
\texttt{cbp|\allowbreak{}lr=0.001,\allowbreak{}cbp\_rho=0.01} & 85.98\% &  \\
\texttt{cbp|\allowbreak{}lr=0.001,\allowbreak{}cbp\_rho=0.1} & 75.39\% &  \\
\texttt{cbp|\allowbreak{}lr=0.0001,\allowbreak{}cbp\_rho=0.1} & 35.84\% &  \\
\texttt{cbp|\allowbreak{}lr=0.0001,\allowbreak{}cbp\_rho=0.01} & 68.59\% &  \\
\texttt{cbp|\allowbreak{}lr=0.0001,\allowbreak{}cbp\_rho=0.001} & 75.03\% &  \\
\texttt{cbp|\allowbreak{}lr=0.0001,\allowbreak{}cbp\_rho=0.0001} & 76.28\% &  \\
\texttt{cbp|\allowbreak{}lr=0.0001,\allowbreak{}cbp\_rho=1e-05} & 76.38\% &  \\
\texttt{redo|\allowbreak{}lr=0.0001,\allowbreak{}redo\_tau=0,\allowbreak{}redo\_period=625} & 77.19\% &  \\
\texttt{redo|\allowbreak{}lr=0.0001,\allowbreak{}redo\_tau=0,\allowbreak{}redo\_period=1250} & 77.26\% &  \\
\texttt{redo|\allowbreak{}lr=0.0001,\allowbreak{}redo\_tau=0,\allowbreak{}redo\_period=3125} & 77.15\% &  \\
\texttt{redo|\allowbreak{}lr=0.0001,\allowbreak{}redo\_tau=0.025,\allowbreak{}redo\_period=625} & 77.37\% &  \\
\texttt{redo|\allowbreak{}lr=0.0001,\allowbreak{}redo\_tau=0.025,\allowbreak{}redo\_period=1250} & 77.39\% &  \\
\texttt{redo|\allowbreak{}lr=0.0001,\allowbreak{}redo\_tau=0.025,\allowbreak{}redo\_period=3125} & 77.24\% &  \\
\texttt{redo|\allowbreak{}lr=0.0001,\allowbreak{}redo\_tau=0.1,\allowbreak{}redo\_period=625} & 77.23\% &  \\
\texttt{redo|\allowbreak{}lr=0.0001,\allowbreak{}redo\_tau=0.1,\allowbreak{}redo\_period=1250} & 76.99\% &  \\
\texttt{redo|\allowbreak{}lr=0.0001,\allowbreak{}redo\_tau=0.1,\allowbreak{}redo\_period=3125} & 76.87\% &  \\
\texttt{ln|\allowbreak{}lr=0.001} & 80.53\% & \textbf{selected} (interior) \\
\texttt{ln|\allowbreak{}lr=0.0001} & 78.39\% &  \\
\texttt{block|\allowbreak{}lr=0.001} & 84.67\% &  \\
\texttt{block|\allowbreak{}lr=0.0001} & 75.50\% &  \\
\texttt{kkt1|\allowbreak{}lr=0.001} & 84.97\% & \textbf{selected} (interior) \\
\texttt{kkt1|\allowbreak{}lr=0.0001} & 76.35\% &  \\
\texttt{none|\allowbreak{}lr=0.003} & 75.38\% &  \\
\texttt{none|\allowbreak{}lr=0.01} & 51.34\% &  \\
\texttt{l2i|\allowbreak{}lr=0.003,\allowbreak{}l2init=0.01} & 85.25\% &  \\
\texttt{l2i|\allowbreak{}lr=0.003,\allowbreak{}l2init=0.001} & 87.70\% &  \\
\texttt{l2i|\allowbreak{}lr=0.003,\allowbreak{}l2init=0.0001} & 86.17\% &  \\
\texttt{l2i|\allowbreak{}lr=0.01,\allowbreak{}l2init=0.01} & 80.46\% &  \\
\texttt{l2i|\allowbreak{}lr=0.01,\allowbreak{}l2init=0.001} & 82.83\% &  \\
\texttt{l2i|\allowbreak{}lr=0.01,\allowbreak{}l2init=0.0001} & 82.72\% &  \\
\texttt{l2|\allowbreak{}lr=0.003,\allowbreak{}l2zero=0.01} & 85.06\% &  \\
\texttt{l2|\allowbreak{}lr=0.003,\allowbreak{}l2zero=0.001} & 87.03\% &  \\
\texttt{l2|\allowbreak{}lr=0.003,\allowbreak{}l2zero=0.0001} & 85.23\% &  \\
\texttt{l2|\allowbreak{}lr=0.01,\allowbreak{}l2zero=0.01} & 80.46\% &  \\
\texttt{l2|\allowbreak{}lr=0.01,\allowbreak{}l2zero=0.001} & 82.14\% &  \\
\texttt{l2|\allowbreak{}lr=0.01,\allowbreak{}l2zero=0.0001} & 80.15\% &  \\
\texttt{snp|\allowbreak{}lr=0.003,\allowbreak{}snp\_eps=0.001,\allowbreak{}snp\_sigma=0.01} & 86.17\% &  \\
\texttt{snp|\allowbreak{}lr=0.003,\allowbreak{}snp\_eps=0.001,\allowbreak{}snp\_sigma=0.001} & 85.80\% &  \\
\texttt{snp|\allowbreak{}lr=0.003,\allowbreak{}snp\_eps=0.001,\allowbreak{}snp\_sigma=0.0001} & 85.69\% &  \\
\texttt{snp|\allowbreak{}lr=0.003,\allowbreak{}snp\_eps=0.0001,\allowbreak{}snp\_sigma=0.01} & 84.03\% &  \\
\texttt{snp|\allowbreak{}lr=0.003,\allowbreak{}snp\_eps=0.0001,\allowbreak{}snp\_sigma=0.001} & 83.54\% &  \\
\texttt{snp|\allowbreak{}lr=0.003,\allowbreak{}snp\_eps=0.0001,\allowbreak{}snp\_sigma=0.0001} & 83.37\% &  \\
\texttt{snp|\allowbreak{}lr=0.003,\allowbreak{}snp\_eps=1e-05,\allowbreak{}snp\_sigma=0.01} & 79.11\% &  \\
\texttt{snp|\allowbreak{}lr=0.003,\allowbreak{}snp\_eps=1e-05,\allowbreak{}snp\_sigma=0.001} & 78.71\% &  \\
\texttt{snp|\allowbreak{}lr=0.003,\allowbreak{}snp\_eps=1e-05,\allowbreak{}snp\_sigma=0.0001} & 78.66\% &  \\
\texttt{snp|\allowbreak{}lr=0.01,\allowbreak{}snp\_eps=0.001,\allowbreak{}snp\_sigma=0.01} & 82.01\% &  \\
\texttt{snp|\allowbreak{}lr=0.01,\allowbreak{}snp\_eps=0.001,\allowbreak{}snp\_sigma=0.001} & 81.84\% &  \\
\texttt{snp|\allowbreak{}lr=0.01,\allowbreak{}snp\_eps=0.001,\allowbreak{}snp\_sigma=0.0001} & 81.85\% &  \\
\texttt{snp|\allowbreak{}lr=0.01,\allowbreak{}snp\_eps=0.0001,\allowbreak{}snp\_sigma=0.01} & 72.69\% &  \\
\texttt{snp|\allowbreak{}lr=0.01,\allowbreak{}snp\_eps=0.0001,\allowbreak{}snp\_sigma=0.001} & 72.20\% &  \\
\texttt{snp|\allowbreak{}lr=0.01,\allowbreak{}snp\_eps=0.0001,\allowbreak{}snp\_sigma=0.0001} & 71.41\% &  \\
\texttt{snp|\allowbreak{}lr=0.01,\allowbreak{}snp\_eps=1e-05,\allowbreak{}snp\_sigma=0.01} & 61.11\% &  \\
\texttt{snp|\allowbreak{}lr=0.01,\allowbreak{}snp\_eps=1e-05,\allowbreak{}snp\_sigma=0.001} & 61.18\% &  \\
\texttt{snp|\allowbreak{}lr=0.01,\allowbreak{}snp\_eps=1e-05,\allowbreak{}snp\_sigma=0.0001} & 58.16\% &  \\
\texttt{cbp|\allowbreak{}lr=0.003,\allowbreak{}cbp\_rho=0.1} & 80.42\% &  \\
\texttt{cbp|\allowbreak{}lr=0.003,\allowbreak{}cbp\_rho=0.01} & 86.29\% &  \\
\texttt{cbp|\allowbreak{}lr=0.003,\allowbreak{}cbp\_rho=0.001} & 86.42\% &  \\
\texttt{cbp|\allowbreak{}lr=0.003,\allowbreak{}cbp\_rho=0.0001} & 84.17\% &  \\
\texttt{cbp|\allowbreak{}lr=0.003,\allowbreak{}cbp\_rho=1e-05} & 80.37\% &  \\
\texttt{cbp|\allowbreak{}lr=0.01,\allowbreak{}cbp\_rho=0.1} & 80.81\% &  \\
\texttt{cbp|\allowbreak{}lr=0.01,\allowbreak{}cbp\_rho=0.01} & 83.79\% &  \\
\texttt{cbp|\allowbreak{}lr=0.01,\allowbreak{}cbp\_rho=0.001} & 81.69\% &  \\
\texttt{cbp|\allowbreak{}lr=0.01,\allowbreak{}cbp\_rho=0.0001} & 77.54\% &  \\
\texttt{cbp|\allowbreak{}lr=0.01,\allowbreak{}cbp\_rho=1e-05} & 71.22\% &  \\
\texttt{redo|\allowbreak{}lr=0.003,\allowbreak{}redo\_tau=0,\allowbreak{}redo\_period=625} & 86.82\% &  \\
\texttt{redo|\allowbreak{}lr=0.003,\allowbreak{}redo\_tau=0,\allowbreak{}redo\_period=1250} & 86.01\% &  \\
\texttt{redo|\allowbreak{}lr=0.003,\allowbreak{}redo\_tau=0,\allowbreak{}redo\_period=3125} & 84.61\% &  \\
\texttt{redo|\allowbreak{}lr=0.003,\allowbreak{}redo\_tau=0.025,\allowbreak{}redo\_period=625} & 86.84\% &  \\
\texttt{redo|\allowbreak{}lr=0.003,\allowbreak{}redo\_tau=0.025,\allowbreak{}redo\_period=1250} & 86.09\% &  \\
\texttt{redo|\allowbreak{}lr=0.003,\allowbreak{}redo\_tau=0.025,\allowbreak{}redo\_period=3125} & 84.61\% &  \\
\texttt{redo|\allowbreak{}lr=0.003,\allowbreak{}redo\_tau=0.1,\allowbreak{}redo\_period=625} & 86.90\% &  \\
\texttt{redo|\allowbreak{}lr=0.003,\allowbreak{}redo\_tau=0.1,\allowbreak{}redo\_period=1250} & 86.06\% &  \\
\texttt{redo|\allowbreak{}lr=0.003,\allowbreak{}redo\_tau=0.1,\allowbreak{}redo\_period=3125} & 84.63\% &  \\
\texttt{redo|\allowbreak{}lr=0.01,\allowbreak{}redo\_tau=0,\allowbreak{}redo\_period=625} & 80.26\% &  \\
\texttt{redo|\allowbreak{}lr=0.01,\allowbreak{}redo\_tau=0,\allowbreak{}redo\_period=1250} & 77.97\% &  \\
\texttt{redo|\allowbreak{}lr=0.01,\allowbreak{}redo\_tau=0,\allowbreak{}redo\_period=3125} & 75.23\% &  \\
\texttt{redo|\allowbreak{}lr=0.01,\allowbreak{}redo\_tau=0.025,\allowbreak{}redo\_period=625} & 79.99\% &  \\
\texttt{redo|\allowbreak{}lr=0.01,\allowbreak{}redo\_tau=0.025,\allowbreak{}redo\_period=1250} & 77.78\% &  \\
\texttt{redo|\allowbreak{}lr=0.01,\allowbreak{}redo\_tau=0.025,\allowbreak{}redo\_period=3125} & 75.06\% &  \\
\texttt{redo|\allowbreak{}lr=0.01,\allowbreak{}redo\_tau=0.1,\allowbreak{}redo\_period=625} & 80.19\% &  \\
\texttt{redo|\allowbreak{}lr=0.01,\allowbreak{}redo\_tau=0.1,\allowbreak{}redo\_period=1250} & 78.15\% &  \\
\texttt{redo|\allowbreak{}lr=0.01,\allowbreak{}redo\_tau=0.1,\allowbreak{}redo\_period=3125} & 75.28\% &  \\
\texttt{ln|\allowbreak{}lr=0.003} & 79.43\% &  \\
\texttt{ln|\allowbreak{}lr=0.01} & 75.67\% &  \\
\texttt{block|\allowbreak{}lr=0.003,\allowbreak{}B\_lam=0.08565} & 84.96\% & \textbf{selected} (edge: B\_lam) \\
\texttt{block|\allowbreak{}lr=0.003} & 84.91\% &  \\
\texttt{block|\allowbreak{}lr=0.003,\allowbreak{}B\_lam=0.3426} & 84.89\% &  \\
\texttt{block|\allowbreak{}lr=0.01,\allowbreak{}B\_lam=0.08565} & 79.32\% &  \\
\texttt{block|\allowbreak{}lr=0.01} & 79.33\% &  \\
\texttt{block|\allowbreak{}lr=0.01,\allowbreak{}B\_lam=0.3426} & 79.27\% &  \\
\texttt{block|\allowbreak{}lr=0.001,\allowbreak{}B\_lam=0.08565} & 84.69\% &  \\
\texttt{block|\allowbreak{}lr=0.001,\allowbreak{}B\_lam=0.3426} & 84.72\% &  \\
\texttt{block|\allowbreak{}lr=0.0001,\allowbreak{}B\_lam=0.08565} & 75.65\% &  \\
\texttt{block|\allowbreak{}lr=0.0001,\allowbreak{}B\_lam=0.3426} & 75.40\% &  \\
\texttt{kkt1|\allowbreak{}lr=0.003,\allowbreak{}kkt\_a0=0.3} & 81.43\% &  \\
\texttt{kkt1|\allowbreak{}lr=0.003} & 84.01\% &  \\
\texttt{kkt1|\allowbreak{}lr=0.003,\allowbreak{}kkt\_a0=1.2} & 82.63\% &  \\
\texttt{kkt1|\allowbreak{}lr=0.01,\allowbreak{}kkt\_a0=0.3} & 76.00\% &  \\
\texttt{kkt1|\allowbreak{}lr=0.01} & 74.72\% &  \\
\texttt{kkt1|\allowbreak{}lr=0.01,\allowbreak{}kkt\_a0=1.2} & 73.62\% &  \\
\texttt{kkt1|\allowbreak{}lr=0.001,\allowbreak{}kkt\_a0=0.3} & 83.89\% &  \\
\texttt{kkt1|\allowbreak{}lr=0.001,\allowbreak{}kkt\_a0=1.2} & 84.02\% &  \\
\texttt{kkt1|\allowbreak{}lr=0.0001,\allowbreak{}kkt\_a0=0.3} & 76.42\% &  \\
\texttt{kkt1|\allowbreak{}lr=0.0001,\allowbreak{}kkt\_a0=1.2} & 75.86\% &  \\
\texttt{crelu|\allowbreak{}lr=0.001} & 84.99\% & \textbf{selected} (interior) \\
\texttt{crelu|\allowbreak{}lr=0.0001} & 76.95\% &  \\
\texttt{crelu|\allowbreak{}lr=0.003} & 83.43\% &  \\
\texttt{crelu|\allowbreak{}lr=0.01} & 78.88\% &  \\
\bottomrule
\end{longtable}}

\subsection{Split of the switch gradient (Eq.~4; last hidden layer; unmodified network at learning rate 0.001; observation)}
\label{app:J8}
agree = fraction of sign agreement between the actual gradient and the common-error term; ratioL = median of $\lvert L\rvert/\lvert S^{\rm rand}\rvert$; sink = fraction in the lowering direction; live = fraction of live units (``switch push'' in the log labels = the switch gradient). Continual ImageNet, then Permuted MNIST (excerpts):

\begin{Verbatim}[fontsize=\scriptsize,breaklines=true,breakanywhere=true]
## 5. switch push split (App. C.7) on the last hidden layer, A0@1e-3: early tasks 2-11 / late 491-500 (eval-seed means)
  early 2-11: agree 0.883 ratioL 0.482 sink_real 0.692 sink_rand 0.692 live 0.810
  late 491-500: agree 0.214 sink_real 0.181 sink_rand 0.239 live 0.096 ratioL 2.072
\end{Verbatim}

\begin{Verbatim}[fontsize=\scriptsize,breaklines=true,breakanywhere=true]
## 5. switch push split (App. C.7) on the last hidden layer, A0@1e-3: early tasks 2-11 / late 491-500 (eval-seed means)
  early 2-11: agree 0.889 ratioL 0.323 sink_real 0.659 sink_rand 0.684 live 0.911
  late 491-500: agree 0.796 ratioL 0.454 sink_real 0.512 sink_rand 0.646 live 0.107
\end{Verbatim}

\subsection{Derivative swap (standard sequences)}
\label{app:J9}
Recipient = the unmodified network at learning rate 0.001. restore = the derivative of the network at the end of task 10 is restored to the network at the end of task 490 (Adam history carried over); reverse = the derivative at the end of task 490 is given to the network at the end of task 10 (Adam initialized for both the control and the intervention). Continual ImageNet: all 4 hidden layers; Permuted MNIST: the second hidden layer (``response swap'' in the log labels = the derivative swap). Continual ImageNet, then Permuted MNIST (excerpts):

\begin{Verbatim}[fontsize=\scriptsize,breaklines=true,breakanywhere=true]
## 7. response swap (App. E.1): swap - ctrl, same seed, online on the next task(s)
restore task 491: ctrl 0.5332 | swap - ctrl +0.1574 [+0.1065, +0.2082] pos 10/10 improved
restore task 492: ctrl 0.5339 | swap - ctrl +0.1270 [+0.0500, +0.2040] pos 9/10 mostly improved
reverse task 11: ctrl 0.7829 | swap - ctrl -0.2122 [-0.2848, -0.1396] pos 0/10 worse
reverse task 12: ctrl 0.8438 | swap - ctrl -0.1725 [-0.2329, -0.1122] pos 0/10 worse
\end{Verbatim}

\begin{Verbatim}[fontsize=\scriptsize,breaklines=true,breakanywhere=true]
## 7. response swap (App. E.1): swap - ctrl, same seed, online on the next task(s)
restore task 491: ctrl 0.7566 | swap - ctrl -0.0340 [-0.0590, -0.0090] pos 1/10 mostly worse
restore task 492: ctrl 0.7531 | swap - ctrl +0.0304 [+0.0184, +0.0425] pos 9/10 mostly improved
reverse task 11: ctrl 0.8109 | swap - ctrl -0.1253 [-0.1374, -0.1132] pos 0/10 worse
reverse task 12: ctrl 0.8574 | swap - ctrl -0.0473 [-0.0549, -0.0396] pos 0/10 worse
\end{Verbatim}

\begin{itemize}
\item The numbers of §6.2 (Continual ImageNet +15.7 and −21.2; Permuted MNIST −12.5 and −3.4) were rounded once from the raw logs of the swaps. ``The second task went in the same direction'' of §6.2 refers to the rows restore task 492 (+0.1270, 9/10) and reverse task 12 (−0.1725, 0/10) of Continual ImageNet in the excerpt above. ``About 55\%'' of §6.2 = +0.1574 of restore task 491 ÷ (early window 0.8176 − control at task 491 0.5332) = 0.553.
\end{itemize}

\subsection{Error-mean removal and removal of the absent-class term (standard sequences; [arm − unmodified network at learning rate 0.001]; task 2, tasks 2--6, tasks 16--20, late window)}
\label{app:J10}
t3 = error-mean removal (T3 of App.~G); abs = removal of the absent-class term. Continual ImageNet (t3), then Permuted MNIST (abs, t3) (excerpts):

\begin{Verbatim}[fontsize=\scriptsize,breaklines=true,breakanywhere=true]
t3|lr=0.001 t2 -0.2091 [-0.2846, -0.1336] pos 0/10 worse || t2-6 -0.2233 [-0.2693, -0.1773] pos 0/10 worse || t16-20 -0.2803 [-0.3210, -0.2396] pos 0/10 worse || late -0.0463 [-0.1074, +0.0148] pos 6/10 inconsistent/small
\end{Verbatim}

\begin{Verbatim}[fontsize=\scriptsize,breaklines=true,breakanywhere=true]
abs|lr=0.001 t2 -0.0031 [-0.0045, -0.0016] pos 0/10 inconsistent/small || t2-6 -0.0028 [-0.0037, -0.0018] pos 0/10 inconsistent/small || t16-20 -0.0064 [-0.0088, -0.0039] pos 0/10 worse || late -0.0054 [-0.0092, -0.0016] pos 1/10 mostly worse
\end{Verbatim}

\begin{Verbatim}[fontsize=\scriptsize,breaklines=true,breakanywhere=true]
t3|lr=0.001 t2 -0.0063 [-0.0084, -0.0043] pos 0/10 worse || t2-6 -0.0086 [-0.0098, -0.0075] pos 0/10 worse || t16-20 +0.0025 [+0.0006, +0.0044] pos 7/10 inconsistent/small || late -0.0083 [-0.0123, -0.0044] pos 0/10 worse
\end{Verbatim}

\subsection{Online accuracy of networks trained from scratch (tasks 100--500; centering + bias suppression − each method; same-seed pairs; selection of runs A and B at learning rates \{0.001, 0.0001\}; §6.3)}
\label{app:J11}
Network trained from scratch = a network trained on that task from initialization by the same method's rule (Table~\ref{tab:terms}). Centering + bias suppression learns faster from initialization. In the selection at learning rates \{0.001, 0.0001\}, about half of the difference in total average online accuracy on Continual ImageNet (+3.1 over L2 Init) corresponds to this speed difference. The speed part of the pairs in the selection with 4 learning rates (run C) is ([post hoc]; excerpt):

\begin{Verbatim}[fontsize=\scriptsize,breaklines=true,breakanywhere=true]
# [post hoc] ci: [centering + bias suppression C:rep:block|lr=0.001] - each method, same-seed pairs: difference in total average online accuracy and the share of the difference between networks trained from scratch (5 points, tasks 100-500)
l2i (C:rep:l2i|lr=0.001,l2init=0.001): total diff +2.390 pt | fresh diff mean +1.965 pt (5 points: +2.17 +1.88 +2.19 +1.73 +1.86) | speed share on tasks 100-500 0.659 / on tasks 2-500 0.821 | fresh diff > 0 in 10/10 seeds
l2 (B:l2|lr=0.001,l2zero=0.001): total diff +4.425 pt | fresh diff mean +2.282 pt (5 points: +2.56 +2.26 +2.46 +1.98 +2.15) | speed share on tasks 100-500 0.414 / on tasks 2-500 0.515 | fresh diff > 0 in 10/10 seeds
snp (A:snp|lr=0.001,snp_eps=0.0001,snp_sigma=0.01): total diff +8.519 pt | fresh diff mean +1.998 pt (5 points: +2.13 +1.98 +2.29 +1.78 +1.81) | speed share on tasks 100-500 0.188 / on tasks 2-500 0.234 | fresh diff > 0 in 10/10 seeds
cbp (A:cbp|lr=0.001,cbp_rho=0.001): total diff +6.313 pt | fresh diff mean +1.925 pt (5 points: +2.11 +1.88 +2.17 +1.70 +1.77) | speed share on tasks 100-500 0.245 / on tasks 2-500 0.304 | fresh diff > 0 in 10/10 seeds
redo (A:redo|lr=0.001,redo_tau=0.025,redo_period=120): total diff +5.590 pt | fresh diff mean +1.928 pt (5 points: +2.08 +1.90 +2.19 +1.69 +1.77) | speed share on tasks 100-500 0.277 / on tasks 2-500 0.344 | fresh diff > 0 in 10/10 seeds
ln (B:ln|lr=0.0001): total diff +23.151 pt | fresh diff mean +2.377 pt (5 points: +2.69 +1.86 +3.27 +2.25 +1.82) | speed share on tasks 100-500 0.082 / on tasks 2-500 0.102 | fresh diff > 0 in 10/10 seeds
crelu (B:crelu|lr=0.001): total diff +3.283 pt | fresh diff mean +1.007 pt (5 points: +1.13 +1.32 +1.04 +0.74 +0.81) | speed share on tasks 100-500 0.246 / on tasks 2-500 0.306 | fresh diff > 0 in 10/10 seeds
\end{Verbatim}

Breakdown (extracted from the raw logs): mean difference over tasks 1--99 +4.73 (30\% of the summed difference; networks trained from scratch were not measured in this range); mean difference over tasks 100--500 +2.74 (70\%). Applying the from-scratch difference (mean of the 5 points below, +1.96) to tasks 100--500 gives 50\% of the total, and to tasks 2--500 gives 62\%. The difference in preservation (difference of the gaps to scratch; positive = centering + bias suppression preserves more) is +1.10, +1.55 and +0.56 at tasks 100, 200 and 300 (6, 7 and 4/10), +0.42 and −0.19 at tasks 400 and 500 (7 and 4/10), and +0.11 for the mean of tasks 400 and 500 (5/10). That centering speeds up learning is classical knowledge (the abstracts of \citealp{schraudolph1998centering} and \citealp{raiko2012deep}), and kunekune, which does not center, also learns faster than L2 Init when trained from scratch, by +5.77 to +6.26 (10/10 each; same analysis). Excerpt:

\begin{Verbatim}[fontsize=\scriptsize,breaklines=true,breakanywhere=true]
ci fresh block - l2i|lr=0.001,l2init=0.001: t100 +2.17 (10/10) | t200 +1.85 (10/10) | t300 +2.22 (10/10) | t400 +1.74 (10/10) | t500 +1.85 (10/10) | 5 tasks x 10 seeds positive 50/50 mean +1.96
ci fresh block - none|lr=0.001: t100 +2.09 (10/10) | t200 +1.87 (10/10) | t300 +2.17 (10/10) | t400 +1.70 (10/10) | t500 +1.78 (10/10) | 5 tasks x 10 seeds positive 50/50 mean +1.92
ci fresh block - l2|lr=0.001,l2zero=0.001: t100 +2.57 (10/10) | t200 +2.24 (10/10) | t300 +2.46 (10/10) | t400 +1.97 (10/10) | t500 +2.15 (10/10) | 5 tasks x 10 seeds positive 50/50 mean +2.28
ci fresh block - crelu|lr=0.001: t100 +1.14 (8/10) | t200 +1.30 (8/10) | t300 +1.04 (9/10) | t400 +0.73 (9/10) | t500 +0.80 (8/10) | 5 tasks x 10 seeds positive 42/50 mean +1.00
pm fresh block - l2i|lr=0.001,l2init=0.001: t100 +1.28 (10/10) | t200 +1.12 (10/10) | t300 +1.32 (10/10) | t400 +1.28 (10/10) | t500 +1.16 (10/10) | 5 tasks x 10 seeds positive 50/50 mean +1.23
pm fresh block - none|lr=0.001: t100 +1.12 (10/10) | t200 +1.01 (10/10) | t300 +1.12 (10/10) | t400 +1.14 (10/10) | t500 +1.04 (10/10) | 5 tasks x 10 seeds positive 50/50 mean +1.09
\end{Verbatim}

\subsection{Fraction of dead units (selection of runs A and B at learning rates \{0.001, 0.0001\}; mean of evaluation seeds 0--9; \%; §6.2)}
\label{app:J12}
Death = the fraction of units with $\lvert\varphi'\rvert<10^{-6}$ on all diagnostic inputs of the task (the task probe) (for convolutions, per channel including all positions; diagnostic values of the training script; the same definition as Table~\ref{tab:terms}). Layers = conv 1, conv 2, fc 1 and fc 2 (the last hidden layer) for Continual ImageNet; the first and second hidden layers for Permuted MNIST. Loss = gap to scratch (mean of tasks 400 and 500; points; positive = loss). The selected settings and the loss values are those of runs A and B at learning rates \{0.001, 0.0001\}, and some values differ from J.3 (the reselection of run C) (e.g., L2 Init on Permuted MNIST is −4.6 in J.3).

\textbf{Continual ImageNet}

\begin{center}\footnotesize\setlength{\tabcolsep}{3pt}
\begin{tabular}{@{}P{4.2cm}llll@{}}
\toprule
\textbf{Method} & \textbf{task 1: layers 1–4} & \textbf{task 50: layers 1–4} & \textbf{task 500: layers 1–4} & \textbf{Loss} \\
\midrule
unmodified (0.001) & 0.0 / 0.0 / 33.8 / 24.2 & 0.0 / 9.4 / 20.2 / 35.4 & 0.0 / 0.0 / 35.0 / 90.0 & +28.2 \\
unmodified (0.0001, calibrated) & 0.0 / 0.0 / 16.3 / 22.2 & 0.0 / 0.0 / 29.5 / 22.6 & 0.0 / 0.0 / 21.8 / 48.6 & −4.1 \\
centering + bias suppression & 0.0 / 0.0 / 0.0 / 0.0 & 0.0 / 0.0 / 0.0 / 0.0 & 0.0 / 0.0 / 0.0 / 0.0 & −7.8 \\
kunekune & 0.0 / 0.0 / 0.0 / 0.0 & 0.0 / 0.0 / 0.0 / 0.0 & 0.0 / 0.0 / 0.0 / 0.0 & −2.1 \\
scale cap & 0.0 / 0.0 / 33.7 / 24.0 & 0.0 / 63.1 / 39.4 / 27.2 & 0.0 / 85.6 / 66.9 / 71.3 & +6.8 \\
L2 Init & 0.0 / 0.6 / 17.3 / 17.0 & 0.0 / 0.6 / 23.0 / 9.1 & 0.0 / 0.0 / 14.7 / 0.0 & −7.7 \\
L2 & 0.0 / 0.0 / 29.2 / 18.8 & 0.0 / 4.4 / 52.9 / 43.7 & 0.0 / 0.0 / 29.5 / 30.7 & −6.9 \\
S\&P & 0.0 / 0.0 / 34.3 / 26.3 & 0.0 / 8.1 / 36.1 / 45.8 & 0.0 / 0.0 / 14.7 / 16.4 & −2.5 \\
CBP & 0.0 / 0.0 / 33.3 / 23.2 & 0.0 / 0.6 / 24.4 / 8.6 & 0.0 / 0.6 / 15.4 / 1.0 & −3.9 \\
ReDo & 0.0 / 0.0 / 33.9 / 23.9 & 0.0 / 11.2 / 29.3 / 27.0 & 0.0 / 0.0 / 31.5 / 22.6 & −4.4 \\
Layer Norm & 0.0 / 0.0 / 0.0 / 0.0 & 0.0 / 0.0 / 22.4 / 19.5 & 0.0 / 9.4 / 24.1 / 40.0 & +17.1 \\
\bottomrule
\end{tabular}\end{center}

\textbf{Permuted MNIST}

\begin{center}\footnotesize\setlength{\tabcolsep}{3pt}
\begin{tabular}{@{}P{5.0cm}llll@{}}
\toprule
\textbf{Method} & \textbf{task 1: layers 1–2} & \textbf{task 50: layers 1–2} & \textbf{task 500: layers 1–2} & \textbf{Loss} \\
\midrule
unmodified (0.001, calibrated) & 6.2 / 11.1 & 35.9 / 18.9 & 15.7 / 80.0 & +8.1 \\
centering + bias suppression & 0.0 / 0.0 & 0.0 / 0.0 & 0.0 / 0.0 & +1.7 \\
kunekune & 0.0 / 0.0 & 0.0 / 0.0 & 0.0 / 0.0 & +2.0 \\
row-norm cap on the weights (cap12, not calibrated) & 6.2 / 11.1 & 79.1 / 4.0 & 91.3 / 18.8 & −3.4 \\
L2 Init & 0.3 / 6.4 & 7.3 / 0.8 & 7.2 / 1.5 & −4.7 \\
L2 & 7.7 / 11.0 & 36.0 / 12.3 & 41.4 / 32.1 & −4.3 \\
S\&P & 6.5 / 11.6 & 41.2 / 0.6 & 42.2 / 0.7 & −2.7 \\
CBP & 0.9 / 0.8 & 3.0 / 0.1 & 3.8 / 0.0 & −3.6 \\
ReDo & 6.2 / 11.1 & 15.2 / 5.0 & 18.7 / 4.6 & −3.7 \\
Layer Norm & 0.0 / 0.0 & 1.5 / 3.3 & 2.9 / 27.0 & +10.5 \\
\bottomrule
\end{tabular}\end{center}

(Tables built from the raw logs of the runs.)

\subsection{Per-task online accuracy (Figure~\ref{fig:std}; selection of runs A and B at learning rates \{0.001, 0.0001\}, and the interventions at learning rate 0.001)}
\label{app:J13}
\begin{center}
\includegraphics[width=0.92\textwidth]{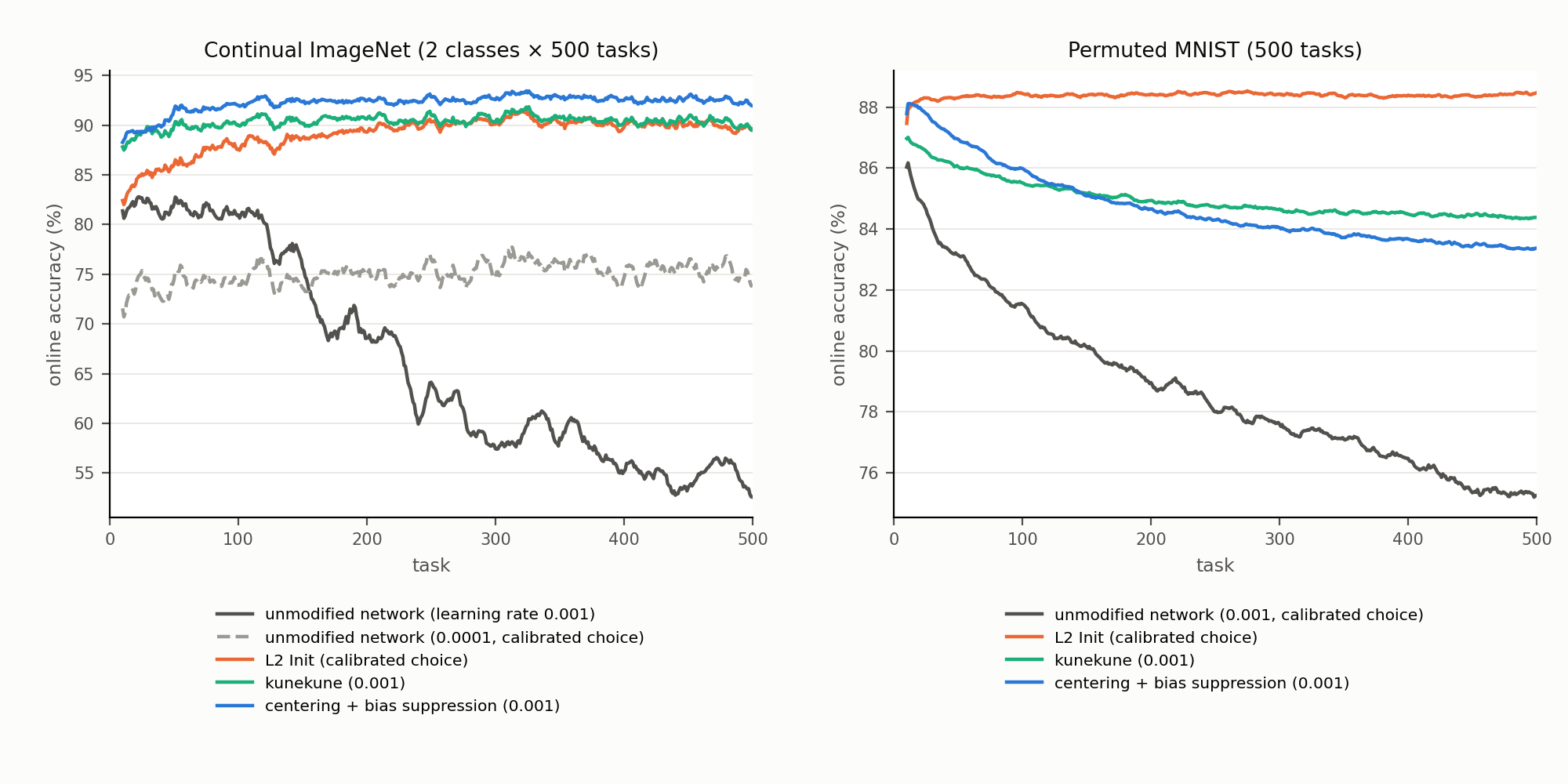}
\captionof{figure}{Online accuracy on the standard task sequences (mean of 10 seeds; 10-task moving average). On Continual ImageNet, the unmodified network at learning rate 0.001 declines; lowering the learning rate to 0.0001 avoids the decline but stays low. Centering + bias suppression and kunekune do not decline at learning rate 0.001. On Permuted MNIST, the unmodified network at learning rate 0.001 declines, and the two interventions reduce the decline but do not stop it.}
\label{fig:std}
\end{center}
(The heights of kunekune and L2 Init are not compared: 0.920 vs 0.913 at task 400 and 0.918 vs 0.919 at task 500, undecided.)

\subsection{Networks with normalization layers (§7, §8)}
\label{app:J14}
\begin{itemize}
\item Question: do networks with normalization layers lose plasticity on the two standard sequences? If they do, does blocking the paths that carry the preactivation mean after normalization remove the loss?
\item Networks and training: the same as in §6 (the Continual ImageNet CNN; the Permuted MNIST MLP 784-100-100-10; Adam; evaluation seeds 0--9; networks trained from scratch at tasks 100--500). The learning rate was not calibrated but fixed at 0.001 (also 0.0001 for Layer Norm on Continual ImageNet), because the question is whether blocking the paths removes the loss, not the ranking of methods.
\item Conditions: bn = batch normalization (BN) placed after the linear (convolutional) map and before the activation of each hidden layer (per unit [per channel including positions for convolutions], subtract the minibatch mean, divide by the standard deviation, multiply by the gain $\gamma$ [initially 1] and add the shift $\beta$ [initially 0]; batch statistics are used in both training and diagnostics). In BN, the per-unit mean of the preactivation over the training batch equals $\beta$ exactly, so the weight and upstream paths of Eq.~\ref{eq:paths} vanish, and the only path that carries the mean is the shift. bn\_b0 = $\beta$ fixed at 0 (all paths closed). ln = the Layer Norm of §6. ln\_ch = ln plus centering (C, subtracting the input mean, and H, subtracting the EMA of the hidden outputs; the linear bias and the shift are learned). ln\_b0 = ln\_ch plus fixing the linear bias and the shift at 0 (all paths closed).
\item Reading rules (fixed before the runs): a loss is present if the mean of the gap to scratch G (mean of tasks 400 and 500) is at least +0.5 points and at least 9 of 10 seeds are positive. J1 = whether the unblocked network has a loss. J2 = the verdict on the difference in preservation $d$ = G(unblocked) − G(blocked) (criterion as in §6.1) together with whether the blocked network has a loss: ``removed'' (X > C and no loss in the blocked network), ``reduced'' (X > C with a remaining loss), ``undecided'' (≈), ``made worse'' (C > X). J3 = split of the Layer Norm paths (secondary). J4 = trace in the last hidden layer at task 500 (descriptive; ``trace present'' if $r\le-0.5$ and death exceeds that of the blocked network). The sha256 of the design and scripts (10 files) were recorded before the runs, and all matched under sha256sum -c after the runs.
\item Verdicts (G and M1 are fractions [0.01 = 1 point]; [ ] are 95\% $t$ intervals; +n/−m are the numbers of positive and negative seeds; r, dphi, beta and gamma are the unit median, the mean derivative, the shift and the gain of the last hidden layer). Excerpt:
\end{itemize}

\begin{Verbatim}[fontsize=\scriptsize,breaklines=true,breakanywhere=true]
# ci (Continual ImageNet) - G = gap to scratch (tasks 400, 500), M1 = total average online, seeds 0-9
  bn_b0|lr=0.001: G +0.1291 [+0.1121, +0.1461] +10/-0 (n 10) | loss present True | M1 +0.8331 [+0.8301, +0.8360] +10/-0 (n 10) | readable True
  bn|lr=0.001: G +0.1096 [+0.0992, +0.1200] +10/-0 (n 10) | loss present True | M1 +0.8490 [+0.8441, +0.8539] +10/-0 (n 10) | readable True
  ln_b0|lr=0.0001: G +0.0398 [+0.0260, +0.0536] +10/-0 (n 10) | loss present True | M1 +0.8282 [+0.8230, +0.8334] +10/-0 (n 10) | readable True
  ln_b0|lr=0.001: G +0.0066 [-0.0183, +0.0315] +8/-2 (n 10) | loss present False | M1 +0.8805 [+0.8764, +0.8846] +10/-0 (n 10) | readable True
  ln_ch|lr=0.001: G +0.0125 [-0.0034, +0.0285] +7/-3 (n 10) | loss present False | M1 +0.8799 [+0.8741, +0.8857] +10/-0 (n 10) | readable True
  ln|lr=0.0001: G +0.1676 [+0.1322, +0.2030] +10/-0 (n 10) | loss present True | M1 +0.6821 [+0.6752, +0.6891] +10/-0 (n 10) | readable True
  ln|lr=0.001: G +0.1778 [+0.1270, +0.2286] +10/-0 (n 10) | loss present True | M1 +0.6817 [+0.6700, +0.6934] +10/-0 (n 10) | readable True
[BN bn|lr=0.001 vs bn_b0|lr=0.001] J1 loss True | J2 made worse (G: C>X -0.0195 [-0.0297, -0.0093] +0/-10 (n 10); blk loss True) | M1: C>X -0.0160 [-0.0211, -0.0108] +0/-10 (n 10)
      J4 last layer l4 task 500: bn|lr=0.001 dead +0.000 m -0.414 r -0.556 dphi +0.299 beta -0.412 gamma +0.759 | bn_b0|lr=0.001 dead +0.000 m +0.000 r +0.000 dphi +0.543 beta +0.000 gamma +0.476 | trace False
sentence: BN lost plasticity on Continual ImageNet (learning rate 0.001) (gap +11.0 points, 10/10), but blocking the paths increased the loss (+12.9). The loss of networks with normalization layers cannot be explained by our paths.
[LN ln|lr=0.001 vs ln_b0|lr=0.001] J1 loss True | J2 removed (G: X>C +0.1712 [+0.1293, +0.2131] +10/-0 (n 10); blk loss False) | M1: X>C +0.1988 [+0.1871, +0.2106] +10/-0 (n 10)
      J4 last layer l4 task 500: ln|lr=0.001 dead +0.835 m -0.327 r -8.282 dphi +0.120 beta -0.192 gamma +1.031 | ln_b0|lr=0.001 dead +0.000 m -0.003 r -0.006 dphi +0.500 beta +0.000 gamma +1.225 | trace True
sentence: Also with a normalization layer, Layer Norm lost plasticity on Continual ImageNet (learning rate 0.001) (gap +17.8 points, 10/10), and fixing the paths of the normalization shift (and bias) at 0 removed the loss (+0.7; difference in preservation +17.1 points, 10/10).
[LN ln|lr=0.0001 vs ln_b0|lr=0.0001] J1 loss True | J2 reduced (G: X>C +0.1278 [+0.0952, +0.1605] +10/-0 (n 10); blk loss True) | M1: X>C +0.1461 [+0.1381, +0.1540] +10/-0 (n 10)
      J4 last layer l4 task 500: ln|lr=0.0001 dead +0.420 m -0.122 r -0.975 dphi +0.472 beta -0.014 gamma +1.029 | ln_b0|lr=0.0001 dead +0.000 m -0.002 r -0.002 dphi +0.495 beta +0.000 gamma +1.032 | trace True
sentence: Also with a normalization layer, Layer Norm lost plasticity on Continual ImageNet (learning rate 0.0001) (gap +16.8 points, 10/10), and blocking the paths reduced the loss, but it remained (+4.0, 10/10).
  [J3 ln_ch|lr=0.001 vs ln|lr=0.001] ln_ch loss False (+0.0125 [-0.0034, +0.0285] +7/-3 (n 10)) | G: X>C +0.1653 [+0.1184, +0.2122] +10/-0 (n 10) | mechanism expectation (inference): ln_ch keeps a loss
  [replication, descriptive] ln|lr=0.0001 (this run) - (run B): M1 -0.0030 [-0.0079, +0.0018] +3/-7 (n 10) | G -0.0032 [-0.0473, +0.0409] +4/-6 (n 10) | M3 -0.0057 [-0.0167, +0.0054] +3/-7 (n 10)

# pm (Permuted MNIST) - G = gap to scratch (tasks 400, 500), M1 = total average online, seeds 0-9
  bn_b0|lr=0.001: G +0.1222 [+0.1185, +0.1260] +10/-0 (n 10) | loss present True | M1 +0.7744 [+0.7740, +0.7747] +10/-0 (n 10) | readable True
  bn|lr=0.001: G +0.0932 [+0.0899, +0.0965] +10/-0 (n 10) | loss present True | M1 +0.7917 [+0.7914, +0.7919] +10/-0 (n 10) | readable True
  ln_b0|lr=0.001: G +0.1073 [+0.1035, +0.1110] +10/-0 (n 10) | loss present True | M1 +0.8176 [+0.8173, +0.8180] +10/-0 (n 10) | readable True
  ln_ch|lr=0.001: G +0.0878 [+0.0837, +0.0919] +10/-0 (n 10) | loss present True | M1 +0.8278 [+0.8275, +0.8282] +10/-0 (n 10) | readable True
  ln|lr=0.001: G +0.1047 [+0.0994, +0.1100] +10/-0 (n 10) | loss present True | M1 +0.8051 [+0.8045, +0.8056] +10/-0 (n 10) | readable True
[BN bn|lr=0.001 vs bn_b0|lr=0.001] J1 loss True | J2 made worse (G: C>X -0.0290 [-0.0324, -0.0256] +0/-10 (n 10); blk loss True) | M1: C>X -0.0173 [-0.0176, -0.0170] +0/-10 (n 10)
      J4 last layer l2 task 500: bn|lr=0.001 dead +0.165 m +0.220 r +0.631 dphi +0.629 beta +0.207 gamma +0.153 | bn_b0|lr=0.001 dead +0.000 m +0.000 r +0.000 dphi +0.492 beta +0.000 gamma +0.268 | trace False
sentence: BN lost plasticity on Permuted MNIST (learning rate 0.001) (gap +9.3 points, 10/10), but blocking the paths increased the loss (+12.2). The loss of networks with normalization layers cannot be explained by our paths.
[LN ln|lr=0.001 vs ln_b0|lr=0.001] J1 loss True | J2 undecided (G: ~ -0.0026 [-0.0083, +0.0031] +5/-5 (n 10); blk loss True) | M1: X>C +0.0126 [+0.0121, +0.0131] +10/-0 (n 10)
      J4 last layer l2 task 500: ln|lr=0.001 dead +0.265 m -0.220 r -0.995 dphi +0.398 beta -0.181 gamma +1.700 | ln_b0|lr=0.001 dead +0.000 m +0.000 r +0.000 dphi +0.489 beta +0.000 gamma +0.539 | trace True
sentence: Layer Norm lost plasticity on Permuted MNIST (learning rate 0.001) (gap +10.5 points, 10/10), but blocking the paths did not change the loss (+10.7). The loss of networks with normalization layers cannot be explained by our paths.
  [J3 ln_ch|lr=0.001 vs ln|lr=0.001] ln_ch loss True (+0.0878 [+0.0837, +0.0919] +10/-0 (n 10)) | G: X>C +0.0169 [+0.0123, +0.0214] +10/-0 (n 10) | mechanism expectation (inference): ln_ch keeps a loss
  [replication, descriptive] ln|lr=0.001 (this run) - (run B): M1 +0.0000 [+0.0000, +0.0000] +0/-0 (n 10) | G +0.0000 [+0.0000, +0.0000] +0/-0 (n 10) | M3 +0.0000 [+0.0000, +0.0000] +0/-0 (n 10)

# abstract phrase (design Sec.5: only when J2 = removed for both BN and LN on Continual ImageNet at lr 0.001): False
\end{Verbatim}

\begin{itemize}
\item Reading (the sentences of §7 and §8 condense the ``sentence:'' lines above into 1--2 sentences, and §7 also states the centering-only result of J3 below):
\begin{itemize}
\item Layer Norm on Continual ImageNet: the loss at learning rate 0.001 (+17.8 points) disappeared when the paths were blocked (+0.7); at 0.0001 it decreased but remained (+16.8 → +4.0). The last hidden layer of the unblocked network had shifted negative and died ($r$ −8.28; death 83.5\%).
\item BN: a loss on both sequences, and fixing the shift at 0 increased the loss (made worse). There is no trace in the sense of J4 ($r\le-0.5$ and more death than the blocked network) (Continual ImageNet: death 0\%, $r$ −0.56; Permuted MNIST: $r$ positive [+0.63], death 16.5\%). ⇒ The loss of the BN network is of a type different from our paths [descriptive; the cause was not investigated].
\item Layer Norm on Permuted MNIST: traces of the negative shift and of death were present ($r$ −0.995; death 26.5\%), but blocking, which brought death to 0, did not change the loss (undecided).
\item J3: on Continual ImageNet, centering alone (ln\_ch) also had no loss (+1.3; positive seeds 7/10, short of the loss criterion [at least 9/10 positive]), and the mechanistic expectation stated before the run [inference], ``with centering alone the shift and bias paths stay open, so a loss remains'', failed (the same pattern as centering without bias suppression in §6.2). In this network as well, the need to block the bias and the shift is not shown. On Permuted MNIST, centering alone left a loss (+8.8; 1.7 points more preserved than the Layer Norm network; 10/10).
\item Replication control (descriptive): Layer Norm on Permuted MNIST (0.001) agreed bit for bit with run B. For Layer Norm on Continual ImageNet (0.0001), the difference from run B is −0.3 points in G, and the interval contains 0. The +17.1 points of Layer Norm in §6.3 is the value of run B; the +16.8 of this section is the value of the same setting rerun in this computation.
\end{itemize}
\end{itemize}

\section{Boundary of the mechanism's scope (Transformers and language models)}
\label{app:K}
\setcounter{subsection}{-1}
\subsection{Sources and limits of the sentences of §7}
\label{app:K0}
\begin{itemize}
\item Verdicts of the image Transformer (``the stage criteria for the input, the upstream and the scale did not hold (1/5, 0/5, 0/5)'' and ``4 of the 8 predictions made before the run failed'' in §7) = the table of verdicts in the experiment record of the image Transformer (J F2 \texttt{SINKS\_VIA\_INPUT} 1/5, J F3 \texttt{UPSTREAM\_CARRIES} 0/5, J F4 \texttt{WIDTH\_DEEPENS} 0/5; J F3 \texttt{PATH\_BLOCKED}, in which the last FFN does not die when the entrance is centered, 5/5) and the rows scoring the predictions (PF2, PF3a, PF4 and PFR0 failed). For J F2, the threshold was determined by measuring on 25 CLS images, while the verdict was measured on all tokens (the original record notes ``an error in the spec; rule unchanged'' in the J F2 row of the verdict table). That the delay at task 30 disappears when only the output layer is reset to 0 comes from a post hoc rerun (R5; predictions written before the run) and does not enter the verdicts. In the table of reruns of the same record, for the network with a centered entrance (SC), resetting the output layer to 0 let all 5 seeds reach the target accuracy even at task 60 (1/5 without centering). The sentence of §7, ``the loss from task 60 on was not reversed in all 5 seeds by any intervention on the side of the output layer and the last FFN'', concerns the network without centering; in the same table, only seed 10 returned from 2,975 to 875 updates when the output layer was reset to 0, and adding Adam initialization to the reset of the output layer let 2 of the 5 seeds reach the target.
\item Limits of ``the interventions that worked acted on the embeddings'' for the language model:
\begin{itemize}
\item The difference −0.086 (5/5) at the single switch (task 2) for the method that equalizes Adam's normalization of the embeddings across tokens (Coupled Adam; \citealp{stollenwerk2025coupled}) is a difference where the unmodified network has no loss at task 2 (gap to scratch −0.003); it is a difference in speed, not a removal of loss. The numbers for the removal of loss are read at cycle 3 and on the long probe (K.3).
\item The worsening of +1.549 (5/5) at cycle 3 for the method that removes the mean of the error backpropagated into the body is a mean pulled up by 2 seeds; the values of the 5 seeds are +0.268, +0.056, +3.341, +0.055, +4.026 (median +0.268). That it worsened in 5/5 is correct, but only for 2 seeds can learning be said to have broken down.
\item The reference value of the loss (gap to scratch of the network without weight decay on the tied embeddings) comes from different runs in the 5-seed table (+0.084) and in the run with the 4 optimizer conditions (+0.076 at cycle 3). Pairwise comparisons ([condition − kept tied]) are read as same-seed pairs within the same run. Larger weight decay in pretraining has been reported to make language models more plastic in fine-tuning \citep{han2026weight}; the conditions with weight decay on the tied embeddings are in the table of all conditions (K.3).
\item The floor of the fluctuation due to batch order (differences of −0.023 to +0.034 between runs with the same initial weights and different batch sequences) is a different yardstick from the threshold for within-run pair effects (0.02). It applies, however, to comparisons across runs (the effects of the output offset on the bodies of 4 layers, 8 layers and Pythia-14m, −0.096, −0.071 and −0.036), so that ordering is kept descriptive.
\end{itemize}
\item On the body of Pythia-14m \citep{biderman2023pythia} (the 1.19 million parameters excluding the embeddings; 5 seeds), the network with tied embeddings and without weight decay also did not learn worse than a network trained from scratch on the new language (−0.0725, 5/5). Whether switching makes learning worse than training from scratch in large networks is not decided at the sizes used in this paper.
\end{itemize}

\subsection{Image Transformer (2 blocks, no layer normalization, shared output layer, 5 seeds)}
\label{app:K1}
The full table of verdicts, the stage-by-stage readings and the table of output-side reruns are in the experiment record of the image Transformer (available from the authors upon request). Sources of the numbers of §7 in that record: all 5 seeds lost adaptation; 1/5, 0/5, 0/5; 4 of the 8 predictions failed; resetting only the output layer at task 30; task 60. We keep the statements that the record declares it does not make: centering at the entrance is not called a decomposition of LN; no generalization to 6 blocks or other depths; no statement about the carrier in the body or about long-term growth laws; post hoc decompositions by the writer and reruns do not enter the verdicts. The sentence of §7, ``the delay at task 30 disappeared when only the output layer was reset to 0'', is R5 of that record (only the head reset to 0; faster than a network trained from scratch in 5/5), which is labeled there as a post hoc rerun (predictions written before the run) and does not enter the record's verdicts (J F2--J F4 of K.0). The sentence of §7, ``the loss from task 60 on was not reversed in all 5 seeds by any intervention on the side of the output layer and the last FFN'', follows the same table (only seed 10 went from 2,975 to 875 with the output layer at 0; adding Adam initialization to the output layer at 0 let 2/5 reach the target). The earlier work of \citet{sun2026vision} (§4.1.2, Finding~2) states that in a 12-block ViT the loss of plasticity strengthens in deep layers and is driven by the collapse of the FFN structure. In our 2-block network without layer normalization, death of the last FFN was not a marker of the delay. Since the depth and the architecture differ, we do not call this a contradiction.

\subsection{Character Transformer (2 blocks, pre-LN, 10 seeds)}
\label{app:K2}
Condensed from the earlier write-up of this experiment (700 states; 53\%; 10/10 on three indicators; $p=.0078$ after Holm over 4 comparisons; recovery ratios 0.37--0.68). Against one of the original result records, we checked $p=0.0078$ after Holm over the 4 comparisons and the median recovery ratios 0.37, 0.54 and 0.68 (not mediation ratios) (they agree). The 700 states and the 53\% were checked only against the write-up; the original record behind them was not opened (unchecked).

\subsection{Language models}
\label{app:K3}
\begin{itemize}
\item All conditions of the 4-layer, width-32, 5-seed model (cycle 3; final loss and AUC; [condition − kept tied]) are in the run records (available from the authors upon request).
\item The 4 conditions that recombine the optimizer of the embeddings (4 layers, 5 seeds). The last section of the excerpt is the difference in training loss on the current language; \textbf{negative = lower loss = learns the current language better}. The value −0.049 (5/5) of Coupled Adam on the long probe (task 24), the gap to scratch on the long probe of +0.025 (unmodified network +0.074; remaining fraction 0.34), and the mean probe loss on task 0 of the method that learns only the embeddings with SGD, 5.08 vs 4.54 nat (mean from the second point on; 5 seeds), were re-extracted from the raw logs of the run with the same definitions as the analysis. Excerpts:
\end{itemize}

\begin{Verbatim}[fontsize=\scriptsize,breaklines=true,breakanywhere=true]
## true LoP (seed mean; end / auc)
tied_cont+wdemb0                   switch1 (task 2): -0.003/-0.122 | cycle 1: +0.039/-0.065 | cycle 2: +0.073/-0.029 | cycle 3: +0.076/-0.010 | long t16-t0: +0.071/+0.050 | long t24-t0: +0.074/+0.062
tied_cont+wdemb0+coupled           switch1 (task 2): -0.089/-0.175 | cycle 1: -0.053/-0.110 | cycle 2: -0.025/-0.048 | cycle 3: -0.008/-0.013 | long t16-t0: +0.025/+0.025 | long t24-t0: +0.025/+0.039
tied_cont+wdemb0+emb_sgd10.0       switch1 (task 2): -0.155/-0.754 | cycle 1: -0.121/-0.690 | cycle 2: -0.085/-0.620 | cycle 3: -0.072/-0.590 | long t16-t0: +0.061/+0.031 | long t24-t0: +0.062/+0.044
tied_cont+wdemb0+gabsent           switch1 (task 2): +0.007/-0.129 | cycle 1: +0.042/+0.363 | cycle 2: +0.084/+0.329 | cycle 3: +0.132/+0.353 | long t16-t0: +0.088/+0.461 | long t24-t0: +0.121/+1.300
tied_cont+wdemb0+gcenter           switch1 (task 2): -0.006/-0.113 | cycle 1: +0.016/-0.061 | cycle 2: +0.372/+0.649 | cycle 3: +1.625/+3.071 | long t16-t0: +0.197/+0.187 | long t24-t0: +2.211/+2.873
\end{Verbatim}

\begin{Verbatim}[fontsize=\scriptsize,breaklines=true,breakanywhere=true]
## the current-language cost: online_last [arm - plain] at the last common task, and online_all of task 1
tied_cont+wdemb0+coupled           task 24 online_last diff mean -0.1256 [-0.111 -0.159 -0.118 -0.137 -0.103] pos 0/5 EFFECT- | task1 online_all diff mean -0.0265 [-0.035 -0.022 -0.012 -0.035 -0.029] pos 0/5 EFFECT-
tied_cont+wdemb0+emb_sgd10.0       task 24 online_last diff mean -0.1295 [-0.118 -0.162 -0.122 -0.125 -0.120] pos 0/5 EFFECT- | task1 online_all diff mean -0.0480 [-0.067 -0.049 -0.050 -0.019 -0.055] pos 0/5 EFFECT-
tied_cont+wdemb0+gabsent           task 24 online_last diff mean +0.0512 [+0.106 -0.007 +0.085 +0.055 +0.017] pos 4/5 mostly+ | task1 online_all diff mean +0.0097 [+0.006 +0.005 +0.017 +0.005 +0.016] pos 5/5 
tied_cont+wdemb0+gcenter           task 24 online_last diff mean +3.2899 [+2.389 +0.007 +12.398 +0.068 +1.588] pos 5/5 EFFECT+ | task1 online_all diff mean +0.0414 [+0.042 +0.050 +0.024 +0.037 +0.053] pos 5/5 EFFECT+
\end{Verbatim}

\begin{itemize}
\item Body of Pythia-14m (the 1.19 million parameters excluding the embeddings; 5 seeds): in the run records. 10 networks with the bias of the final normalization stopped (3 seeds); excerpt:
\end{itemize}

\begin{Verbatim}[fontsize=\scriptsize,breaklines=true,breakanywhere=true]
## (i) [bfin0 - G1 same arm, same seed] (end, cycle 3) and the floor (unchanged A1 s0 in c4 - G1 A1 s0)
tied_cont+wdemb0         mean +0.0050 [+0.016 +0.011 -0.012] pos 2/3 
untied_cont+wdemb0       mean -0.0186 [-0.031 -0.015 -0.010] pos 0/3 
tied_cont+wdemb0+obias   mean +0.0174 [+0.027 +0.022 +0.003] pos 3/3 
floor A1 s0 (c4 unchanged - G1): -0.0088 (end, cycle 3)

## (ii) within-run arm differences (end, cycle 3 and long t24): bfin0 vs G1
A7 - A1 (receiver)     bfin0 cycle3 mean -0.0289 [-0.034 -0.018 -0.035] pos 0/3 EFFECT- | long24 mean -0.0427 [-0.064 -0.036 -0.028] pos 0/3 EFFECT-
A7 - A1 (receiver)     G1    cycle3 mean -0.0413 [-0.045 -0.028 -0.051] pos 0/3 EFFECT- | long24 mean -0.0385 [-0.047 -0.027 -0.041] pos 0/3 EFFECT-
A3 - A1 (untied)       bfin0 cycle3 mean -0.1047 [-0.115 -0.105 -0.094] pos 0/3 EFFECT- | long24 mean -0.0683 [-0.096 -0.052 -0.057] pos 0/3 EFFECT-
A3 - A1 (untied)       G1    cycle3 mean -0.0811 [-0.068 -0.080 -0.096] pos 0/3 EFFECT- | long24 mean -0.0533 [-0.043 -0.051 -0.066] pos 0/3 EFFECT-
\end{Verbatim}

\begin{itemize}
\item Ordering of the size of the effect of the output offset (mean at the end of the cycle; [condition − kept tied]): 4 layers −0.096 (3 seeds; the −0.0955 of the analysis output lies at a rounding boundary, so it was re-extracted from the raw log as −0.09553; in the 5-seed run S5 of the same 4 layers it is −0.086; 4 and 8 layers are compared between 3-seed runs of the same procedure), 8 layers −0.071 (3 seeds), and the body of Pythia-14m −0.036 (5 seeds). As fractions of the loss of the network trained from scratch: 2.3\%, 1.7\% and 1.0\% (denominator = the final loss of the kept-tied network trained from scratch of each network).
\end{itemize}

\subsection{Where the common-error gradient lands (analysis of saved weights; descriptive; outside §7)}
\label{app:K4}
\begin{itemize}
\item Question and setting: in the first update at a task switch, where in the parameters does the part of the gradient that does not depend on the label assignment (the left side of identity B.1 = the gradient with the correct labels of the new task replaced by the frequencies of the classes [tokens] in the minibatch) land? The setting (quantities, groups, reading, stopping conditions) was fixed by an independent checker before the results. No new training was done (CPU, 8 threads).
\item Quantities (per group G): Q1 = squared share of the common-error gradient; Q2 = directional agreement (cos) with the actual gradient, and the ratio of the label term; Q3 = fraction of coordinates where the signs of the actual and common-error gradients agree and the common-error part exceeds the label term; \textbf{Q4 (primary) = the first-order decrease of the loss by the first step $\Delta\theta$, $-\langle g_G,\Delta\theta_G\rangle$, split into the common-error part and the label-term part}; Q5 = the part of the common-error gradient that is the first term of identity B.1 (the part due to the mean output error). First step = for the language model, which reinitializes Adam at each task, $\Delta\theta=-\eta\,\mathrm{sign}(g)$; for the CNN, an Adam step using the saved state ($m$, $v$). Expected gradient = mean over 8 minibatches.
\item Networks: language model = the 4-layer, width-32 network of K.3 (kept tied, 5 seeds; ends of cycles 1--3 = tasks 8, 16, 24; switch to the new language vi; as controls, the 4 arms with the same seeds, untied networks with 5 seeds [2 + 3 seeds from two runs] and the kept-tied network of the same run). CNN = the unmodified network at learning rate 0.001 on Continual ImageNet of §6 (10 seeds; tasks 10→11 and 490→491).
\item Checks: weight-only diagnostic values recomputed from the saved weights differ from the run records by at most 5.0e−05 (language model; 105 network × time points; threshold 1e−4), and the fraction of dead units of the CNN by at most 2.9e−08. The residual of the identity (actual = common error + label term) is at most 2.8e−06 times the maximum gradient. Non-finite values: 0. Minibatches were generated with a CPU generator (for the language model not the same minibatches as in the run; for the CNN the same). The spread across seeds was larger than that across minibatches (K = 8 suffices, as stated in the setting).
\item Results (excerpt of the analysis output):
\end{itemize}

\begin{Verbatim}[fontsize=\scriptsize,breaklines=true,breakanywhere=true]
## Computation A (language model, kept tied tied_cont+wdemb0, c3_main_v2, 5 seeds, switch to the probe language vi, expected gradient K=8)
t8 Q4 frequency part / total first-step decrease: body 0.894 (0.868-0.923, n=5) token table 0.006 (0.004-0.007, n=5) gains 0.009 (0.004-0.024, n=5)
t8 Q1 squared share of the frequency part: body 0.949 (0.447-0.998, n=5) token table 0.000 (0.000-0.000, n=5) gains 0.051 (0.002-0.553, n=5)
t16 Q4 frequency part / total first-step decrease: body 0.923 (0.890-1.006, n=5) token table 0.005 (0.005-0.010, n=5) gains 0.011 (0.008-0.014, n=5)
t16 Q1 squared share of the frequency part: body 0.910 (0.753-0.955, n=5) token table 0.000 (0.000-0.000, n=5) gains 0.090 (0.045-0.247, n=5)
t24 Q4 frequency part / total first-step decrease: body 0.897 (0.870-1.069, n=5) token table 0.006 (0.003-0.008, n=5) gains 0.010 (0.006-0.017, n=5)
t24 Q1 squared share of the frequency part: body 0.945 (0.833-0.978, n=5) token table 0.000 (0.000-0.001, n=5) gains 0.054 (0.022-0.167, n=5)
t8 untied networks (S5 2 seeds + X4 3 seeds), embeddings (input + output), Q4 frequency part: 0.101 (0.088-0.112, n=5) / kept tied, same run: 0.007 (0.004-0.011, n=5)
t16 untied networks (S5 2 seeds + X4 3 seeds), embeddings (input + output), Q4 frequency part: 0.082 (0.063-0.108, n=5) / kept tied, same run: 0.007 (0.004-0.014, n=5)
t24 untied networks (S5 2 seeds + X4 3 seeds), embeddings (input + output), Q4 frequency part: 0.076 (0.061-0.105, n=5) / kept tied, same run: 0.005 (0.004-0.011, n=5)
1 batch (120 rows = 5 seeds x 3 time points x 8 batches): fraction of token-table coordinates in rows of tokens absent from the batch targets 0.781 (0.770-0.797, n=120), Q3 of those rows 1.000 (0.999-1.000, n=120), cos(actual, frequency) 1.000 (0.999-1.000, n=120), first-step decrease of those rows / total 0.002 (0.000-0.004, n=120)
number of coordinates: token table 131072, body 49152, gains 288 (token-table share 0.726)
t8 Q5 squared mean-error part / squared frequency part: last layer (o, up, dn)0.964 (0.755-1.126, n=15), first layer 0.709 (0.433-1.007, n=20), whole body 0.956 (0.720-1.012, n=5)
t16 Q5 squared mean-error part / squared frequency part: last layer (o, up, dn)0.921 (0.369-1.231, n=15), first layer 0.369 (0.104-0.988, n=20), whole body 0.742 (0.240-0.951, n=5)
t24 Q5 squared mean-error part / squared frequency part: last layer (o, up, dn)0.951 (0.155-1.138, n=15), first layer 0.482 (0.045-1.923, n=20), whole body 0.696 (0.175-0.816, n=5)

## Control B (Continual ImageNet CNN, unmodified network at learning rate 0.001, 10 seeds, first 8 batches of task t+1, Adam step with the saved m, v)
t10: networks whose hidden layers receive gradient with first-step loss change >= 1e-4: 10/10. of these, hidden > output 10/10, hidden frequency part / total 0.711 (0.212-1.860, n=10), output layer 0.046 (0.012-0.065, n=10), Q5 hidden squared mean-error part / squared frequency part 0.532 (0.025-1.006, n=10), squared share along the input-mean direction of the frequency gradient of fully connected hidden-unit weights (f1, f2)0.890 (0.396-0.976, n=20)
t490: networks whose hidden layers receive gradient with first-step loss change >= 1e-4: 2/10. of these, hidden > output 2/2, hidden frequency part / total 0.475 (0.386-0.564, n=2), output layer 0.066 (0.024-0.108, n=2), Q5 hidden squared mean-error part / squared frequency part 0.035 (0.014-0.055, n=2), squared share along the input-mean direction of the frequency gradient of fully connected hidden-unit weights (f1, f2)0.649 (0.489-0.926, n=4)
\end{Verbatim}

\begin{itemize}
\item Primary count of the setting: in the kept-tied language model, the number of seeds in which the common-error part of Q4 was larger in the embeddings (sum of rows a--c) than in the body (attention and FFN matrices) was 0/5, 0/5 and 0/5 at tasks 8, 16 and 24 (0 also for the 4 arms, the untied networks and the switch to the next training language). For the CNN at tasks 10→11, hidden > output in 10/10. ⇒ ``body'' under the reading fixed in the setting. Therefore the main text does not claim that the parameters that receive the common error determine the form of the loss and the effective interventions, and it states only where the effective interventions acted (§7) and the distribution measured here.
\item Where the results split (per the rule of the setting, ``if it splits, write that it split''): in loss decrease (Q4) and size (Q1) the body receives most, but in direction (Q3) the rows of the embedding for tokens that do not appear among the minibatch targets (about 78\% of the embedding coordinates) have their direction determined by the common error alone (Q3 and cos both 1.000). Because the first step of the language model moves each coordinate by nearly the same size, these rows move a lot even though their contribution to the loss decrease is 0.2\%. In large language models, the output-embedding rows of tokens absent from the new data likewise receive one-sided softmax gradients that Adam's normalization turns into nearly full-sized steps \citep[§4.3]{han2026deafening}.
\item {[Post hoc]} (outside the setting; not used for verdicts): (i) the count is the same with the first Adam step including $\epsilon$, $-g/(|g|+10^{-8})$ (0/5). (ii) Counting with each arm's own first-step rule, the embedding share is 0.08--0.20 in the Coupled Adam network and 0.09--0.62 in the network that learns the embeddings with SGD (embedding > body only for 1 seed at task 8). (iii) In the CNN at task 490, 7 of the 10 seeds had all units of the last hidden layer dead, and the hidden layers receive no gradient (1 seed had 91\% dead with a first-step loss change of 3e−5). In the 2 seeds remaining after excluding these 8, hidden > output (exclusion line = first-step loss change below 1e−4 or zero hidden gradient; a line placed after seeing the results). (iv) ``Tokens appearing among the minibatch targets'' for the embedding share were, for the expected gradient, tokens appearing in any of the 8 minibatches.
\item Limits of the reading: all of these are descriptions of saved states, not interventions. Differences across cycles and arms are concurrent associations with the size of the loss and are not read as causes. These are first-step quantities only; the accumulation of updates within the task was not measured.
\end{itemize}

\end{document}